\documentclass[ejs]{imsart}

\RequirePackage[OT1]{fontenc}
\RequirePackage{amsthm,amsmath}
\RequirePackage[numbers]{natbib}
\RequirePackage[colorlinks,citecolor=blue,urlcolor=blue]{hyperref}
\usepackage{graphicx}
\usepackage{amsmath,amsfonts,graphicx,rotating}
\usepackage{amssymb, mathtools,cases}
\usepackage{subcaption,color,url}
\usepackage{algorithmicx}
\usepackage{algorithm}
\usepackage{algpseudocode,enumerate}

\newcommand{\bd}{\boldsymbol}
\newcommand{\mb}{\mathbf}

\newcommand{\be}{\begin{equation}}
	\newcommand{\ee}{\end{equation}}

\DeclareMathOperator*{\argmin}{arg\,min}
\DeclareMathOperator*{\argmax}{arg\,max}

\DeclareMathOperator{\diag}{diag}
\DeclareMathOperator{\vdg}{vdg}

\DeclareMathOperator{\rank}{rank}
\DeclareMathOperator{\cov}{cov}
\DeclareMathOperator{\corr}{corr}
\DeclareMathOperator{\var}{var}

\DeclareMathOperator{\tr}{tr}
\DeclareMathOperator{\lspan}{span}

\DeclareMathOperator{\colsp}{colsp}

\DeclareMathOperator{\SNR}{SNR}

\newtheorem{theorem}{Theorem}

\newtheorem{definition}{Definition}
\newtheorem{remark}{Remark}
\newtheorem{assump}{Assumption}

\begin{document}

\begin{frontmatter}
\title{CDPA: Common and Distinctive Pattern Analysis between High-dimensional Datasets\thanksref{T1}\thanksref{T2}}
\runtitle{Common and Distinctive Pattern Analysis}
\thankstext{T1}{Dr. Shu's research was partially supported by the grant R21AG070303 from the National Institutes of Health and a startup fund from New York University. The content is solely the responsibility of the authors and does not necessarily represent the official views of the National Institutes of Health, New York University, or Tulane University.
}
\thankstext{T2}{This is the author's final version of the article published in {\it Electronic Journal of Statistics}, 2022, 16 (1), 2475--2517. The publisher's version is available at \url{https://doi.org/10.1214/22-EJS2008}. 
}

\begin{aug}
\author{\fnms{Hai} \snm{Shu}\ead[label=e1]{hs120@nyu.edu}}

\address{Department of Biostatistics, School of Global Public Health, New York University\\
\printead{e1}}

\author{\fnms{Zhe} \snm{Qu}
\ead[label=e3]{third@somewhere.com}
\ead[label=u1,url]{www.foo.com}}

\address{Department of Mathematics, School of Science and Engineering, Tulane University}

\runauthor{H. Shu and Z. Qu}

\end{aug}

\begin{abstract}
A representative model in integrative analysis of two high-dimensional correlated datasets is to decompose each data matrix into a low-rank common matrix generated by latent factors shared across datasets, a low-rank distinctive matrix corresponding to each dataset, and an additive noise matrix. 
Existing decomposition methods claim that their common matrices capture the common pattern of the two datasets. However, their so-called common pattern only denotes the common latent factors but ignores the common pattern between the two coefficient matrices of these common latent factors.
We propose a new unsupervised learning method, called the common and distinctive pattern analysis (CDPA),  which appropriately defines the two types of data patterns by further incorporating
the common and distinctive patterns of the coefficient matrices.
A consistent estimation approach is developed for high-dimensional settings, and shows reasonably good finite-sample performance in simulations.
Our simulation studies and real data analysis
corroborate that the proposed CDPA
can provide better characterization
of common and distinctive patterns 
and thereby benefit data mining.
\end{abstract}


\tableofcontents
\end{frontmatter}

\section{Introduction}\label{sec: intro}
Modern biomedical studies often collect multiple types of large-scale datasets on a common set of objects \citep{Craw16,Jens17}. 
For example, The Cancer Genome Atlas (TCGA) \citep{Hoad18} 
collected for tumor samples the multi-platform genomic data such as mRNA expression and DNA methylation;
the Human Connectome Project (HCP) \citep{Van13} 
acquired 
multi-modal brain imaging data, including structural MRI and functional MRI,
from healthy adults.
The use of  multiple data types can allow us to  enhance  understanding the mechanisms underlying complex diseases like cancers 
\citep{Kobo12,Camp18}
and neurodegenerative diseases \citep{Wein13,Saee17}, or to improve the performance in various learning tasks such as clustering and classification \citep{Kloe16, Smil17}.

The most straightforward approach to the integrative analysis of multi-type datasets is to concatenate all their data matrices into one matrix and then implement standard data analysis tools. One such example is the simultaneous component analysis (SCA) \citep{Smil03}, which applies the principal component analysis (PCA) to the concatenated data matrix and thus is also known as SUM-PCA.
These methods are simple to implement, but they are unable to explore or interpret the relationships among datasets.
As pioneers to overcome this drawback, the canonical correlation analysis (CCA) \citep{Hote36,mai2019iterative} 
and its various generalizations \citep{Carr68,Kett71,tenenhaus2011regularized}
measure the correlations and extract the most correlated components among datasets.
The CCA methods only account for correlated features and fail to reveal a more detailed relationship on the common and distinctive patterns across datasets.

A family of data integration methods has emerged recently 
to identify and separate the common and distinctive variations across datasets, including orthogonal $n$-block partial least squares (OnPLS) \citep{Lofs11}, 
distinctive and common components with SCA (DISCO-SCA) \citep{Scho13},
common orthogonal basis extraction (COBE) \citep{Zhou16},
joint and individual variation explained (JIVE) \citep{Lock13},
angle-based	JIVE (AJIVE) \citep{Feng18}, 
and decomposition-based CCA (D-CCA) \citep{Shu18}. 
Consider the case with two datasets. 
All these methods decompose each data matrix into a low-rank {\it common matrix} generated by latent factors shared across datasets,\footnote{\label{ft: OnPLS}
		The common matrices of OnPLS may have different sets of latent factors. As a post-processing step \citep{Kloe16}, 
		the same set of latent factors can be obtained as
		an  orthonormal basis of the vector space spanned by all these sets of latent factors. The common matrices remain unchanged after this post-processing step. 
}
a low-rank {\it distinctive matrix} corresponding to each dataset, and an additive noise matrix.
Despite different constraints in the decomposition,
these methods refer the common pattern of the two datasets to 
the common latent factors, but ignore the common pattern
between the two 
coefficient matrices of these common latent factors.
It may be more appropriate to name their ``common" and ``distinctive" matrices as {\it common-source} and {\it distinctive-source matrices}.

We propose a new unsupervised learning method, called the common and distinctive pattern analysis (CDPA), to improve the delineation of the common and distinctive patterns between two correlated datasets.
The CDPA method defines the common pattern by incorporating both the common latent factors and
the common pattern of their coefficient matrices, and
determines each distinctive pattern as the residual part of the corresponding signal dataset.
In factor analysis \citep{Harm76}, a coefficient matrix
of latent factors
is called
a {\it factor pattern matrix},
containing the factor loadings (i.e., coefficients)
that represent the contributions of 
latent factors
to the signal data.
A coefficient matrix is also known as a {\it mixing channel} in signal processing \citep{Papa00, Parr03}
which introduces correlations into the uncorrelated 
source variables to generate the output data.
Hence, 
the two coefficient matrices
of the common latent factors 
for two correlated datasets
may contain common and distinctive patterns of the ways in which these common latent factors 
form their corresponding common-source matrices. Such
common and distinctive patterns in the two coefficient matrices are also important and
should be separated into the common and distinctive patterns of the two datasets.
Our defined common-pattern and distinctive-pattern matrices together with the aforementioned common-source and distinctive-source matrices 
constitute a more comprehensive picture that depicts the relationship of two datasets.

Three challenging issues arise in the construction
and estimation of common-pattern and distinctive-pattern matrices: 
(i) There exists the row matching problem of the two coefficient matrices, or 
equivalently the variable pairing problem of the two datasets, if the rows of either observed data matrix can be arbitrarily ordered independent of the other matrix;
(ii) The common pattern of the two coefficient matrices must be identified;
(iii) Recovering the high-dimensional common-pattern and distinctive-pattern matrices confronts the curse of dimensionality
where the unknown large covariance matrices may not be consistently estimated by the traditional sample covariance matrices \citep{Yin88}.
We successfully convert the row matching problem (i) into the classic graph matching problem~\citep{Lu16}. We extract
the common pattern in (ii) by
our extended analogy of the state-of-the-art D-CCA. 
To address the challenge (iii), 
we develop consistent estimators of proposed common-pattern and distinctive-pattern matrices under the high-dimensional spiked covariance model \citep{Fan13,Wang17,Shu18}, which has been widely used in
various fields, such as signal processing \citep{Nada10}, machine learning \citep{Huang17}, and economics~\citep{Cham83}.

The rest of this article is organized as follows.
Section~\ref{CDPA sec: pre} introduces the CCA and D-CCA methods as preliminaries.
Our CDPA method and its consistent estimation are established in Section~\ref{CDPA sec: method}.
Section~\ref{CDPA sec: simulations} examines
the finite-sample performance of proposed estimators via simulations.
We also compare CDPA with six D-CCA-type methods
through simulated data in Section~\ref{CDPA sec: simulations} and
through two real-data examples from HCP and TCGA in Section~\ref{CDPA sec: real data}. 
Section~\ref{sec: discussion} concludes with discussion.
All theoretical proofs and additional simulation and real-data results are provided in Appendices.
A Python package for the proposed CDPA method is available at \url{https://github.com/shu-hai/CDPA}.

\section{Preliminaries}\label{CDPA sec: pre}

Let $\mb{Y}_k\in \mathbb{R}^{p_k\times n}$ for $k\in\{1,2\}$ be the $k$-th dataset obtained on a common set of $n$ objects, where $p_k$ is the number of variables. 
The decomposition model considered in aforementioned existing methods (e.g., D-CCA) is
\be\label{decomp mat}
\mb{Y}_k=\mb{X}_k+\mb{E}_k=\mb{C}_k+\mb{D}_k+\mb{E}_k\in \mathbb{R}^{p_k\times n}
\ee
for which the $n$ columns of each matrix are assumed to be independent and identically distributed
(i.i.d.) copies of the corresponding mean-zero random vector in 
\be\label{decomp vec}
\bd{y}_k =\bd{x}_k+\bd{e}_k=\bd{c}_k+\bd{d}_k+\bd{e}_k\in\mathbb{R}^{p_k}
\ee
where $\{\mb{X}_k,\bd{x}_k\}_{k=1}^2$ and $\{\mb{E}_k,\bd{e}_k\}_{k=1}^2$ are signals and noises, respectively, $\{\mb{C}_k\}_{k=1}^2$ and $\{\bd{c}_k\}_{k=1}^2$ are common-source matrices and random vectors that are generated from 
the common latent factors of the two datasets, and 
$\mb{D}_k$ and $\bd{d}_k$ are the distinctive-source matrix and random vector from distinctive latent factors of the $k$-th dataset. 
Write each $k$-th common-source random vector by
$\bd{c}_k=\mb{B}_k(c_1,\dots,c_{L_{12}})^\top$,
where $c_1,\dots,c_{L_{12}}$ are the common latent factors
and $\mb{B}_k$ is their coefficient matrix.
The common pattern of $\mb{B}_1$ and $\mb{B}_2$ is not considered by the existing methods, which motivates our current research.

We start with signal vectors $\{\bd{x}_k\}_{k=1}^2$ for simplicity, and introduce the CCA and D-CCA methods in the two following subsections.
The signal estimation is deferred to Section~\ref{sec: estimation}.

{\bf Notation.}
For any matrix $\mb{M}=(M_{ij})_{1\le i\le p,1\le j\le n}\in \mathbb{R}^{p\times n}$, denote the $\ell$-th largest singular value
and the $\ell$-th largest eigenvalue (if $p=n$) by
$\sigma_\ell(\mb{M})$ and $\lambda_\ell(\mb{M})$ respectively,
the spectral norm $\| \mb{M} \|_2=\sigma_1(\mb{M})$,
the Frobenius norm $\| \mb{M} \|_F=\sqrt{\sum_{i=1}^p\sum_{j=1}^nM_{ij}^2}$,
the matrix $\mathcal{L}^{\infty}$ norm $\|\mb{M}  \|_\infty=\max_{1\le i\le p}\sum_{j=1}^n |M_{ij}|$, and the max norm $\|\mb{M}  \|_{\max}=\max_{i,j}|\mb{M}_{ij}|$.
Let $\mb{M}^{[s:t,u:v]}$, $\mb{M}^{[s:t,:]}$ and $\mb{M}^{[:,u:v]}$ denote the submatrices 
$(M_{ij})_{s\le i\le t, u\le j\le v}$, $(M_{ij})_{s\le i\le t, 1\le j\le n}$
and $(M_{ij})_{1\le i\le p, u\le j\le v}$ of $\mb{M}$, respectively.
Write the Moore-Penrose pseudoinverse and the column space
of $\mb{M}$ by $\mb{M}^\dag$
and $\colsp(\mb{M})$, respectively.
Let $[\mb{M}_1;...;\mb{M}_L]=[\mb{M}_1^\top,\dots,\mb{M}_L^\top]^\top$ for matrices $\mb{M}_1,\dots, \mb{M}_L$ with the same number of columns.
Denote the $j$-th entry of a vector $\bd{v}\in\mathbb{R}^p$ by $\bd{v}^{[j]}$.
Write
$\lspan(\bd{v}^\top)=\lspan(\{\bd{v}^{[j]}\}_{j=1}^p)=\{\sum_{j=1}^p a_j \bd{v}^{[j]}: \forall a_j\in \mathbb{R}\}$.
For vectors $\bd{v}_1,\dots,\bd{v}_L$ of same length, let $[\bd{v}_\ell]_{\ell=1}^L=(\bd{v}_1,\dots,\bd{v}_L)$.
Denote the angle between two elements $v_1$ and $v_2$ in an inner product space by
$\theta(v_1,v_2)$.
Let $(\mathcal{L}_0^2,\cov)$ be the inner product space
composed of all real-valued random variables with zero mean and finite variance, and endowed with the covariance operator as the inner product.
Let both $x:=y$ and $y=:x$ mean that $x$ is defined by $y$.
Write $a\propto b$ if $a=\kappa b$ for some constant $\kappa \in \mathbb{R}$.
Denote $a\wedge b=\min(a,b)$ and $a\vee b=\max(a,b)$.
For signal vectors $\{\bd{x}_k\}_{k=1}^2$, 
denote $\mb{\Sigma}_k=\cov(\bd{x}_k)$,
$\mb{\Sigma}_{12}=\cov(\bd{x}_1,\bd{x}_2)$,
$r_k=\rank(\mb{\Sigma}_k)$,
$r_{\min}=r_1 \wedge r_2$,
$r_{\max}=r_1\vee r_2$,
and $r_{12}=\rank(\mb{\Sigma}_{12})$.
Throughout the paper, 
our asymptotic arguments are by
default under $n \to \infty$.

\subsection{Canonical correlation analysis}\label{subsec: CCA}
The CCA method \citep{Hote36} sequentially finds the most correlated variables, called {\it canonical variables}, between the two subspaces 
$\{\lspan(\bd{x}_k^\top)\}_{k=1}^2$ in $(\mathcal{L}_0^2,\cov)$. For $1\le \ell\le  r_{12}$, the $\ell$-th pair of  canonical variables are defined as
\be
\begin{split}\label{CCA def}
	&
	\{z_{1\ell},z_{2\ell}\}	\in \argmax_{\{z_k\}_{k=1}^2}\ \corr(z_1,z_2)\quad \text{subject to}\\
	&\quad \var(z_k)=1 \ \text{and}\ z_k\in	\lspan(\bd{x}_k^\top)\setminus\lspan(\{z_{km}\}_{m=1}^{\ell-1}),
\end{split}
\ee
where $\lspan(\bd{x}_k^\top)\setminus\lspan(\{z_{km}\}_{m=1}^0):=\lspan(\bd{x}_k^\top)$,
and 
for $\ell>1$,
$\lspan(\bd{x}_k^\top)\setminus\lspan(\{z_{km}\}_{m=1}^{\ell-1})$
denotes
the 
orthogonal complement of $\lspan(\{z_{km}\}_{m=1}^{\ell-1})$ in $\lspan(\bd{x}_k^\top)$.
The correlation $\rho_\ell:=\corr(z_{1\ell},z_{2\ell})$  is called the $\ell$-th {\it canonical correlation} of $\bd{x}_1$ and $\bd{x}_2$.
Augment 
$\{z_{k\ell}\}_{\ell=1}^{r_{12}}$ with any $(r_k-r_{12})$ standardized variables to be $\bd{z}_k=(z_{k1},\dots,z_{kr_k}   )^\top$
such that $\bd{z}_k^\top$ is an orthonormal basis of $\lspan(\bd{x}_k^\top)$.
We have the bi-orthogonality \citep{Shu18} that 
\be\label{bi-orthogonality}
\cov(\bd{z}_1,\bd{z}_2)=\diag(\rho_1,\dots,\rho_{r_{12}},\mb{0}_{(r_1-r_{12})\times (r_2-r_{12})}).
\ee
The augmented canonical variables $\{\bd{z}_k\}_{k=1}^2$ can be obtained by 
$
\bd{z}_k=\mb{U}_{\theta k}^\top \bd{z}_k^*,
$
where $\bd{z}_k^* =\mb{\Lambda}_k^{-1/2}\mb{V}_k^\top \bd{x}_k$,
$\mb{\Sigma}_k=\mb{V}_k\mb{\Lambda}_k\mb{V}_k^\top$ is a compact 
singular value decomposition (SVD)
with $\mb{\Lambda}_k=\diag(\sigma_1(\mb{\Sigma}_k),\dots,\sigma_{r_k}(\mb{\Sigma}_k))$,
and  $
\mb{\Theta}:=\cov(\bd{z}_1^*,\bd{z}_2^*  )= \mb{U}_{\theta 1} \mb{\Lambda}_{\theta} \mb{U}_{\theta 2}^\top
$ is a full SVD with $\mb{\Lambda}_{\theta}=\diag(\rho_1,\dots,\rho_{r_{12}},\mb{0}_{(r_1-r_{12})\times (r_2-r_{12})})$.
Note that in the inner product space 
$(\mathcal{L}_0^2,\cov)$,
$\cos\theta(\cdot,\cdot)=\corr(\cdot,\cdot)$ and $\|\cdot\|=\sqrt{\var(\cdot)}$.

A similar method to CCA is the principal angle analysis (PAA) \citep{Bjor73}, which investigates the closeness of any two subspaces, denoted by $F$ and $G$,  in the Euclidean dot product space $(\mathbb{R}^p,\cdot)$. 
For $1\le \ell\le q:=\min\{\dim(F),\dim(G)\}$,
the $\ell$-th {\it principal angle} $\theta_\ell\in[0,\pi/2]$ between $F$ and $G$ is defined  by 
\be\label{PAA def}
\begin{split}
	&\cos\theta_\ell=\max_{\bd{u}\in F}\max_{\bd{v}\in G} \bd{u}^\top\bd{v}=\bd{u}_\ell^\top\bd{v}_\ell
	\quad \text{subject to}\\
	& \|\bd{u}\|_F=\|\bd{v}  \|_F=1,
	~\text{and}~
	\bd{u}^\top  \bd{u}_j=\bd{v}^\top \bd{v}_j=0
	~\text{for}~
	j=1,\dots, \ell-1.
\end{split}
\ee
The vectors $\{\bd{u}_\ell,\bd{v}_\ell\}$ are called the $\ell$-th pair of {\it principal vectors} of $F$ and $G$.
Let $\mb{Q}_F$ and $\mb{Q}_G$ be the matrices 
whose columns form the orthonormal bases of $F$ and $G$, respectively.
The principal angles and principal vectors can be obtained
by
\be\label{PA,PV}
\cos \theta_\ell =\sigma_\ell (\mb{Q}_F^\top \mb{Q}_G),
\quad
(\bd{u}_1,\dots,\bd{u}_q)=\mb{Q}_F\mb{U}_{Q},
\quad
(\bd{v}_1,\dots,\bd{v}_q)=\mb{Q}_G\mb{V}_{Q},
\ee
where
$\mb{Q}_F^\top \mb{Q}_G=\mb{U}_{Q}\diag\{\sigma_1(\mb{Q}_F^\top \mb{Q}_G),\dots,\sigma_q(\mb{Q}_F^\top \mb{Q}_G) \}\mb{V}_{Q}^\top$
is a SVD of $\mb{Q}_F^\top \mb{Q}_G$.

The PAA and CCA methods are essentially the same except their respective inner product spaces $(\mathbb{R}^p,\cdot)$ and $(\mathcal{L}_0^2,\cov)$. 
The principal vectors and the cosines of principal angles of PAA correspond to 
the canonical variables and the canonical correlations of CCA.
The cosines of principal angles are also called canonical correlations in PAA \citep{Bjor73}.
Similar to \eqref{bi-orthogonality},
the bi-orthogonality between different pairs of principal vectors also holds.

\subsection{Decomposition-based canonical correlation analysis}\label{subsec: D-CCA}
For random vectors $\{\bd{x}_k\}_{k=1}^2$, the D-CCA method \citep{Shu18}
aims to decompose each $\bd{x}_k$ into a common-source vector $\bd{c}_k$ and a distinctive-source vector $\bd{d}_k$ by
\be\label{x=c+d}
\bd{x}_k=\bd{c}_k+\bd{d}_k
\ee
subject to three desirable constraints in
$(\mathcal{L}_0^2,\cov)$:
 \begin{numcases}{}
	\lspan(\bd{c}_1^\top)= \lspan(\bd{c}_2^\top),\nonumber\\
	\lspan(\bd{d}_1^\top)\perp \lspan(\bd{d}_2^\top),\label{contr space}\\
	\lspan([\bd{x}_1;\bd{x}_2]^\top)=\lspan([\bd{c}_1;\bd{c}_2;\bd{d}_1;\bd{d}_2]^\top).\nonumber
\end{numcases}

\begin{figure}[hb!]	
	\begin{subfigure}{1\textwidth}
		\centering
		\includegraphics[width=0.5\textwidth]{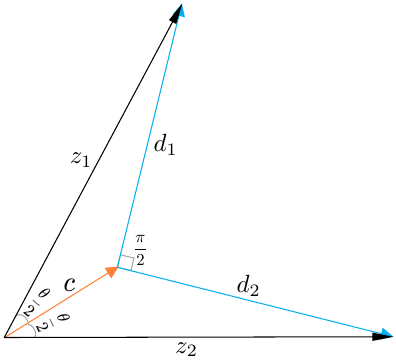}
		\bigskip
		\caption{The geometry of D-CCA for two standardized random variables.}
	\end{subfigure}
	\begin{subfigure}{1\textwidth}
		\centering
		\includegraphics[width=0.9\textwidth]{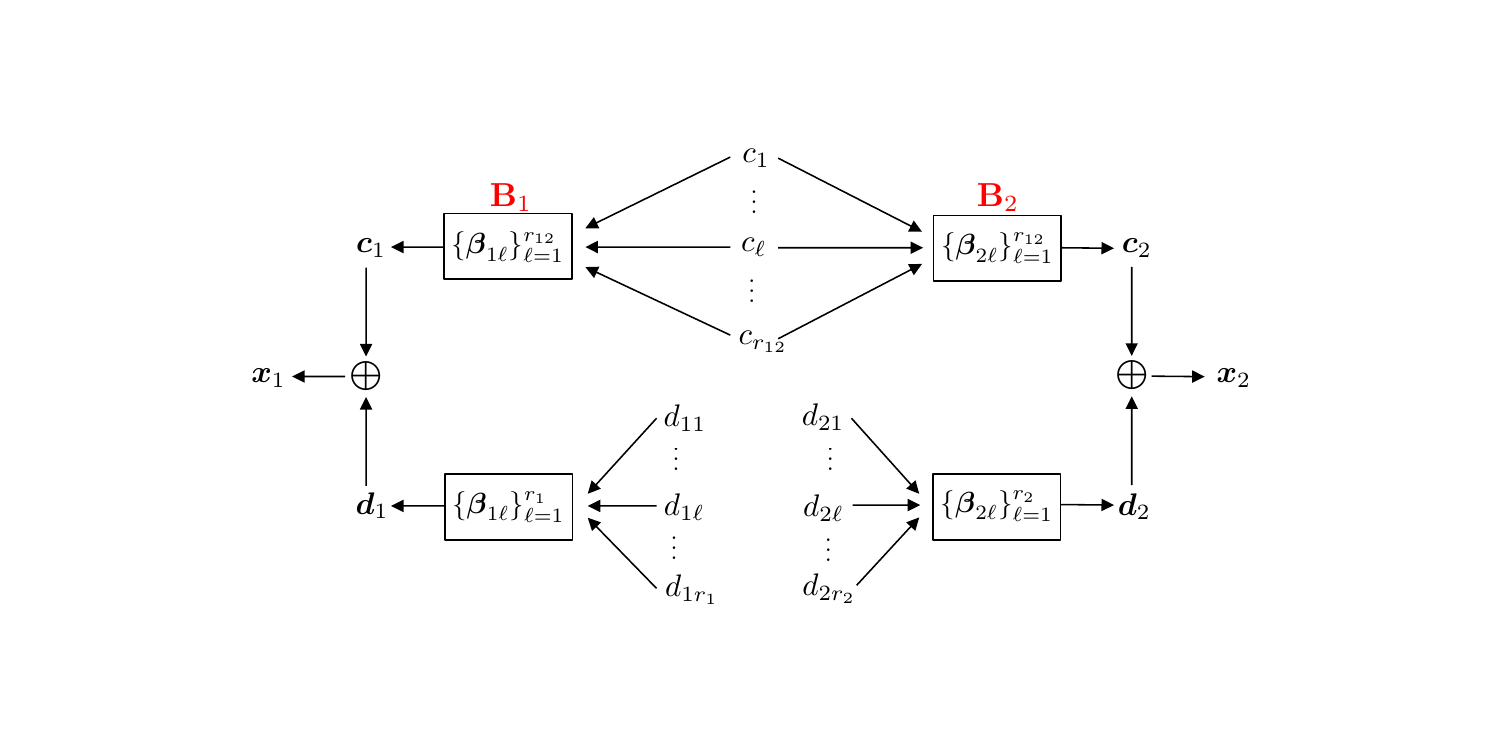}
		\caption{The D-CCA decomposition for two signal random vectors.}
	\end{subfigure}	
	\caption{The D-CCA decomposition structure.
		In subfigure (a), the distinctive variables $d_1$ and $d_2$ are orthogonal (i.e., uncorrelated); the norm (i.e., standard deviation) of the common variable $c$ indicates the correlation strength of the two standardized variables $z_1$ and $z_2$.		
		In subfigure (b), D-CCA refers the common pattern of $\{\bd{x}_1, \bd{x}_2\}$ to the common latent factors $\{c_\ell\}_{\ell=1}^{r_{12}}$, but ignores the common pattern of their coefficient
		matrices $\mb{B}_k = (\bd{\beta}_{k1},\dots,\bd{\beta}_{kr_{12}})$ for $k = 1,2$.}
	\label{D-CCA general struc}
\end{figure}

To this end, guided by the bi-orthogonality \eqref{bi-orthogonality}
of augmented canonical variables $\bd{z}_k^\top=[z_{k\ell}]_{\ell=1}^{r_k}$, $k=1,2$, D-CCA divides the decomposition problem~\eqref{x=c+d} of $\lspan([\bd{x}_1;\bd{x}_2]^\top)$ into $r_{\max}$ subproblems, 
each within one of
the mutually orthogonal subspaces
$\big\{\lspan(\{z_{k\ell}\}_{k=1}^2)\big\}_{\ell=1}^{r_{\max}}$ as
\be\label{z=c+d}
z_{k\ell}=c_\ell+d_{k\ell},
\ee
where $z_{k\ell}=0$ for $\ell>r_k$, and $c_\ell=0$ for $\ell>r_{\min}$.
For $\ell\le r_{\min}$, the common variable $c_\ell$ is defined by
\be\label{angle constraint} 
c_\ell \propto \underset{w\in (\mathcal{L}_0^2,\cov)}{\argmax}\left\{ \cos^2\theta(z_{1\ell},w)+\cos^2\theta(z_{2\ell},w) \right\}  
\ee
such that
\begin{numcases}{}
	d_{1\ell}\perp d_{2\ell},\label{D orth}\\
	\|c_\ell\| \ \text{increases as}\  \theta_{z\ell}:=\theta(z_{1\ell},z_{2\ell})\  \text{decreases on} \ [0,\pi/2]. \label{norm constraint}
\end{numcases}
Constraint~\eqref{norm constraint} equivalently
says that $\|c_\ell\|$ indicates the correlation strength of 
$z_{1\ell}$ and $z_{2\ell}$.
The unique solution of  \eqref{angle constraint}  subject to \eqref{D orth} and \eqref{norm constraint} is 
\be\label{C for K=2}
c_\ell=\left(1-\sqrt{\frac{1-\cos\theta_{z\ell}}{1+\cos\theta_{z\ell}}     }\right) \frac{z_{1\ell}+z_{2\ell}}{2}
=\left[1-\tan\left(\frac{\theta_{z\ell}}{2}\right)\right] \frac{z_{1\ell}+z_{2\ell}}{2}.
\ee
Figure~\ref{D-CCA general struc} (a) geometrically illustrates the solution~\eqref{C for K=2} with $\ell$ omitted in the subscriptions.

Combining the solutions of subproblems yields the D-CCA decomposition:
f{\tiny }or $k=1,2$,
\be\label{X=betaC+betaD}
\bd{x}_k=
\sum_{\ell=1}^{r_k}\bd{\beta}_{k\ell}z_{k\ell}
=\sum_{\ell=1}^{r_{12}}\bd{\beta}_{k\ell}c_\ell+\sum_{\ell=1}^{r_k}\bd{\beta}_{k\ell}d_{k\ell}=: \bd{c}_k+ \bd{d}_k
\ee
with
$\bd{\beta}_{k\ell}=\cov(\bd{x}_k,z_{k\ell})$. Here,
$\{c_\ell\}_{\ell=1}^{r_{12}}$ are the {\it common latent factors} of $\bd{x}_1$ and $\bd{x}_2$, and $\{d_{k\ell}\}_{\ell=1}^{r_k}$ are the {\it distinctive latent factors} of $\bd{x}_k$.
Figure~\ref{D-CCA general struc}\,(b) shows the decomposition structure of D-CCA.

\section{Common and Distinctive Pattern Analysis}\label{CDPA sec: method}

The CDPA method aims to more comprehensively define the common and distinctive patterns of two datasets by incorporating the common and distinctive patterns of the two coefficient matrices of common latent factors. We use a graph matching approach to match the unpaired rows between the coefficient matrices. Consistent estimators are established for the CDPA-defined common-pattern and distinctive-pattern matrices.

\subsection{Common and distinctive patterns}\label{subsec: com and dis}
As shown in Figure~\ref{D-CCA general struc} (b),  
D-CCA only focuses on the common latent factors $\{c_\ell\}_{\ell=1}^{r_{12}}$
of $\{\bd{x}_k\}_{k=1}^2$, and ignores
the common pattern of their
coefficient matrices $\mb{B}_k=(\bd{\beta}_{k1},\dots,\bd{\beta}_{kr_{12}})$ for $k=1,2$. So do its five previous methods mentioned in Section~\ref{sec: intro}.
In factor analysis \citep{Harm76},
the coefficient matrix $\mb{B}_k$
of latent factors $\{c_\ell\}_{\ell=1}^{r_{12}}$
in $\bd{c}_k=\mb{B}_k([c_\ell]_{\ell=1}^{r_{12}})^\top$
is called
their factor pattern matrix,
and the entry
$\mb{B}_k^{[i_k,\ell]}$ is the factor loading on
$c_\ell$ for  variable $\bd{c}_k^{[i_k]}=\sum_{\ell=1}^{r_{12}}\mb{B}_k^{[i_k,\ell]}c_\ell$,
representing the contribution of $c_\ell$
in the linear combination of $\{c_\ell\}_{\ell=1}^{r_{12}}$ to forming $\bd{c}_k^{[i_k]}$.
In signal processing, $\mb{B}_k$ is called a mixing channel \citep{Papa00, Parr03}, 
which introduces correlations into the uncorrelated input sources $\{c_\ell\}_{\ell=1}^{r_{12}}$ to generate the output signal $\bd{c}_k $ that
has $\cov(\bd{c}_k)=\mb{B}_k\diag(\var(c_1),\dots,\var(c_{r_{12}}))\mb{B}_k^\top$. Thus, $\mb{B}_1$ and $\mb{B}_2$ may possess 
common and distinctive patterns of the respective ways in which
the common latent factors $\{c_\ell\}_{\ell=1}^{r_{12}}$ 
constitute
$\bd{c}_1$ and $\bd{c}_2$.
In CDPA,
we define a common-pattern vector $\bd{c}$ for $\{\bd{x}_k\}_{k=1}^2$ which takes into account both the common latent sources $\{c_\ell\}_{\ell=1}^{r_{12}}$ and
the common pattern of their mixing channels $\{\mb{B}_k\}_{k=1}^2$. The distinctive-pattern vector of signal $\bd{x}_k$ is then defined as the residual part of the signal after removing~$\bd{c}$.

In the process $\bd{c}_k=\mb{B}_k([c_\ell]_{\ell=1}^{r_{12}})^\top=\sum_{\ell=1}^{r_{12}}\bd{\beta}_{k\ell} c_\ell$, the $\ell$-th column $\bd{\beta}_{k\ell}$ of the mixing channel $\mb{B}_k$ is the sub-channel transmitting $c_\ell$, and the linear mixture of sub-channel outputs $\{\bd{\beta}_{k\ell} c_\ell\}_{\ell=1}^{r_{12}}$ reflects the ``mixing" performance of the channel $\mb{B}_k$.
We disentangle the common and distinctive latent structures for the two sub-channel spaces $\{\colsp(\mb{B}_k)\}_{k=1}^2$ in a similar way as D-CCA does for the two signal spaces $\{\lspan(\bd{x}_k^\top)\}_{k=1}^2$.


Two issues need to be solved before the analysis.
First, the sub-channel vectors $\{\bd{\beta}_{k\ell}\}_{k\le 2, \ell\le r_{12}}$ may have unequal lengths $p_1$ and $p_2$,
Without loss of generality, we let $p_1\ge p_2$ throughout the paper.
When $p_1>p_2$,
we zero-pad $\mb{B}_2$ to be a $p_1\times r_{12}$ matrix $\mb{B}_{2A}=[\mb{B}_2;\mb{0}_{(p_1-p_2)\times r_{12}}]$.
This zero padding is equivalent to adding $(p_1-p_2)$ zeros into $\bd{x}_2$.
In other words, we are now equivalently studying the patterns between $\bd{x}_1$ and $\bd{x}_{2A}=[\bd{x}_2;\bd{0}_{(p_1-p_2)\times 1}]$.
Second, sometimes the rows between $\mb{B}_1$ and $\mb{B}_{2A}$ or equivalently the entries between $\bd{x}_1$ and $\bd{x}_{2A}$ are not one-to-one matched due to their arbitrary ordering. For this scenario, we match their rows by permuting the rows of $\mb{B}_{2A}$ with a permutation matrix $\mb{P}$. 
The permutation can be defined so that  
$\colsp(\mb{B}_1)$ and $\colsp(\mb{P}\mb{B}_{2A})$ are closest to each other by maximizing
$\sum_{\ell=1}^{r_{12}} \cos^2 \theta_{B\ell}$,
where $\theta_{B\ell}$ is their $\ell$-th principal angle. 
This row-matching procedure will be discussed in detail in Section~\ref{sec: graph matching}. 
For the generalization of our results to other row-matching criteria, we
assume that the permutation matrix $\mb{P}$ is prespecified in the following text.
For notational simplicity,
we use 
the superscript ``$\diamond$" to indicate adding zero padding and $\mb{P}$ to the given vector or matrix if necessary, for example,
$(\bd{x}_1^\diamond,\bd{x}_2^\diamond, \mb{B}_1^\diamond, \mb{B}_2^\diamond)=(\bd{x}_1, \mb{P}\bd{x}_{2A}, \mb{B}_1, \mb{P}\mb{B}_{2A})$.

We now consider the latent structure of
the two sub-channel spaces $\{\colsp(\mb{B}_k^\diamond)\}_{k=1}^2$ by using an analogy of D-CCA on $(\mathbb{R}^{p_1},\cdot)$,
where constraints \eqref{x=c+d}-\eqref{norm constraint} are translated for
the columns of $\{\mb{B}_k^\diamond\}_{k=1}^2$ and CCA is replaced by PAA. 
Let $\theta_{B\ell}$ and $\{\bd{v}_{B_1\ell},\bd{v}_{B_2\ell}\}$ be the $\ell$-th principal angle and the $\ell$-th pair of principal vectors of $\{\colsp(\mb{B}_k^\diamond)\}_{k=1}^2$. There are $r_{12}$ such pairs since $\mb{B}_k=\mb{V}_k\mb{\Lambda}_k^{1/2}\mb{U}_{\theta k}^{[:,1:r_{12}]}$ is a rank-$r_{12}$ matrix.
We define the common and distinctive components of $\{\bd{v}_{B_1\ell},\bd{v}_{B_2\ell}\}$ using a decomposition similar to that in 
\eqref{z=c+d} and \eqref{C for K=2}:
\be\label{c_{B,ell} vector}
\bd{c}_{B\ell}=
\left(1-\sqrt{\frac{1-\cos\theta_{B\ell}}{1+\cos\theta_{B\ell}}     }\right) \frac{(\bd{v}_{B_1\ell}+\bd{v}_{B_2\ell})}{2}
\quad\text{and}\quad
\bd{d}_{B_k\ell}=\bd{v}_{B_k\ell}-\bd{c}_{B\ell}
\ee
for $k=1,2$ and $\ell=1,\dots, r_{12}$.
Because the principal vectors 
$(\bd{v}_{B_k1},\dots,\bd{v}_{B_kr_{12}})=:\mb{V}_{B_k}$ form an  orthonormal basis of $\colsp(\mb{B}_k^\diamond)$, 
the mixing-channel matrix can be written as
\be\label{B1 decomp}
\mb{B}_k^\diamond= \mb{V}_{B_k} (\mb{V}_{B_k}^\top\mb{B}_k^\diamond)=   \big([\bd{c}_{B\ell}]_{\ell=1}^{r_{12}}+
[\bd{d}_{B_k\ell}]_{\ell=1}^{r_{12}}\big) (\mb{V}_{B_k}^\top\mb{B}_k^\diamond).
\ee
The part of $\bd{x}_k^\diamond$ that
contains the  common latent factors (source variables) $\{c_\ell\}_{\ell=1}^{r_{12}}$
and the common mixing-channel basis $\{\bd{c}_{B\ell}\}_{\ell=1}^{r_{12}}$
is
\be\label{c1*}
\bd{c}_k^*:=[\bd{c}_{B\ell}]_{\ell=1}^{r_{12}}\mb{V}_{B_k}^\top \mb{B}_k^\diamond ([c_\ell]_{\ell=1}^{r_{12}})^\top.
\ee
The difference between $\bd{c}_1^*$ and $\bd{c}_2^*$ is the matrices
$
\mb{S}_k:=
\mb{V}_{B_1}^\top \mb{B}_k^\diamond
$, $k=1,2$
in the middle of their formulas, which contain the weights dually owned by
$\{\bd{c}_{B\ell}\}_{\ell=1}^{r_{12}}$
and 
$\{c_\ell\}_{\ell=1}^{r_{12}}$.
We define the common part of the two dual weight matrices $\{\mb{S}_k\}_{k=1}^2$ as
\be\label{S matrix}
\mb{S}
=\argmin_{\mb{M}\in \mathbb{R}^{r_{12}\times r_{12}}}
\sum_{k=1}^2\Big\|\mb{M}-[\tr(\mb{\Sigma}_k)]^{-1/2}\mb{S}_k \Big\|_F^2
=\frac{1}{2}\sum_{k=1}^2[\tr(\mb{\Sigma}_k)]^{-1/2}\mb{S}_k.
\ee
To avoid overweighting a dataset when signals 
$\bd{x}_1$ and $\bd{x}_2$ have different scales,
we weight $\mb{S}_k$ by
the scale factor $[\tr(\mb{\Sigma}_k)]^{-1/2}$ in \eqref{S matrix}.
This is equivalent to rescaling each $\bd{x}_k$ by the
factor $[\tr(\mb{\Sigma}_k)]^{-1/2}$ at the very beginning as in \citep{Lock13}.
Our definition of $\mb{S}$ in \eqref{S matrix}
is motivated by the {\it consensus configuration}
in generalized procrustes analysis \citep{Gowe75,Gowe04}
which minimizes the sum of squared Euclidean distances
to transformed configurations of interest (i.e., the scaled $\{\mb{S}_k\}_{k=1}^2$ in our case).
This minimization is equivalent to that of the sum of Kullback-Leibler divergences \citep[Lemma 17.4.3]{Moak06} and yields a closed-form solution.

We combine the three types of common parts $\{\bd{c}_{B\ell}\}_{\ell=1}^{r_{12}}$, $\{c_\ell\}_{\ell=1}^{r_{12}}$
and $\mb{S}$
to define the {\it common-pattern vector} of the scaled signals $\bd{x}_k^S:=[\tr(\mb{\Sigma}_k)]^{-1/2}\bd{x}_k^\diamond$, $k=1,2$ as
\be\label{c vector}
\bd{c} 
=\mb{B}_c([c_\ell]_{\ell=1}^{r_{12}})^\top
=[\bd{c}_{B\ell}]_{\ell=1}^{r_{12}}\mb{S}([c_\ell]_{\ell=1}^{r_{12}})^\top
=
\frac{1}{2}\sum_{k=1}^2[\tr(\mb{\Sigma}_k) ]^{-1/2}\bd{c}_k^*,
\ee
where $\mb{B}_c=[\bd{c}_{B\ell}]_{\ell=1}^{r_{12}}\mb{S}$
is defined as the common pattern of $\mb{B}_1^\diamond$ and $\mb{B}_2^\diamond$.

For each individiual unscaled signal vector $\bd{x}_k^\diamond$,
we rescale $\bd{c}$ to be 
$\bd{c}^{(k)}=[\tr(\mb{\Sigma}_k)]^{1/2}\bd{c}$ 
and express the CDPA decomposition as
\be\label{CDPA decomp}
\bd{x}_k^\diamond=\bd{c}_k^\diamond+\bd{d}_k^\diamond=:(\bd{c}^{(k)}+\bd{h}_k)+\bd{d}_k^\diamond
=\bd{c}^{(k)}+(\bd{h}_k+\bd{d}_k^\diamond)=:\bd{c}^{(k)}+\bd{\delta}_k.
\ee
For signal vector  $\bd{x}_k^\diamond$, the vector $\bd{h}_k$ represents the distinctive pattern retained within the common-source vector $\bd{c}_k^\diamond$,
and the vector $\bd{\delta}_k$ characterizes  the {\it total} distinctive pattern by incorporating both $\bd{h}_k$ and the distinctive-source vector $\bd{d}_k^\diamond$.
We denote $\{\mb{C},\mb{C}^{(k)},\mb{H}_k,\mb{\Delta}_k\}$ to be
the corresponding sample matrices of $\{\bd{c},\bd{c}^{(k)},\bd{h}_k,\bd{\delta}_k\}$ associated with $\mb{X}_k$.

\begin{definition}\label{def: C, Delta}
	We define  the common-pattern vector of $\{\bd{x}_1^\diamond,\bd{x}_2^\diamond\}$ (more precisely, $\{\bd{x}_1^S,\bd{x}_2^S\}$) as the vector $\bd{c}$ given in \eqref{c vector}, 
	and the {\it scaled} common-pattern vector for $\bd{x}_k^\diamond$ as $\bd{c}^{(k)}=[\tr(\mb{\Sigma}_k)]^{1/2}\bd{c}$.  The
	distinctive-pattern vector of $\bd{x}_k^\diamond$ is $\bd{\delta}_k=\bd{x}_k^\diamond-\bd{c}^{(k)}$.
	As the sample matrices of $\bd{c}$, $\{\bd{c}^{(k)}\}_{k=1}^2$ and $\{\bd{\delta}_k\}_{k=1}^2$,
	matrices 
	$\mb{C}$, $\{\mb{C}^{(k)}\}_{k=1}^2$ and $\{\mb{\Delta}_k\}_{k=1}^2$
	are called the common-pattern, the scaled common-pattern, and distinctive-pattern matrices of
	$\{\mb{X}_k\}_{k=1}^2$, respectively.
\end{definition}

The population CDPA decomposition is summarized in Algorithm~\ref{Oralce CDPA}
and its uniqueness is given in Theorem~\ref{thm: uniq}.

\begin{algorithm}
	\caption{Population CDPA}
	\label{Oralce CDPA}
	\begin{algorithmic}[1]
		\Require Signal vectors $\bd{x}_k\in \mathbb{R}^{p_k}, k=1,2$

		\State Obtain common latent factors $[c_\ell]_{\ell=1}^{r_{12}}$ and coefficient matrix
		$\mb{B}_k$ by the D-CCA in \eqref{X=betaC+betaD};

		\State Add zero rows to $\mb{B}_1$ or $\mb{B}_2$ if dimensions $p_1\ne p_2$;

		\State Match the rows of $\mb{B}_1$ and $\mb{B}_2$ by the graph-matching based approach (Section~\ref{sec: graph matching}),		
		if the variables in $\bd{x}_1$ and $\bd{x}_2$ are not paired;

		\State Compute the common mixing-channel basis $\{\bd{c}_{B\ell}\}_{\ell=1}^{r_{12}}$ by \eqref{c_{B,ell} vector} for $\{\colsp(\mb{B}_k)\}_{k=1}^2$;

		\State	Compute the common dual-weight matrix $\mb{S}$ by \eqref{S matrix};

		\State Obtain
		the common-pattern vector of $\bd{x}_1$ and $\bd{x}_2$ as
		$
		\bd{c}=[\bd{c}_{B\ell}]_{\ell=1}^{r_{12}}\mb{S}([c_\ell]_{\ell=1}^{r_{12}})^\top;
		$

		\State Rescale $\bd{c}$ to be
		the scaled common-pattern vector
		$\bd{c}^{(k)}=[\tr(\mb{\Sigma}_k)]^{1/2}\bd{c}$;

		\State Obtain the distinctive-pattern vector of $\bd{x}_k$ as 
		$\bd{\delta}_k=\bd{x}_k-\bd{c}^{(k)}$;


		\Ensure	Common-pattern vector $\bd{c}$, scaled common-pattern vectors $\{\bd{c}^{(k)}\}_{k=1}^2$,  distinctive-pattern vectors $\{\bd{\delta}_k\}_{k=1}^2$
		
	\end{algorithmic}
\end{algorithm}

\begin{theorem}\label{thm: uniq}
	Given any $p_1\times p_1$ permutation matrix $\mb{P}$, the common-pattern vector $\bd{c}$ defined in \eqref{c vector} for $(\bd{x}_1^\diamond, \bd{x}_2^\diamond)=(\bd{x}_1,\mb{P}\bd{x}_{2A})$ is unique, regardless of  the non-unique choices of canonical variables $\{z_{1\ell},z_{2\ell}\}_{\ell=1}^{r_{12}}$	and principal vectors $\{\bd{v}_{B_1\ell},\bd{v}_{B_2\ell}\}_{\ell=1}^{r_{12}}$.
\end{theorem}

\begin{remark}
	Since $\bd{c}$ is the common-pattern vector of the scaled signal vectors $\bd{x}_1^S$ and $\bd{x}_2^S$,
	$\tr\{\cov(\bd{c})\}=\frac{\tr\{\cov(\bd{c})\}}{\frac{1}{2}\sum_{k=1}^2 \tr\{\cov(\bd{x}_k^S)\}}$
	represents the
	proportion of the average variance of $\bd{x}_1^S$ and $\bd{x}_2^S$ explained 
	by $\bd{c}$, which reflects the similarity strength of the two signal vectors.
\end{remark}

\begin{remark}
	The common-pattern vector $\bd{c}$ differs only in its sign
	for 
	$\{\bd{x}_1,\mb{P}\bd{x}_{2A}\}$ and $\{-\bd{x}_1,$
	$-\mb{P}\bd{x}_{2A}\}$, but
	is usually quite different for $\{\bd{x}_1,\mb{P}\bd{x}_{2A}\}$ and $\{\bd{x}_1,-\mb{P}\bd{x}_{2A}\}$. 
	We assume the sign of each entry in $\bd{y}_k$ or $\bd{x}_k$ cannot be arbitrarily changed, but the sign of $\bd{y}_k$ or equivalently that of $\bd{x}_k$ may change.
	The assumption is generally true if each dataset represents a data type.
	For example, let $\bd{y}_2$ be mRNA expression data and its entry $\bd{y}_2^{[i]}$ measure the mRNA expression level on the $i$-th gene. The arbitrary entry-wise sign changes can result in two different measurements applied to $\bd{y}_2$. 
	Regarding the different $\bd{c}$'s due to the sign change (if allowed) of entirely $\bd{y}_2$ or $\bd{x}_2$,
	we suggest to choose the one with larger variance $\tr\{\cov(\bd{c})\}$
	or, in practice, larger $\frac{1}{n}\|\widehat{\mb{C}} \|_F^2=\tr(\frac{1}{n}\widehat{\mb{C}}\widehat{\mb{C}}^\top)$, where $\widehat{\mb{C}}$ is the estimate of $\mb{C}$ that will be introduced in Section~\ref{sec: estimation}.
	It will be shown later in Theorem~\ref{thm: CC} that
	$\frac{1}{n}\|\widehat{\mb{C}} \|_F^2\overset{P}{\to}\tr\{\cov(\bd{c})\}$
	under mild conditions.
	The confidence interval (CI) of  $\frac{1}{n}\|\widehat{\mb{C}} \|_F^2$ can be constructed by bootstrapping samples \citep{Efro93} once the ranks $\{r_1,r_2,r_{12}\}$ and the permutation matrix $\mb{P}$ are determined.
\end{remark}

\subsection{Row matching of coefficient matrices}\label{sec: graph matching}

When the rows of coefficient matrices $\mb{B}_1$ and $\mb{B}_{2A}$ are not one-to-one matched, we match them by permuting the rows of $\mb{B}_{2A}$ with the following permutation matrix 
\be\label{P obj in cos theta}
\mb{P}_*=\argmax_{\mb{P}\in \Pi_{p_1}}\sum_{\ell=1}^{r_{12}} \cos^2 \theta_{B\ell},
\ee
where $\theta_{B\ell}$ is the $\ell$-th principal angle of 
$\colsp(\mb{B}_1)$ and $\colsp(\mb{P}\mb{B}_{2A})$, and $\Pi_{p_1}$ is the set of all $p_1\times p_1$ permutation matrices.
This optimization is equivalent to minimizing the Frobenius distance
$(2\sum_{\ell=1}^{r_{12}}\sin^2\theta_{B\ell})^{1/2}$.
Commonly-used distances between vector spaces \citep{Deza14}
also include the geodesic distance 
$(\sum_{\ell=1}^{r_{12}} \theta_{B\ell}^2)^{1/2}$,
the Martin distance
$(\log\prod_{\ell=1}^{r_{12}} \cos^{-2}\theta_{B\ell})^{1/2}$,
the Asimov distance $\theta_{B1}$, and the gap distance $\sin \theta_{B1}$.
The four alternative distances appear more difficult to be minimized,
but our criterion based on the Frobenius distance
can be converted to the famous graph matching problem \citep{Lu16}. 

Specifically, 
by equations in \eqref{PA,PV},
the optimization problem in \eqref{P obj in cos theta} is equivalent to
\begin{align}
\mb{P}_*&=\argmax_{\mb{P}\in \Pi_{p_1}} \tr\left(\mb{Q}_1^\top \mb{P} \mb{Q}_{2A} (\mb{Q}_1^\top \mb{P} \mb{Q}_{2A})^\top\right)\nonumber\\
&= \argmax_{\mb{P}\in \Pi_{p_1}} \tr\left(
\mb{Q}_1\mb{Q}_1^\top \mb{P} \mb{Q}_{2A}\mb{Q}_{2A}^\top \mb{P}^\top
\right),\label{B's opt obj}
\end{align}
where $\mb{Q}_k\in \mathbb{R}^{p_k\times r_{12}}$ is a matrix whose columns are an orthonormal basis of $\colsp(\mb{B}_k)$, which can be the $r_{12}$ left singular vectors of $\mb{B}_k$,
and $\mb{Q}_{2A}=[\mb{Q}_2;\mb{0}_{(p_1-p_2)\times r_{12}}]$ whose columns are still an orthonormal basis of $\mb{B}_{2A}$.
Let $\mb{M}_1=\mb{Q}_1\mb{Q}_1^\top$ and $\mb{M}_2=\mb{Q}_{2A}\mb{Q}_{2A}^\top$.
For $k=1,2$,
let $\mb{M}_k^+$ be the matrix obtained by all elements of $\mb{M}_k$ minus the smallest element of $[\mb{M}_1,\mb{M}_2]$.
For any $p_1\times p_1$ matrix $\mb{M}$,
denote $\overline{\diag}(\mb{M})$ to be the
$p_1\times p_1$ matrix having the same off-diagonal part of
$\mb{M}$ but with zero diagonal, 
and $\vdg(\mb{M})$ to be the vector consisting of 
the diagonal elements of $\mb{M}$. 
We have
\begin{align}
	\lefteqn{\max_{\mb{P}\in \Pi_{p_1}} \tr\left(
\mb{M}_1 \mb{P} \mb{M}_2 \mb{P}^\top
\right)}\nonumber\\
&\Leftrightarrow
\min_{\mb{P}\in \Pi_{p_1}}\left\|\mb{M}_1-\mb{P}\mb{M}_2 \mb{P}^\top  \right\|_F^2\nonumber\\
&\Leftrightarrow
\min_{\mb{P}\in \Pi_{p_1}}\left\|\mb{M}_1^+-\mb{P}\mb{M}_2^+ \mb{P}^\top  \right\|_F^2  \nonumber\\
&\Leftrightarrow
\min_{\mb{P}\in \Pi_{p_1}}\left\{\left\|\overline{\diag}(\mb{M}_1^+)-\mb{P}\overline{\diag}(\mb{M}_2^+ )\mb{P}^\top  \right\|_F^2 
+\left\| \vdg(\mb{M}_1^+)-\mb{P}\vdg(\mb{M}_2^+)   \right\|_F^2
\right\}
\nonumber\\
&\Leftrightarrow
\max_{\mb{P}\in \Pi_{p_1}} \left\{
\tr\left(
\mb{P}^\top\overline{\diag}(\mb{M}_1^+) \mb{P} \overline{\diag}(\mb{M}_2^+)
\right)
+\tr\left( \mb{P}^\top \vdg(\mb{M}_1^+)[\vdg(\mb{M}_2^+)]^\top \right)
\right\},\label{equal to graph match}
\end{align}
where 
the last objective function is the formula (4) of \citep{Lu16} for the graph matching problem.
Graph matching is known to be NP-hard for the optimal solution.
We use the doubly stochastic projected fixed-point (DSPFP) algorithm of \citep{Lu16} to obtain an efficient approximation of $\mb{P}_*$,
which has time complexity only $O(p_1^3)$ per iteration and space complexity $O(p_1^2)$. 
For ultra-large $p_1$, one may further apply the approximation procedure of \citep{Oliv16} that employs a clustering method before DSPFP.

\subsection{Estimation}\label{sec: estimation}
Often in practice, the data matrices $\{\mb{Y}_k\}_{k=1}^2$ are high-dimensional and are the only observable data in decomposition \eqref{decomp mat}. 
The literature of \eqref{decomp mat} regularly assumes high-dimensional $\{\mb{Y}_k\}_{k=1}^2$ to be ``low-rank plus noise". Indeed, big data matrices are often approximately low-rank in many real-world applications \citep{Udel19}, so their low-rank approximations provide feasible or more efficient computation and meanwhile preserve the major information \citep{Kish17}. Moreover, the low-rank plus noise structure can circumvent
the curse of dimensionality \citep{Yin88,Kolt17} in recovering the common-source and distinctive-source matrices $\{\mb{C}_k, \mb{D}_k\}_{k=1}^2$ from which our defined common-pattern and distinctive-pattern matrices are derived.
Following the D-CCA paper \citep{Shu18},
we consider the low-rank plus noise structure:
\begin{align}
	\mb{Y}_k&=\mb{X}_k+\mb{E}_k=\mb{B}_{f_k}\mb{F}_k+\mb{E}_k,
	\label{matrix AFM}\\
	\bd{y}_k&=\bd{x}_k+\bd{e}_k=\mb{B}_{f_k}\bd{f}_k+\bd{e}_k,
	\label{variable AFM}
\end{align}
where $\mb{B}_{f_k}\in \mathbb R^{p_k\times r_k}$ is a real deterministic matrix,
the columns of $\mb{F}_k$ and $\mb{E}_k$ are respectively the $n$ i.i.d. copies of mean-zero random vectors $\bd{f}_k$ and $\bd{e}_k$,
the columns of $\mb{F}=[\mb{F}_1;\mb{F}_2]$ are also statistically independent,
and the vector $\bd{f}_k\in \mathbb{R}^{r_k}$
contains $r_k$ latent factors such that $\cov(\bd{f}_k)=\mb{I}_{r_k\times r_k}$, $\cov(\bd{f}_k,\bd{e}_k)=\mb{0}_{r_k\times p_k}$, 
and $\lspan(\bd{f}_k^\top)$ is a fixed subspace
in $(\mathcal{L}_0^2,\cov)$ that is independent of $\{n,p_1,p_2\}$.
Hence, $r_1$, $r_2$ and $r_{12}$ are fixed numbers.
We can choose $\bd{f}_k^\top$ to be the augmented canonical variables $\bd{z}_k^\top$.
The covariance matrix $\cov(\bd{y}_k)=\mb{B}_{f_k}\mb{B}_{f_k}^\top +\cov(\bd{e}_k)$ is assumed to be a spiked covariance matrix for which the largest $r_k$ eigenvalues are significantly larger than the rest, namely, signals are distinguishably stronger than noises.

Before recovering our common-pattern and distinctive-pattern matrices,
we introduce the
D-CCA's estimators of $\mb{X}_k$ and $\mb{C}_k$.
For simplicity, 
we write all estimators with true matrix ranks $\{r_1,r_2,r_{12}\}$.
In practice, as implemented in D-CCA,
ranks $\{r_k\}_{k=1}^2$ and $r_{12}$ can be well selected by 
the edge distribution (ED) method of \citep{Onat10} and
the minimum description length information-theoretic criterion (MDL-IC)
of \citep{Song16}, respectively; 
see Appendix~\ref{sec: rank selection}. 
The estimator of $\mb{X}_k$ is defined by using 
the soft-thresholding method of \citep{Wang17} as
\be\label{X tilde}
\widehat{\mb{X}}_k=\mb{U}_{k1} \diag(\hat{\sigma}_1^S(\mb{Y}_k),\dots, \hat{\sigma}_{r_k}^S(\mb{Y}_k))\mb{U}_{k2}^\top,
\ee
where
$\mb{U}_{k1}\diag(\sigma_1(\mb{Y}_k),\dots,\sigma_{r_k}(\mb{Y}_k))\mb{U}_{k2}^\top$ forms the top-$r_k$ SVD of $\mb{Y}_k$,
and the soft-thresholded singular value
$
\hat{\sigma}_\ell^S(\mb{Y}_k)=\sqrt{\max\{\sigma_\ell^2(\mb{Y}_k)- \tau_kp_k,0\}}
$ with
$
\tau_k=\sum_{\ell=r_k+1}^{p_k} \sigma_\ell^2(\mb{Y}_k)/(np_k-nr_k-p_kr_k).
$ 
Then from $\widehat{\mb{X}}_k$,
define the estimator of $\mb{\Sigma}_k$ by
$\widehat{\mb{\Sigma}}_k=\frac{1}{n}\widehat{\mb{X}}_k\widehat{\mb{X}}_k^\top$,
and denote its SVD by 
$\widehat{\mb{\Sigma}}_k=\widehat{\mb{V}}_k\widehat{\mb{\Lambda}}_k\widehat{\mb{V}}_k^\top$,
where $\widehat{\mb{V}}_k\in \mathbb{R}^{p_k\times r_k}$ has orthonormal columns
and
$\widehat{\mb{\Lambda}}_k=\diag(\sigma_1(\widehat{\mb{\Sigma}}_k),\dots,\sigma_{r_k}(\widehat{\mb{\Sigma}}_k))$.
Following Section~\ref{subsec: CCA},
let $\widehat{\mb{Z}}_k^*=(\widehat{\mb{\Lambda}}_k^\dag)^{1/2}\widehat{\mb{V}}_k^\top \widehat{\mb{X}}_k$ and
$\widehat{\mb{\Theta}}=\frac{1}{n}\widehat{\mb{Z}}_1^*(\widehat{\mb{Z}}_2^*)^\top$, and write
the latter's full SVD by
$\widehat{\mb{\Theta}}=\widehat{\mb{U}}_{\theta 1}\widehat{\mb{\Lambda}}_\theta \widehat{\mb{U}}_{\theta 2}^\top$
with
$\widehat{\mb{\Lambda}}_\theta=\diag(\sigma_1(\widehat{\mb{\Theta}}),\dots,\sigma_{\hat{r}_{\theta}}(\widehat{\mb{\Theta}}),\mb{0}_{(r_1-\hat{r}_{\theta})\times (r_2-\hat{r}_{\theta})})$ and $\hat{r}_{\theta}=\rank(\widehat{\mb{\Theta}})$.
Define the estimated sample matrix of $\bd{z}_k$
by 
$\widehat{\mb{Z}}_k=\widehat{\mb{U}}_{\theta k}^\top \widehat{\mb{Z}}_k^*$.
Let $\widehat{\mb{A}}_C=\diag(\hat{a}_1,\dots,\hat{a}_{r_{12}})$,
where $\hat{a}_\ell=\frac{1}{2}\big[1-(\frac{1-\sigma_\ell (\widehat{\mb{\Theta}})}{1+\sigma_\ell (\widehat{\mb{\Theta}})})^{1/2}\big]$
for $\ell\le \hat{r}_{\theta}$, and otherwise $\hat{a}_\ell=0$.
The estimators of $\mb{C}_k$ and $\mb{D}_k$ are defined by
\[
\widehat{\mb{C}}_k=\frac{1}{n}
\widehat{\mb{X}}_k(\widehat{\mb{Z}}_k^{[1:r_{12},:]})^\top 
\widehat{\mb{A}}_C\sum_{j=1}^2 \widehat{\mb{Z}}_j^{[1:r_{12},:]}
=\widehat{\mb{B}}_k \widehat{\mb{C}}_0
\quad\text{and}\quad
\widehat{\mb{D}}_k=\widehat{\mb{X}}_k-\widehat{\mb{C}}_k.
\]
where
$\widehat{\mb{B}}_k=\frac{1}{n}\widehat{\mb{X}}_k(\widehat{\mb{Z}}_k^{[1:r_{12},:]})^\top=\widehat{\mb{V}}_k\widehat{\mb{\Lambda}}_k^{1/2}\widehat{\mb{U}}_{\theta k}^{[:,1:r_{12}]}$ similar to
$\mb{B}_k=\mb{V}_k\mb{\Lambda}_k^{1/2}\mb{U}_{\theta k}^{[:,1:r_{12}]}$,
and $\widehat{\mb{C}}_0=\widehat{\mb{A}}_C\sum_{j=1}^2 \widehat{\mb{Z}}_j^{[1:r_{12},:]}$
is the estimated sample matrix of $(c_1,\dots,c_{r_{12}})^\top$.

We now derive the estimators of our common-pattern and distinctive-pattern matrices. 
Let $\widehat{\mb{B}}_{2A}=[\widehat{\mb{B}}_2;\mb{0}_{ (p_1-p_2)\times r_{12}}]$,
$\widehat{\mb{Q}}_k\in \mathbb{R}^{p_k\times r_{12}}$ be the left singular matrix of $\widehat{\mb{B}}_k$,
$\widehat{\mb{Q}}_{2A}=[\widehat{\mb{Q}}_2;\mb{0}_{(p_1-p_2)\times r_{12}}]$, and
$\widehat{\mb{\Theta}}_B=\widehat{\mb{Q}}_1^\top\mb{P} \widehat{\mb{Q}}_{2A}$.
Recall that we assume the permutation matrix $\mb{P}$ is prespecified. 
If the row matching of $\mb{B}_1$ and $\mb{B}_{2A}$ is necessary,
one may choose $\mb{P}$ to be the matrix $\mb{P}_*$ in the NP-hard problem~\eqref{B's opt obj},
approximated by the DSPFP method with data samples. 
Note that $\mb{P}_*$, as a permutation matrix, is either obtained exactly 
or approximated with at least two wrong entries. 
To ease theoretical analysis without such misspecification, we assume that $\mb{P}$ is well determined.
Write the full SVD of $\widehat{\mb{\Theta}}_B$ by
$
\widehat{\mb{\Theta}}_B=\widehat{\mb{U}}_{B_1} \widehat{\mb{\Lambda}}_B\widehat{\mb{U}}_{B_2}^\top
$,
where $\widehat{\mb{\Lambda}}_B$
has nonincreasing diagonal elements,
and define
$\widehat{\mb{V}}_{B_1}=\widehat{\mb{Q}}_1\widehat{\mb{U}}_{B_1}$ 
and $\widehat{\mb{V}}_{B_2}=\mb{P}\widehat{\mb{Q}}_{2A}\widehat{\mb{U}}_{B_2}$.
It follows from \eqref{PA,PV} that 
the diagonal elements of $\widehat{\mb{\Lambda}}_B$ and the
columns of $\{\widehat{\mb{V}}_{B_k}\}_{k=1}^2$ are respectively the cosines of principal angles and the principal vectors of $\colsp(\widehat{\mb{B}}_1)$ and $\colsp(\mb{P}\widehat{\mb{B}}_{2A})$.
Substituting them
for their true counterparts in \eqref{c_{B,ell} vector} yields
our estimator $\hat{\bd{c}}_{B\ell}$ for $\bd{c}_{B\ell}$.
Then from \eqref{c vector}, we define the estimator of $\mb{C}$ by
\be\label{C_hat matrix}
\widehat{\mb{C}}=\frac{1}{2}[\hat{\bd{c}}_{B\ell}]_{\ell=1}^{r_{12}}\Big(\widehat{\mb{V}}_{B_1}^\top\widehat{\mb{B}}_1[\tr(\widehat{\mb{\Sigma}}_1)]^{-1/2}+ \widehat{\mb{V}}_{B_2}^\top \mb{P}  \widehat{\mb{B}}_{2A}[\tr(\widehat{\mb{\Sigma}}_2)]^{-1/2}\Big)\widehat{\mb{C}}_0,
\ee
where $[\tr(\widehat{\mb{\Sigma}}_k)]^{1/2}=[\tr(\frac{1}{n}\widehat{\mb{X}}_k\widehat{\mb{X}}_k^\top)]^{1/2}=\frac{1}{\sqrt{n}}\|\widehat{\mb{X}}_k\|_F$ estimates $[\tr(\mb{\Sigma}_k)]^{1/2}$. The estimator of the scaled version $\mb{C}^{(k)}$ is defined 
by 
\[
\widehat{\mb{C}}^{(k)}=[\tr(\widehat{\mb{\Sigma}}_k)]^{1/2}\widehat{\mb{C}}.
\]
Given $\{r_1,r_2,r_{12},\mb{P}\}$, the computational complexity of obtaining $\widehat{\mb{C}}$ and $\widehat{\mb{C}}^{(k)}$ is $O(np_1^2 \wedge n^2p_1)$ majorly 
due to the SVD of $\{\mb{Y}_k\}_{k=1}^2$.
We define the estimators
$\widehat{\mb{H}}_k=\widehat{\mb{C}}_k^\diamond-\widehat{\mb{C}}^{(k)}$
and $\widehat{\mb{\Delta}}_k=\widehat{\mb{H}}_k+\widehat{\mb{D}}_k^\diamond$
for $\mb{H}_k$ and $\mb{\Delta}_k$, respectively.

The estimation approach for the CDPA decomposition is summarized in Algorithm~\ref{Prac CDPA}.

\begin{algorithm}[h!]
	\caption{CDPA estimation}
	\label{Prac CDPA}
	\begin{algorithmic}[1]
		\Require Observed datasets $\mb{Y}_k\in \mathbb{R}^{p_k\times n}$, $k=1,2$

		\State Select ranks $r_k=\rank(\mb{\Sigma}_k)$ 
		and $r_{12}=\rank(\mb{\Sigma}_{12})$, respectively,
		by the ED method and the MDL-IC method (Appendix~\ref{sec: rank selection}).

		\State Obtain the denoised data $\widehat{\mb{X}}_k$ by the soft thresholding in \eqref{X tilde}.

		\State Obtain coefficient matrix estimates $\{\widehat{\mb{B}}_k\}_{k=1}^2$ 
		and the sample matrix $\widehat{\mb{C}}_0$ of common latent factors
		by the sample D-CCA (Section~\ref{sec: estimation}).

		\State	If necessary, zero-pad and/or row-match $\{\widehat{\mb{B}}_k\}_{k=1}^2$
		by the graph-matching based approach (Section~\ref{sec: graph matching}).

		\State Compute the common-pattern matrix estimate $\widehat{\mb{C}}$ by
		\eqref{C_hat matrix}.

		\State Obtain the scaled common-pattern matrix estimate
		$
		\widehat{\mb{C}}^{(k)}=[\tr(\widehat{\mb{\Sigma}}_k)]^{1/2}\widehat{\mb{C}}
		$
		with $\widehat{\mb{\Sigma}}_k=\widehat{\mb{X}}_k\widehat{\mb{X}}_k^\top/n$,
		and the distinctive-pattern matrix estimate
		$
		\widehat{\mb{\Delta}}_k=\widehat{\mb{X}}-\widehat{\mb{C}}^{(k)}.
		$

		\Ensure	Common-pattern matrix estimate $\widehat{\mb{C}}$, scaled common-pattern matrix estimates $\{\widehat{\mb{C}}^{(k)}\}_{k=1}^2$,  distinctive-pattern matrix estimates $\{\widehat{\mb{\Delta}}_k\}_{k=1}^2$		
	\end{algorithmic}
\end{algorithm}

The following assumption 
given in 
\citep{Wang17,Shu18}, which  guarantees the consistency of $\{\widehat{\mb{X}}_k\}_{k=1}^2$, is also used to derive our asymptotic results. 
Readers are referred to \citep{Wang17,Shu18} for detailed discussions on this assumption.

\begin{assump}\label{assump1}	
	We assume the following conditions for model given in~\eqref{matrix AFM} and~\eqref{variable AFM}.
	\begin{enumerate}[(I)]
		\item Let $\lambda_{k,1}>\cdots>\lambda_{k,r_k}>\lambda_{k,r_k+1}\ge \cdots\ge \lambda_{k,p_k}>0$ be
		the eigenvalues of $\cov(\bd{y}_k)$. There exist positive constants $\kappa_1,\kappa_2$ and $\delta_0$ such that $\kappa_1\le \lambda_{k,\ell}\le \kappa_2$ for $\ell>r_k$
		and  $\min_{\ell\le r_k} (\lambda_{k,\ell}-\lambda_{k,\ell+1})/\lambda_{k,\ell}\ge \delta_0$.
		
		\item Assume that $p_k>\kappa_0 n$ with a constant $\kappa_0>0$. When $n\to \infty$, assume $\lambda_{k,r_k}\to \infty$,
		$p_k/(n\lambda_{k,\ell})$ is upper bounded for $\ell\le r_k$, $\lambda_{k,1}/\lambda_{k,r_k}$ is bounded from above and below,
		and $\sqrt{p_k}(\log n)^{1/\gamma_{k2}}=o(\lambda_{k,r_k})$ with $\gamma_{k2}$ given in~(V).
		
		\item The columns of $\mb{Z}_{k}^{(y)}:= (\mb{\Lambda}_{k}^{(y)})^{-1/2}(\mb{V}_{k}^{(y)})^\top \mb{Y}_k$ are i.i.d. copies of 
		$\bd{z}_{k}^{(y)}:= (\mb{\Lambda}_{k}^{(y)})^{-1/2}(\mb{V}_{k}^{(y)})^\top \bd{y}_k$, where $\mb{V}_{k}^{(y)}\mb{\Lambda}_{k}^{(y)}(\mb{V}_{k}^{(y)})^\top$ is the full SVD of $\cov(\bd{y}_k)$ with
		$\mb{\Lambda}_{k}^{(y)}=\diag(\lambda_{k,1},\dots,\lambda_{k,p_k})$.
		Vector $\bd{z}_{k}^{(y)}$'s entries $\{z_{ki}^{(y)}\}_{i=1}^{p_k}$ are independent with $E(z_{ki}^{(y)})=0$, $\var(z_{ki}^{(y)})=1$, and the sub-Gaussian norm $\sup_{q\ge 1}q^{-1/2}(E|z_{ki}^{(y)}|^q)^{1/q}\le \kappa_s$ 
		with a constant $\kappa_s>0$ for all $i\le p_k$.
		
		\item The matrix $\mb{B}_{f_k}^\top\mb{B}_{f_k}$ is a diagonal matrix.
		For all $i\le p_k$ and $\ell\le r_k$, $| \mb{B}_{f_k}^{[i,\ell]}  |\le \kappa_B\sqrt{\lambda_{k,\ell}/p_k}$ with a constant $\kappa_B>1$.
		
		\item Denote $\bd{e}_k=(e_{k1},\dots,e_{kp_k})^\top$ and $\bd{f}_k=(f_{k1},\dots, f_{kr_k})^\top$.
		Assume that $\| \cov(\bd{e}_k) \|_\infty<s_0$ with a constant $s_0>0$.
		For all $i\le p_k$ and $\ell\le r_k$, there exist positive constants $\gamma_{k1},\gamma_{k2},b_{k1}$ and $b_{k2}$ such that for $t>0$,
		$
		P(|e_{ki}|>t)\le \exp(-(t/b_{k1})^{\gamma_{k1}})
		$
		and
		$
		P(|f_{k\ell}|>t)\le \exp(-(t/b_{k2})^{\gamma_{k2}}).
		$
	\end{enumerate}
\end{assump}

\begin{theorem}\label{thm: CC}
	Suppose that Assumption~1 and $r_{12}\ge 1$ hold.
	Assume that
	any distinct values in $\{\cos\theta_{B\ell}\}_{\ell=1}^{r_{12}}\cup\{0,-\infty\}$
	are separated by at least a positive constant.
	Define
	\[
	\delta_\theta=
	\left(\frac{1}{\sqrt{n}}+\sum_{k=1}^2 \sqrt{\frac{\log p_k}{n
			\SNR_k
	}}\right)\wedge 1,
	\]
	where $\SNR_k=\frac{\tr\{\cov(\bd{x}_k)\}}{\tr\{\cov(\bd{e}_k)\}}$ is the signal-to-noise ratio of $\bd{y}_k$.
	For $k=1,2$, we have that
\[
			\frac{\|  \widehat{\mb{C}}-   \mb{C}\|_{\star}^2}{\frac{1}{2}(\| \mb{X}_1^S \|_{\star}^2+\| \mb{X}_2^S \|_{\star}^2 )}
		\ \vee \
		\frac{\|  \widehat{\mb{C}}^{(k)}-   \mb{C}^{(k)}\|_{\star}^2}{\| \mb{X}_k\|_{\star}^2}
		=O_P( \delta_\theta),
\]		
and
\[
			\left|\tr(\frac{1}{n}\widehat{\mb{C}}\widehat{\mb{C}}^\top)-\tr\{\cov(\bd{c})\}\right|
		=O_P( \delta_\theta^{1/2}),
\]
	where $\|\cdot\|_{\star}$ denotes either the Frobenius norm or the spectral norm, and $\mb{X}_k^S=[\tr(\mb{\Sigma}_k)]^{-1/2} \mb{X}_k$.
\end{theorem}

\begin{remark}\label{remark2}
	From Theorem~3 and Corollary~1 of \citep{Shu18},
	we have $\frac{\|\widehat{\mb{M}}_k-\mb{M}_k\|_{\star}^2}{\|\mb{X}_k\|_{\star}^2}=O_P(\delta_\theta)$
	for $\mb{M}_k\in \{\mb{X}_k, \mb{C}_k,\mb{D}_k\}$.
	Additionally by our Theorem~\ref{thm: CC} and the triangle inequality of norms,
	we also have this 
	error bound for $\mb{M}_k\in \{\mb{H}_k, \mb{\Delta}_k\}$. 
	Note that the scaled squared error in the Frobenius norm 
	indicates the scaled loss in matrix variation (sum of squares).
\end{remark}

\begin{theorem}\label{thm: P converge}
	Let $\widehat{\mb{P}}_*=\argmax_{\mb{P}\in \Pi_{p_1}} \tr(\widehat{\mb{Q}}_1^\top \mb{P} \widehat{\mb{Q}}_{2A} (\widehat{\mb{Q}}_1^\top \mb{P} \widehat{\mb{Q}}_{2A})^\top)$.
	Suppose that Assumption~1 and $r_{12}\ge 1$ hold.
	Then, we have 
	\[
	\left|
	\tr\big(\mb{Q}_1^\top \widehat{\mb{P}}_* \mb{Q}_{2A} (\mb{Q}_1^\top \widehat{\mb{P}}_* \mb{Q}_{2A})^\top\big)
	-
	\tr\big(\mb{Q}_1^\top \mb{P}_* \mb{Q}_{2A} (\mb{Q}_1^\top \mb{P}_* \mb{Q}_{2A})^\top\big)
	\right|=O_P(\delta_\theta).
	\]
\end{theorem}

For the row matching problem of $\mb{B}_1$ and $\mb{B}_{2A}$,
Theorem~\ref{thm: P converge}
provides an asymptotically vanishing bound on the change in 
the objective function value of \eqref{B's opt obj}
when the optimal solution $\mb{P}_*$ is replaced by $\widehat{\mb{P}}_*$.

\section{Simulation Studies}\label{CDPA sec: simulations}

In this section, we evaluate the finite-sample performance of the proposed CDPA estimation via simulations, comparing with the six D-CCA-type methods mentioned in Section~\ref{sec: intro}.

\subsection{Simulation setups}

We consider the following two simulation setups for signals $\{\bd{x}_k\}_{k=1}^2$.

\medskip

\noindent {\it Setup 1}: Let variable dimensions $p_1=p_2$, ranks $r_1=r_2=5$, and eigenvalues $\lambda_\ell(\mb{\Sigma}_k )=500-100(\ell-1)$ for $\ell\le 5$. The signals are
$\bd{x}_k=\mb{V}_k\mb{\Lambda}_k^{1/2}\bd{z}_k$ for $k=1,2$,
where canonical variables $[\bd{z}_1;\bd{z}_2]$ follow a multivariate Gaussian distribution with
$\bd{z}_k\sim \mathcal{N}(\bd{0}_{r_k\times 1},\mb{I}_{r_k\times r_k})$
and
$\cov(\bd{z}_1,\bd{z}_2)=
\diag\big\{\cos(\theta\wedge30^\circ),
\cos(\theta\wedge60^\circ),
\cos\theta,
\cos(\theta+15^\circ),
\cos((\theta+30^\circ) \wedge 90^\circ)\big\}$.
Let $\mb{Q}_k=\mb{V}_k^{[:,1:r_{12}]}$,
$\mb{P}=\mb{I}_{(p_1\vee p_2)\times (p_1\vee p_2)}$, 
and $(\mb{Q}_1^0)^\top \mb{Q}_2^0
=\cov(\bd{z}_1,\bd{z}_2)^{[1:r_{12},1:r_{12}]}$ of which the diagonal contains the cosines of principal angles
of $\colsp(\mb{B}_1^0)$ and $\colsp(\mb{B}_{2}^0)$, where
$\mb{M}_k^0=[\mb{M}_k;\mb{0}_{(p_k-p_1\wedge p_2)\times r_{12}}]$ with $\mb{M}\in \{\mb{Q},\mb{B}\}$.
Matrices $\{\mb{V}_k\}_{k=1}^2$ are randomly generated under the above constraints %
and are fixed for all simulation replications.

\medskip

\noindent {\it Setup 2}: We vary $p_1$ but fix $p_2=900$. The other settings are the same as in Setup~1. This setup aims to evaluate the performance of considered methods when $p_1\ne p_2$.

\medskip

We generate noises $\{e_{ki}\}_{k\le 2, i\le p_k} \overset{i.i.d.}{\sim}\mathcal{N}(0,\sigma_e^2)$ independent of signals $\{\bd{x}_k\}_{k=1}^2$.
Simulations are conducted with sample size $n=300$, variable dimension $p_1$ ranging from 100 to 1500, angle $\theta$ from $0^\circ$ to $75^\circ$, noise variance $\sigma_e^2$ from 0.25 to 9, and 1000 replications under each setting.
The proportion of average variance of $\bd{x}_1^S$ and $\bd{x}_2^S$ explained by $\bd{c}$, that is, $\tr\{\cov(\bd{c})\}$, has values
0.890, 0.479, 0.213, 0.126, 0.092 and 0.088
corresponding to
$\theta$ from $0^\circ$ to $75^\circ$ by a step $15^\circ$.
This pattern of the explained proportion of variance persists across all chosen values of $p_1$.



\subsection{Finite-sample performance of CDPA estimators}\label{subsec: CDPA siml}
We numerically evaluate the finite-sample performance of proposed CDPA estimators 
by comparing to the asymptotic results given in Section~\ref{sec: estimation}.
Since the signal-to-noise ratio $\SNR_k=1500/(p_k\sigma_e^2)$ in the above simulation setups, for simplicity we examine the trend of estimation errors 
with respect to $(p_k,\sigma_e^2)$ instead of $(p_k, \SNR_k)$ in the theorems. 
We use the true $\{r_k\}_{k=1}^2$, $r_{12}$, and $\mb{P}$ in our matrix estimation here to exclude the error induced by their misspecification. The ranks $\{r_k\}_{k=1}^2$ and $r_{12}$ can be well selected by the ED and MDL-IC methods, respectively,  as shown in \cite{Shu18}.
The selection of $\mb{P}$ by the DSPFP-based row-matching method in Section~\ref{sec: graph matching} is evaluated later in this subsection.

We first investigate the performance of our common-pattern matrix estimator $\widehat{\mb{C}}$ defined in \eqref{C_hat matrix}.
The first two rows of Figures~\ref{Fig: Setup 1, theta=15} and~\ref{Fig: Setup 1, theta=75}  summarize the scaled squared errors of $\widehat{\mb{C}}$ 
as studied in Theorem~\ref{thm: CC} and also its relative squared errors under Setup~1 with $\theta\in \{ 15^\circ, 75^\circ\}$.
The squared errors in the Frobenius norm represent
the scaled or relative losses in matrix variation (sum of squares).  The average estimation errors increase as the dimension $p_1$ or the noise variance $\sigma_e^2$ 
grows,  and are even well controlled under 0.1 for many cases with large $p_1\ge 900$ and large $\sigma_e^2\ge 4$ (or $\SNR_k\le 0.42$).
These results are consistent with the influence of $p_1$ and $\sigma_e^2$ ($=1500/(p_k\SNR_k)$, here) on the convergence rates given in Theorem~\ref{thm: CC}.
Similar numerical results are observed
for the scaled version $\widehat{\mb{C}}^{(k)}=[\tr(\widehat{\mb{\Sigma}}_k)]^{1/2}\widehat{\mb{C}}$ and the distinctive-pattern matrix estimator $\widehat{\mb{\Delta}}_{k}=\widehat{\mb{X}}_k-\widehat{\mb{C}}^{(k)}$ for $k\in\{1,2\}$,
and hence are omitted for brevity.

As a similarity indicator of signals $\bd{x}_1$ and $\bd{x}_2$, 
the common-pattern explained proportion of signal variance, $\tr\{\cov(\bd{c})\}$, is estimated by $\tr(\frac{1}{n}\widehat{\mb{C}}\widehat{\mb{C}}^\top)=\frac{1}{n}\| \widehat{\mb{C}}\|_F^2$. 
The third rows of 
Figures~\ref{Fig: Setup 1, theta=15} and~\ref{Fig: Setup 1, theta=75} plot the average absolute error and the average relative error of this estimator for Setup~1 with $\theta\in \{ 15^\circ, 75^\circ\}$. Same with Theorem~\ref{thm: CC}, the row shows that the average  estimation errors grow with increasing $p_1$ or $\sigma_e^2$ and have a larger magnitude than those squared errors of $\widehat{\mb{C}}$ as shown in the first two rows of the figure. 
The errors are controlled below 0.1 even for some cases with large $p_1\ge 900$ or  $\sigma_e^2\ge 4$.

For the row-matching approach of coefficient matrices $\{\mb{B}_k\}_{k=1}^2$ described in Section~\ref{sec: graph matching}, its theoretical performance stated in
Theorem~\ref{thm: P converge} is numerically investigated
with the intractable $\mb{P}_*$ and $\widehat{\mb{P}}_*$ 
being replaced by their DSPFP approximations denoted as 
$\mb{P}_a$ and $\widehat{\mb{P}}_a$. 
The fourth rows of Figures~\ref{Fig: Setup 1, theta=15} and~\ref{Fig: Setup 1, theta=75} display
the average  absolute and relative errors of $\tr\{(\mb{Q}_1^0)^\top \widehat{\mb{P}}_a \mb{Q}_{2}^0 [(\mb{Q}_1^0)^\top \widehat{\mb{P}}_a \mb{Q}_{2}^0]^\top\}$ for Setup~1 with $\theta\in \{ 15^\circ, 75^\circ\}$.
Although its absolute error
seems to have larger values than that of its oracle version (with $\widehat{\mb{P}}_*$) expected in Theorem~\ref{thm: P converge},
its relative error is controlled 
under or around 0.1 even for some cases with large $p_1\ge 900$ or $\sigma_e^2\ge 4$, and moreover,  the two types of errors both follow 
the influence of $p_1$ and $\sigma_e^2$ ($=1500/(p_k\SNR_k)$, here) on the convergence rate shown in the theorem.

The above result patterns also generally hold for settings with more different
values of $\theta$ (or equivalently $\tr\{\cov(\bd{c})\}$) and for those under Setup~2 where $p_1\ne p_2$, which are provided in  Appendix~\ref{sec: add siml}.

\begin{figure}[p!]
	\includegraphics[width=0.8\textwidth]{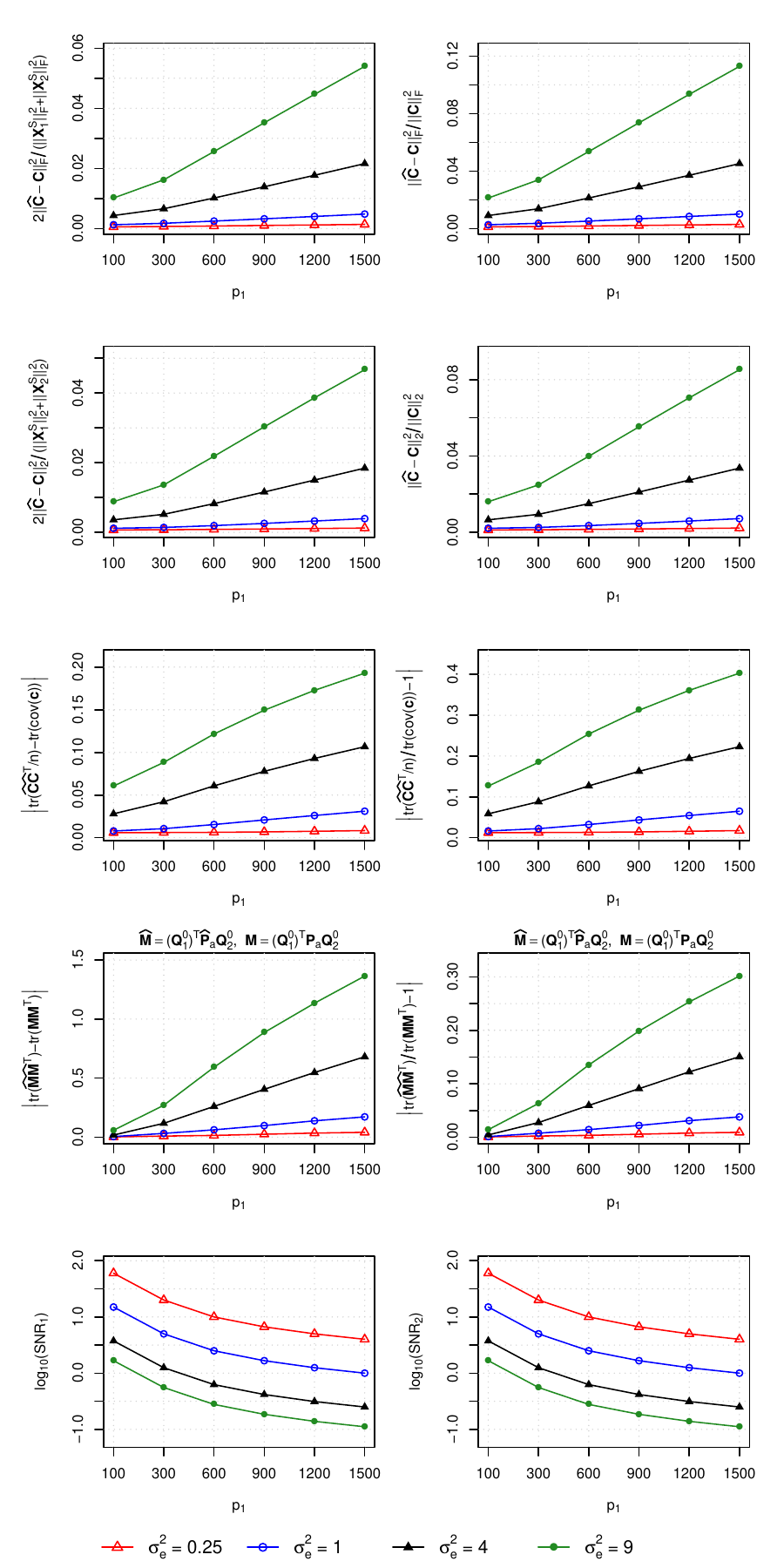}
	\caption{Average errors of CDPA estimates over 1000 replications and the signal-to-noise ratios for Setup 1 with $\theta=15^\circ$.}
	\label{Fig: Setup 1, theta=15}
\end{figure}

\begin{figure}[p!]
	\includegraphics[width=0.8\textwidth]{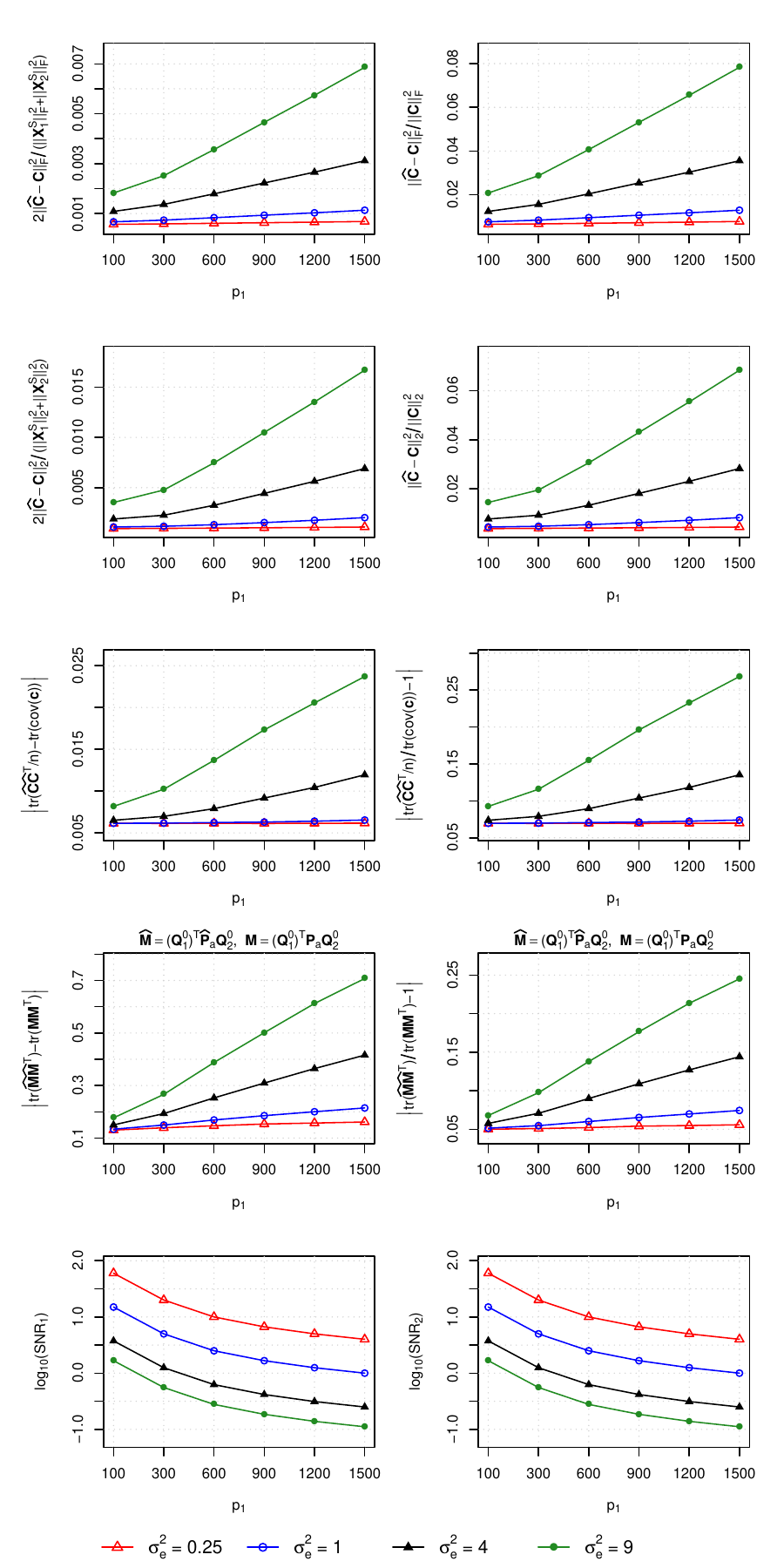}
	\caption{Average errors of CDPA estimates over 1000 replications and the signal-to-noise ratios for Setup 1 with $\theta=75^\circ$.}
	\label{Fig: Setup 1, theta=75}
\end{figure}

\subsection{Performance of related methods}\label{sec: siml others}
We now investigate the numerical performance of the six competing methods, including D-CCA, OnPLS, COBE, JIVE, AJIVE, and DISCO-SCA. 
Unlike our CDPA, all the six D-CCA-type methods
are developed without taking into account
the common and distinctive patterns between the two coefficient matrices $\{\mb{B}_k\}_{k=1}^2$ of their common latent factors.\footnote{\label{ft: for OnPLS siml} 
		We implement the OnPLS with the post-processing step in footnote~\ref{ft: OnPLS}
		to obtain the coefficient matrices $\{\mb{B}_k\}_{k=1}^2$ of its common latent factors.
}

The simulation here aims
to corroborate the existence of both common and distinctive patterns
in their coefficient matrices $\{\mb{B}_k\}_{k=1}^2$.
The existence can be
shown if 
$\colsp(\mb{B}_1^0)$ and $\colsp(\mb{B}_2^0)$
are neither overlapping nor orthogonal, that is,
their first principal angle
$\theta_{B1}\notin \{0^\circ, 90^\circ\}$
or equivalently their first canonical correlation
$\cos\theta_{B1}\notin \{1,0\}$.
Since the six D-CCA-type methods have different definitions of
$\{\bd{c}_k\}_{k=1}^2$
for decomposition~\eqref{decomp vec}
due to their different constraints, they may have different $\{\mb{B}_k^0\}_{k=1}^2$ and thus different values of $\theta_{B1}$ under our simulation setups. 
The ground-truth $\theta_{B1}$ is $\theta\wedge30^\circ$ for D-CCA in our simulation, 
but may not be easy to theoretically determine for the other five methods and is thus estimated by simulated data.

Table~\ref{tab: others' principal angle} summarizes the first principal angle $\theta_{B1}$ and its cosine $\cos\theta_{B1}$ of $\colsp(\mb{B}_1^0)$ and $\colsp(\mb{B}_2^0)$ estimated by the six methods
under the two simulation setups with $(\theta,p_1,\sigma_e^2)=(75^\circ, 300, 1)$.
We see that the COBE and AJIVE give zero common-source matrix estimates
and thus fail to discover any common pattern of the two correlated signal datasets.
The average estimates of $\theta_{B1}$ and $\cos\theta_{B1}$ from the other four methods are all close to $30^\circ$ and $0.866$, respectively, with very small standard deviations.
Therefore, there is significant statistical
evidence that 
their $\theta_{B1}\notin \{0^\circ, 90^\circ\}$
and $\cos\theta_{B1}\notin \{1,0\}$.
This indicates the non-negligible existence 
of both the common and the distinctive patterns between their coefficient matrices, 
but these patterns are unfortunately not
considered by these D-CCA-type methods.

\begin{table}[tb!]
	\caption{Averages (and standard deviations) of the estimates for
		the first principal angle
		$\theta_{B1}$ and its cosine $\cos \theta_{B1}$ of $\colsp(\mb{B}_1^0)$ and $\colsp(\mb{B}_2^0)$ obtained from existing methods over 1000 simulation replications with $(\theta,p_1,\sigma_e^2)=(75^\circ, 300, 1)$.}
	\begin{tabular}{cccc cc c cccc}
		\hline\noalign{\smallskip}
		Method
		& Setup 1
		&Setup 2
		\\
		\noalign{\smallskip}\hline \noalign{\smallskip}
		D-CCA & 30.6$^\circ$(0.374$^\circ$)/0.860(0.003) & 30.8$^\circ$(0.339$^\circ$)/0.859(0.003)
		\\
		OnPLS & 31.8$^\circ$(0.931$^\circ$)/0.850(0.009) & 32.0$^\circ$(0.881$^\circ$)/0.848(0.008)
		\\
		COBE& NA & NA
		\\
		JIVE & 32.0$^\circ$(0.998$^\circ$)/0.848(0.009) & 32.9$^\circ$(1.210$^\circ$)/0.840(0.012)
		\\
		AJIVE & NA  & NA
		\\
		DISCO-SCA & 31.1$^\circ$(0.573$^\circ$)/0.856(0.005) & 31.3$^\circ$(0.545$^\circ$)/0.855(0.005)
		\\		\noalign{\smallskip}\hline
	\end{tabular}
	\vspace{0.5\baselineskip}
	
	{Note: NA means the result is not available due to zero common-source matrix estimates. }
	\label{tab: others' principal angle}
\end{table}

\section{Real Data Analysis}\label{CDPA sec: real data}

We apply our CDPA to two real-world data examples, respectively, from the HCP and TCGA, 
comparing with the six D-CCA-type methods
mentioned in Section~1.
We focus on the comparison with the state-of-the-art D-CCA in this section, and present the results of the other five methods in Appendix~\ref{sec: add data analysis}.

\subsection{Application to HCP motor-task functional MRI data}\label{sec: HCP app}

We consider the HCP motor-task functional MRI data obtained from 1080 healthy young adults \citep{Barc13}.
All participants were asked by visual cues to perform five motor tasks during the image scanning, including tapping left and right fingers, squeezing left and right toes, and moving tongue.
From the acquired brain images, for every participant and each task,
the HCP computed
a $z$-statistic map of the task's contrast against the fixation baseline at 91,282 grayordinates including 59,412 cortical surface vertices and 31,870 subcortical gray matter voxels.
The $z$-statistic maps of all participants for each individual task constitute a 91,282$\times$1080 data matrix.
We focus on the left-hand and right-hand tasks,
and 
apply the proposed CDPA to discover their common pattern on the brain, 
with comparison to the D-CCA method.

\begin{figure}[b!]	\vspace{1\baselineskip}
	\begin{subfigure}{0.45\textwidth}
		\centering\includegraphics[width=1\textwidth]{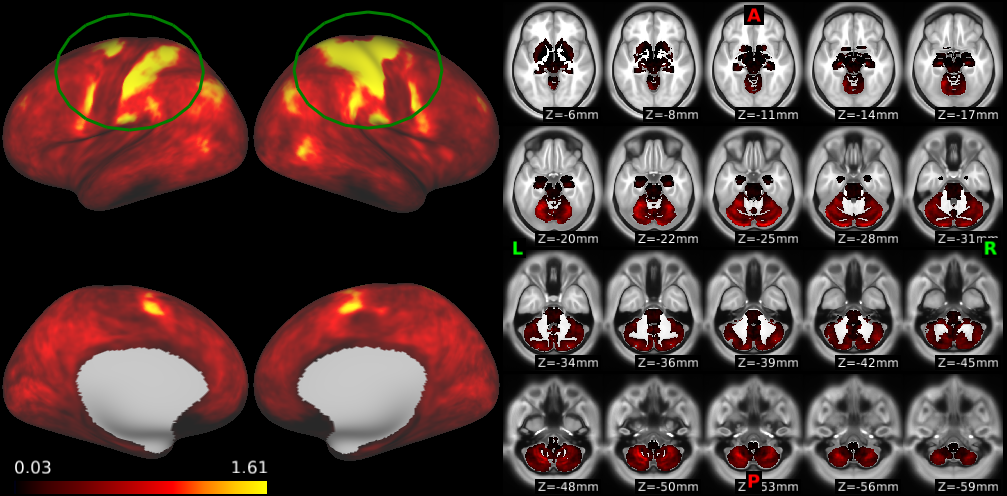}
		\caption{$\widehat{\var}(\bd{x}_L)$ of D-CCA}
	\end{subfigure}
	\begin{subfigure}{0.45\textwidth}
		\centering\includegraphics[width=1\textwidth]{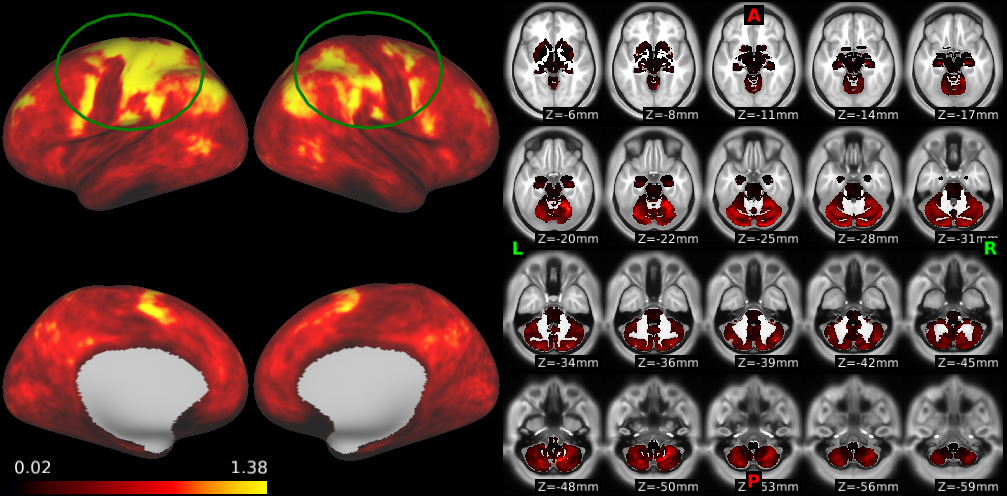}
		\caption{$\widehat{\var}(\bd{x}_R)$ of D-CCA}
	\end{subfigure}
	\begin{subfigure}{0.45\textwidth}\bigskip
		\centering\includegraphics[width=1\textwidth]{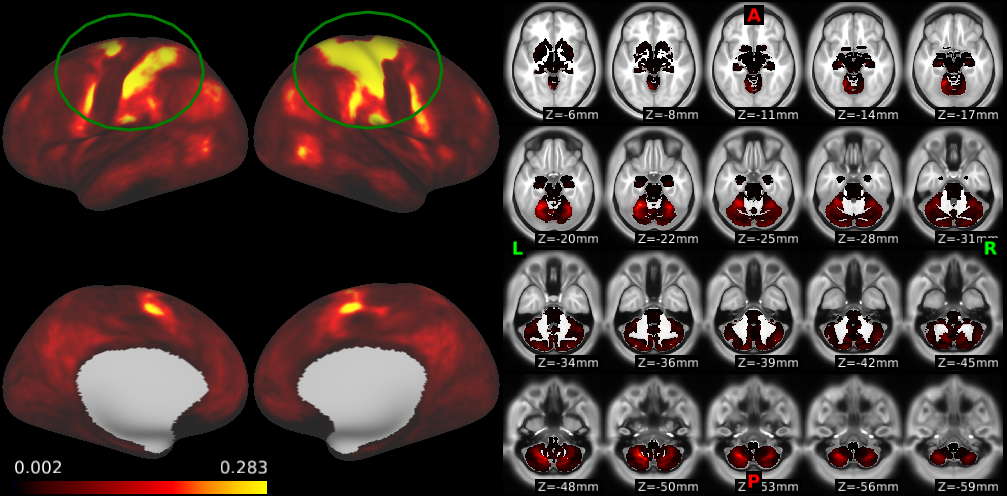}
		\caption{$\widehat{\var}(\bd{c}_L)$ of D-CCA}
	\end{subfigure}
	\begin{subfigure}{0.45\textwidth}\bigskip
		\centering\includegraphics[width=1\textwidth]{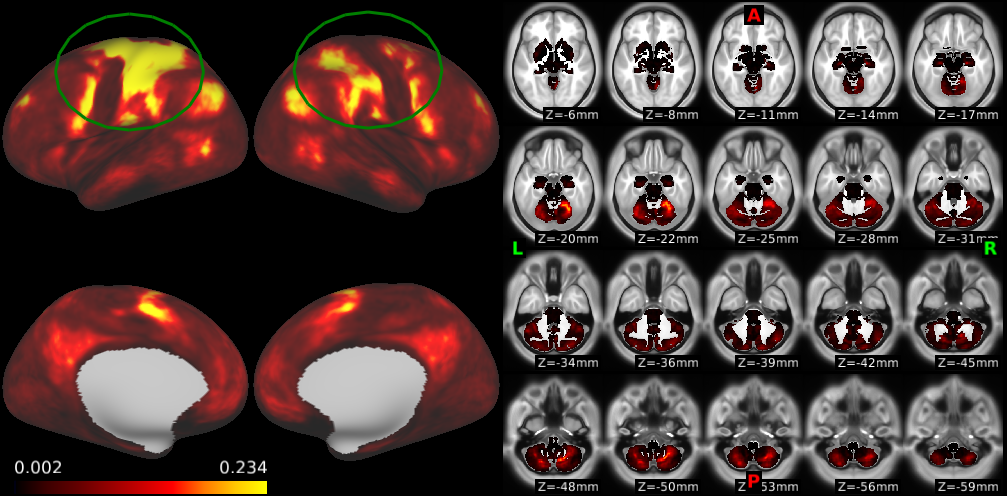}
		\caption{$\widehat{\var}(\bd{c}_R)$ of D-CCA}
	\end{subfigure}
	\begin{center}\smallskip
		\begin{subfigure}{1\textwidth}
			\centering\includegraphics[width=0.45\textwidth]{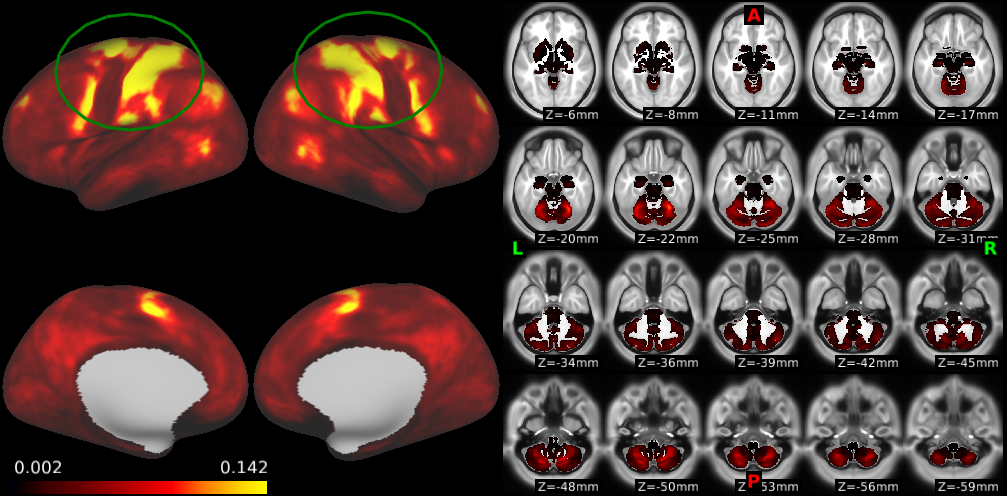}
			\caption{$\widehat{\var}(\bd{c})\cdot [\tr^{\frac{1}{2}}\{\widehat{\cov}(\bd{x}_L)\}+\tr^{\frac{1}{2}}\{\widehat{\cov}(\bd{x}_R)\}]^2/4$ 
				of CDPA}
		\end{subfigure}
	\end{center}
	\caption{The variance maps estimated by the D-CCA and CDPA methods for HCP motor-task functional MRI data. The notations $\widehat{\var}$ and $\widehat{\cov}$ denote the sample variance vector and sample covariance matrix obtained from the corresponding recovered sample matrix. In each subfigure, the left part displays the cortical surface with the outer side shown in the first row and the inner side in the second row; the right part shows the subcortical area on 20 $xy$ slides at the $z$ axis. The somatomotor cortex is annotated by green circles.}
	\label{CDPA: HCP real data}
\end{figure}

Each of the two observed data matrices is row-centered by subtracting the average within each row.
Since all $z$-statistic maps of the two motor tasks are obtained from the same measurement and at the same set of grayordinates, there is no need to choose the signs or match the rows of 
the two data matrices. We consider the variance maps of
$\{\bd{x}_L,\bd{x}_R, \bd{c}_L,\bd{c}_R,\bd{c}\}$ on the brain,
which are estimated by the sample variances computed from 
the sample matrix estimates $\{\widehat{\mb{X}}_L,\widehat{\mb{X}}_R,\widehat{\mb{C}}_L,\widehat{\mb{C}}_R,\widehat{\mb{C}}\}$ 
obtained by D-CCA and CDPA.
Here, the subscripts $L$ and $R$ denote the left-hand and right-hand tasks. 
The ranks $\{r_L,r_R\}$ and $r_{12}$ are all selected as two 
by the ED  and MDL-IC methods, respectively.
The proportions of corresponding signal variances explained by
common-source vectors $\bd{c}_L$ and $\bd{c}_R$ are $\frac{\tr\{\cov(\bd{c}_L)\}}{\tr\{\cov(\bd{x}_L) \}} \approx\frac{\| \widehat{\mb{C}}_L\|_F^2}{\| \widehat{\mb{X}}_L\|_F^2}=0.113$
and  $\frac{\tr\{\cov(\bd{c}_R)\}}{\tr\{\cov(\bd{x}_R) \}} \approx\frac{\| \widehat{\mb{C}}_R\|_F^2}{\| \widehat{\mb{X}}_R\|_F^2}=0.111$. The common-pattern explained proportion of signal variance is
$\tr\{\cov(\bd{c})\}\approx \frac{1}{n}\| \widehat{\mb{C}}\|_F^2=0.077$.

Figure~\ref{CDPA: HCP real data} presents the estimated variance maps of D-CCA and CDPA.
For all the five maps, the estimated variances of cortical surface vertices overall dominate those of subcortical voxels. We hence focus on the part of each variance map for the cortical surface. 
From the estimated signal variance maps $\widehat{\var}(\bd{x}_L)$ and $\widehat{\var}(\bd{x}_R)$ shown in Figure~\ref{CDPA: HCP real data}~(a) and (b), we see that
the right half brain is more active, with larger variances, on the cortical surface for the left-hand task, while the pattern is almost opposite for the right-hand task.
In particular, the contralateral pattern is clearly seen on the somatomotor cortex annotated by green circles, a brain region known to be linked with hand tasks \citep{Buck11}.
A similar contralateral pattern is also observed for D-CCA's $\widehat{\var}(\bd{c}_L)$ and $\widehat{\var}(\bd{c}_R)$ in 
Figure~\ref{CDPA: HCP real data}~(c) and (d). 
This indicates that the $\bd{c}_k$ vector of D-CCA retains some distinctive pattern of $\bd{x}_k$ for $k\in \{L,R\}$. It is not surprising because
$\bd{c}_L$ and $\bd{c}_R$ have different coefficient matrices of the common latent factors,  which are $r_{12}$ columns in the coefficient matrices of canonical variables for $\bd{x}_L$ and $\bd{x}_R$, respectively, as shown in equation \eqref{X=betaC+betaD}.
In contrast, 
our CDPA's common-pattern vector $\bd{c}$ in Figure~\ref{CDPA: HCP real data} (e)  has an estimated variance map that is nearly symmetric on the two hemispheres, 
and thus is more reasonable than D-CCA's common-source vectors $\bd{c}_L$ and $\bd{c}_R$ to represent the common pattern of the left-hand and right-hand tasks on the brain.

\subsection{Application to TCGA breast cancer genomic datasets}\label{sec: TCGA}
With the aim to discover new breast cancer subtypes,
we apply the proposed CDPA to two TCGA breast cancer genomic datasets \citep{Kobo12}, and compare the results with the D-CCA.
We consider the DNA methylation data and mRNA expression data obtained from a common set of 703 tumor samples.
Following the preprocessing procedure of \citep{Lock13b},
we select the top 1100 variable probes for the DNA methylation dataset
and the top 896 variably expressed genes for the mRNA expression dataset.
The tumor samples
are categorized by the classic PAM50 model \citep{Park09}
into four intrinsic subtypes, including
124 Basal-like, 58 HER2-enriched, 348 Luminal A, and 173 Luminal B tumors. 

The two data matrices of interest have sizes $1100{\times}703$ and $896{\times}703$, 
and are row-centered
before analysis.
The ranks 
$(r_{\text{DNA}},r_{\text{mRNA}},r_{12})$ are selected by
the ED and MDL-IC methods as $(3,2,2)$.
From the D-CCA,
the proportions of signal variances explained by common-source vectors $\bd{c}_{\text{DNA}}$ and $\bd{c}_{\text{mRNA}}$ are
$\frac{\tr\{\cov(\bd{c}_{\text{DNA}})\}}{\tr\{\cov(\bd{x}_{\text{DNA}}) \}} \approx\frac{\| \widehat{\mb{C}}_{\text{DNA}}\|_F^2}{\| \widehat{\mb{X}}_{\text{DNA}}\|_F^2}=0.210$ and 
$\frac{\tr\{\cov(\bd{c}_{\text{mRNA}})\}}{\tr\{\cov(\bd{x}_{\text{mRNA}}) \}} \approx\frac{\| \widehat{\mb{C}}_{\text{mRNA}}\|_F^2}{\| \widehat{\mb{X}}_{\text{mRNA}}\|_F^2}=0.422$, 
indicating 
different influences of the common latent factors on the two signal datasets. 
Thus, by ignoring these different common-source influences, their $\bd{c}_{\text{DNA}}$ and $\bd{c}_{\text{mRNA}}$ are not appropriate to be viewed as the common pattern of $\bd{x}_{\text{DNA}}$ and $\bd{x}_{\text{mRNA}}$.

Since only 126 (11.5\%) DNA methylation probes can be mapped to the genes of the considered mRNA expression data,
for simplicity we match the rows of the two data matrices by using the graph-matching based approach described in Section~\ref{sec: graph matching} before implementing CDPA. The CDPA method shows that the common-pattern explained proportion of signal variance
$\tr\{\cov(\bd{c})\}\approx\frac{1}{n}\| \widehat{\mb{C}}\|_F^2$ is 0.161 (95\% CI = [0.154, 0.185]) for $\bd{x}_{\text{DNA}}$ and $\bd{x}_{\text{mRNA}}$,
but is only 0.049 (95\% CI = [0.046, 0.057])  for $\bd{x}_{\text{DNA}}$ and $-\bd{x}_{\text{mRNA}}$,
where each 95\% CI is computed by 5000 bootstrapping samples.
We hence focus on the common and distinctive patterns extracted from
$\{\bd{x}_{\text{DNA}},\bd{x}_{\text{mRNA}}\}$ rather than $\{\bd{x}_{\text{DNA}},-\bd{x}_{\text{mRNA}}\}$.



We explore new cancer subtypes by conducting clustering analysis on each observed 
or recovered matrix from the CDPA and D-CCA methods. We use the Ward's hierarchical clustering method \citep{Ward63} with the Euclidean distance, and simply specify the number of clusters to be four, which is the same number of the PAM50 intrinsic subtypes.

Table~\ref{tab: log-rank table for CDPA} compares the differences in survival curves of identified clusters or given subtypes using
two most popular methods, the log-rank test \citep{Mant66} 
and the Peto-Peto's Wilcoxon test \citep{Peto72}, where the latter test is more sensitive to early survival differences.
Our CDPA's $\widehat{\mb{\Delta}}_{\text{mRNA}}$-identified clusters and
the PAM50 intrinsic subtypes
both have very significantly distinct survival behaviors
with the two smallest p-values $\le 0.009$ in both tests,  while the other matrices generate much less pronounced clusters, in particular, the matrices $\{\widehat{\mb{C}}_k,\widehat{\mb{D}}_k\}_{k\in \{\text{DNA},\text{mRNA}\}}$ of D-CCA all have large p-values $\ge 0.290$.
By comparing the p-values of 
$\widehat{\mb{C}}$,
$\widehat{\mb{X}}_k$ and $\widehat{\mb{\Delta}}_k$ for each $k$, 
the improved discriminative power of 
distinctive-pattern matrix estimate $\widehat{\mb{\Delta}}_k$ can be attributed to
removing the less sensitive common-pattern matrix estimate
$\widehat{\mb{C}}$ from the denoised data matrix $\widehat{\mb{X}}_k$.
The adjusted Rand index \citep{Hube85}
between our $\widehat{\mb{\Delta}}_{\text{mRNA}}$-identified clusters and PAM50 subtypes 
is 0.343 (95\% CI = [0.335, 0.352]),
indicating a poor agreement.
It is evident that, built on top of D-CCA, our CDPA 
can benefit data mining with additional pattern matrices 
$\big\{\widehat{\mb{C}},\{\widehat{\mb{\Delta}}_k, \widehat{\mb{H}}_k\}_{k\in \{\text{DNA},\text{mRNA}\}}\big\}$.

\begin{figure}[b!]
	\begin{subfigure}{0.45\textwidth}
		\centering\includegraphics[width=1\textwidth]{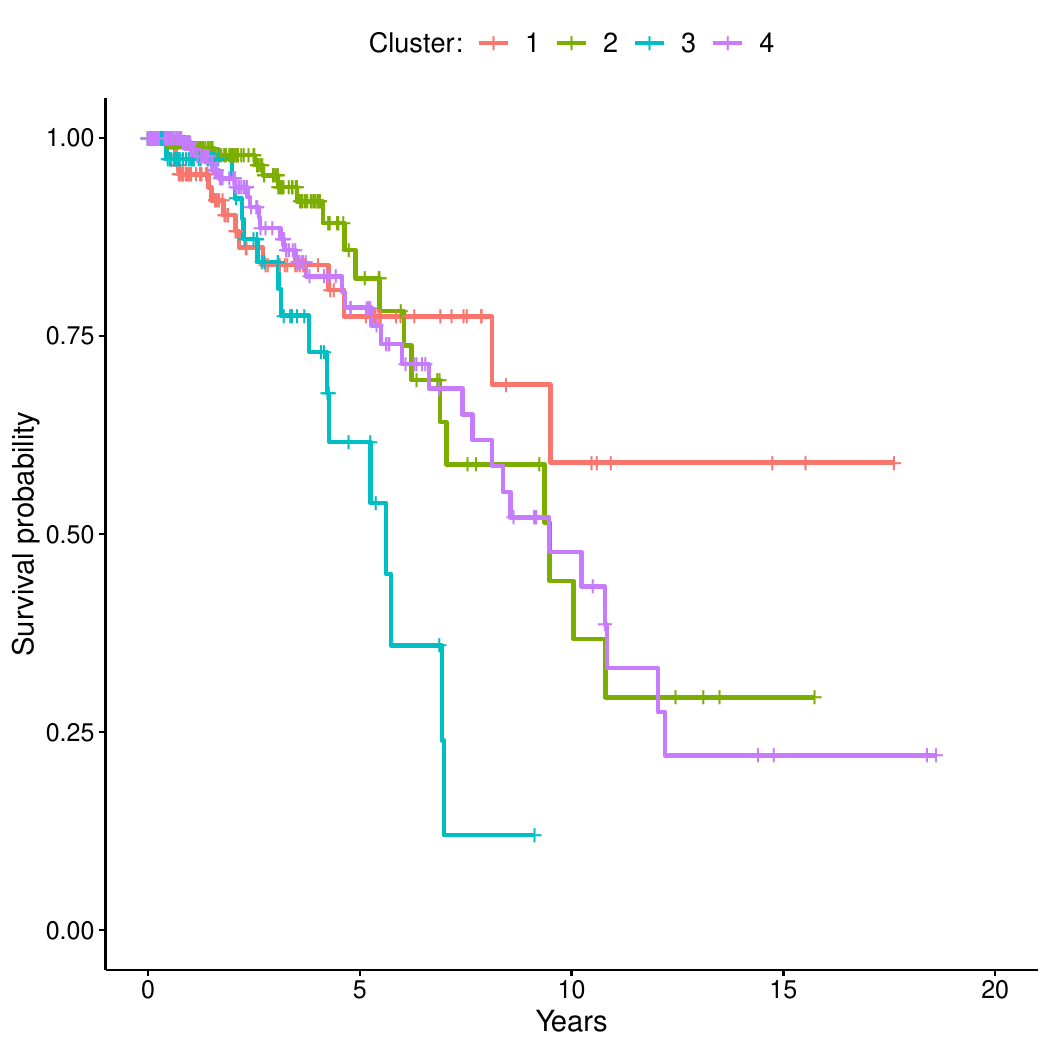}
		\caption{CDPA's $\widehat{\mb{\Delta}}_{\text{mRNA}}$-identified clusters}
	\end{subfigure}
	\begin{subfigure}{0.45\textwidth}
		\centering\includegraphics[width=1\textwidth]{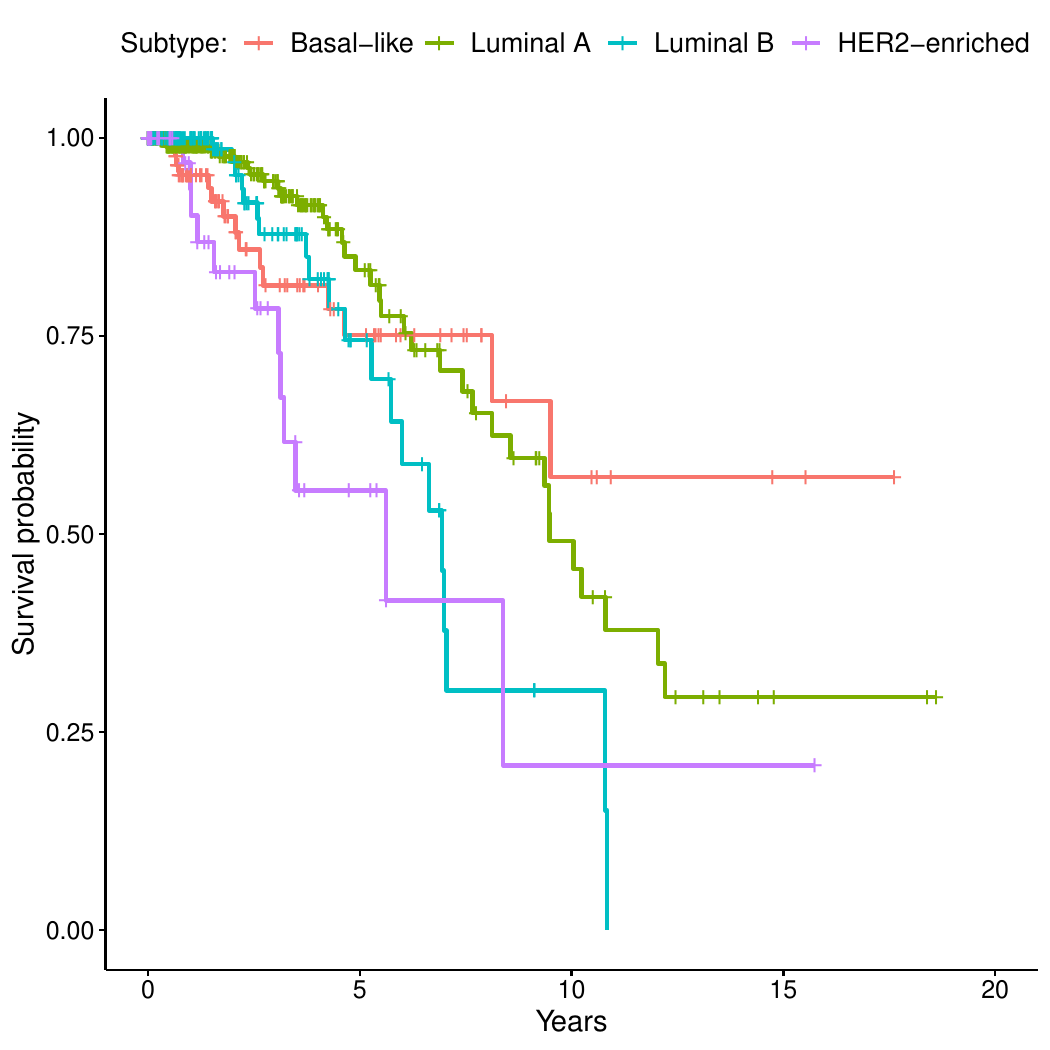}
		\caption{PAM50 subtypes}
	\end{subfigure}
	\caption{Kaplan-Meier survival curves of TCGA breast cancer clusters  and subtypes.}
	\label{fig: KM curves}
\end{figure}

\begin{table}[b!]
	\caption{Log-rank test and Peto-Peto's Wilcoxon test for survival curve differences among the clusters identified from each observed 
		or recovered matrix from the CDPA and D-CCA methods for TCGA breast cancer datasets. }	
\scalebox{0.85}{
	\begin{tabular}{cccc cc c cccc}
		\hline\noalign{\smallskip}
		& Log-rank/Peto's&&  &       Log-rank/Peto's
		&  &&   Log-rank/Peto's
		\\
		Data
		& p-values& 
		&Data
		& p-values& 
		&Data
		& p-values\\
		\noalign{\smallskip}	
		\cline{1-2} 	\cline{4-5} 	\cline{7-8}		
		\noalign{\smallskip}
		${\mb{Y}}_{\text{DNA}}$&0.175/0.230 &&${\mb{Y}}_{\text{mRNA}}$&0.251/0.299 &&$[{\mb{Y}}_{\text{DNA}}^N;{\mb{Y}}_{\text{mRNA}}^N]$&0.245/0.129
		\\
		$\widehat{\mb{X}}_{\text{DNA}}$&\textbf{0.077/0.112} &&$\widehat{\mb{X}}_{\text{mRNA}}$&\textbf{0.063/0.061} &&$[\widehat{\mb{X}}_{\text{DNA}}^N;\widehat{\mb{X}}_{\text{mRNA}}^N]$&0.565/0.619
		\\
		$\widehat{\mb{C}}_{\text{DNA}}$& 0.820/0.979 &&$\widehat{\mb{C}}_{\text{mRNA}}$& 0.619/0.704 &&$[\widehat{\mb{C}}_{\text{DNA}}^N;\widehat{\mb{C}}_{\text{mRNA}}^N]$&0.752/0.751
		\\
		$\widehat{\mb{D}}_{\text{DNA}}$&0.515/0.417 &&$\widehat{\mb{D}}_{\text{mRNA}}$&0.290/0.354 &&$[\widehat{\mb{D}}_{\text{DNA}}^N;\widehat{\mb{D}}_{\text{mRNA}}^N]$&0.149/0.223
		\\
		$\widehat{\mb{H}}_{\text{DNA}}$&0.430/0.502 &&$\widehat{\mb{H}}_{\text{mRNA}}$&0.330/0.409 &&$[\widehat{\mb{H}}_{\text{DNA}}^N;\widehat{\mb{H}}_{\text{mRNA}}^N]$&0.337/0.369
		\\
		$\widehat{\mb{\Delta}}_{\text{DNA}}$&\textbf{0.058/0.075} &&$\widehat{\mb{\Delta}}_{\text{mRNA}}$&\textbf{0.004/0.009} &&
		$[\widehat{\mb{\Delta}}_{\text{DNA}}^N;\widehat{\mb{\Delta}}_{\text{mRNA}}^N]$&0.218/0.208
		\\
		$\widehat{\mb{C}}$&\textbf{0.106/0.163}
		&&    PAM50 & \textbf{0.003}/\textbf{0.001}
		&&
		\\
		\noalign{\smallskip}	\hline
	\end{tabular}}
	
	\vspace{0.5\baselineskip}
	{Note: Denote $\mb{M}^N=\mb{M}/\|\mb{M}\|_F$ for any matrix $\mb{M}$. 
	}
	\label{tab: log-rank table for CDPA}
\end{table}

\begin{table}[h!]
	\caption{Log-rank test and Peto-Peto's Wilcoxon test for survival curve differences among CDPA's $\widehat{\mb{\Delta}}_{\text{mRNA}}$-identified clusters and PAM50 subtypes
		for TCGA breast cancer data.}
	\scalebox{0.88}{
		\begin{tabular}{cccc cc c cccc}
			\hline\noalign{\smallskip}
			& Log-rank/Peto's&&  &       Log-rank/Peto's
			\\
			Comparison
			& p-values& 
			&Comparison
			& p-values	\\
			\noalign{\smallskip}\cline{1-2}\cline{4-5}\noalign{\smallskip}
			$\widehat{\mb{\Delta}}_{\text{mRNA}}$-1 vs. $\widehat{\mb{\Delta}}_{\text{mRNA}}$-2&0.895/0.550 &&
			$\widehat{\mb{\Delta}}_{\text{mRNA}}$-1 vs. $\widehat{\mb{\Delta}}_{\text{mRNA}}$-3&0.022/0.070 
			\\
			$\widehat{\mb{\Delta}}_{\text{mRNA}}$-1 vs. $\widehat{\mb{\Delta}}_{\text{mRNA}}$-4&0.491/0.816 &&
			$\widehat{\mb{\Delta}}_{\text{mRNA}}$-2 vs. $\widehat{\mb{\Delta}}_{\text{mRNA}}$-3&3.34e-4/5.35e-4
			\\
			$\widehat{\mb{\Delta}}_{\text{mRNA}}$-2 vs. $\widehat{\mb{\Delta}}_{\text{mRNA}}$-4&0.375/0.320&&
			$\widehat{\mb{\Delta}}_{\text{mRNA}}$-3 vs. $\widehat{\mb{\Delta}}_{\text{mRNA}}$-4&0.006/0.013
			\\
			$\widehat{\mb{\Delta}}_{\text{mRNA}}$-3 vs. Basal-like&0.041/0.121&&
			$\widehat{\mb{\Delta}}_{\text{mRNA}}$-3 vs. Luminal A&5.89e-5/1.26e-4
			\\
			$\widehat{\mb{\Delta}}_{\text{mRNA}}$-3 vs. Luminal B&0.069/0.070 &&
			$\widehat{\mb{\Delta}}_{\text{mRNA}}$-3 vs. HER2-enriched&0.585/0.361
			\\		\noalign{\smallskip}\hline
	\end{tabular}}
	\label{tab: log-rank pairwise}
\end{table}



\begin{table}[t!]\bigskip
	\caption{Matching matrix and clinical features of CDPA's
		$\widehat{\mb{\Delta}}_{\text{mRNA}}$-identified clusters
		and PAM50 subtypes
		for TCGA breast cancer data analysis.}
	\scalebox{0.81}{	\hspace{-1.3cm}
		\begin{tabular}{cccccccccccccccccccccc}
			\hline	\noalign{\smallskip}
			PAM50&&$\widehat{\mb{\Delta}}_{\text{mRNA}}$-1 & $\widehat{\mb{\Delta}}_{\text{mRNA}}$-2 & $\widehat{\mb{\Delta}}_{\text{mRNA}}$-3 & $\widehat{\mb{\Delta}}_{\text{mRNA}}$-4 && Total 
			&ER$+$/$-$ & PR$+$/$-$ &HER2$+$/$-$\\
			\noalign{\smallskip}		\cline{1-1} \cline{3-6} \cline{8-11} \noalign{\smallskip}
			Basal-like&& 122 &  0 &  0 &  2 &&124 &6\%/81\% &6\%/79\%&7\%/54\%
			\\
			Luminal A&&0 &194 &31  &123 &&348 &89\%/1\% &82\%/8\% &9\%/53\%
			\\
			Luminal B&&0  &13 & 77  &83 &&173 &87\%/2\% &72\%/17\%  &16\%/46\%
			\\
			HER2-enriched&&8  & 8 & {\bf 10}  &32 &&58 &{\bf 33\%/52\%} &{\bf 17\%/71\%} & {\bf 62\%/16\%}
			\\
			\noalign{\smallskip}	\hline\noalign{\smallskip}
			Total	&& 130 &	215 & 118 & 240 &&703
			\\
			ER$+$/$-$ && 6\%/80\% &  91\%/3\%  & {\bf 80\%/4\%}&79\%/10\%
			\\
			PR$+$/$-$ && 5\%/79\% & 84\%/10\%  &{\bf 64\%/20\%}&68\%/20\%
			\\
			HER2$+$/$-$ && 8\%/52\% & 10\%/57\%&{\bf 19\%/42\%} & 21\%/41\% 
			\\
			\noalign{\smallskip}	\hline
		\end{tabular}
	}	
	
	\vspace{0.5\baselineskip}	
	{Notes: 
		The columns of the matching matrix are well reordered such that its diagonal sum is maximized. Receptor status for estrogen (ER), progesterone (PR) and human epidermal growth factor~2 (HER2) includes positive ($+$), negative ($-$), and N/A or equivocal. \par}
	\label{tab: match table}
\end{table}


Let $\widehat{\mb{\Delta}}_{\text{mRNA}}$-$i$ denote the $i$-th cluster identified from $\widehat{\mb{\Delta}}_{\text{mRNA}}$.
Figure~\ref{fig: KM curves} displays the Kaplan-Meier survival curves 
of $\widehat{\mb{\Delta}}_{\text{mRNA}}$-identified clusters 
and PAM50 subtypes.
With the worst survival curve
among the four identified clusters,
$\widehat{\mb{\Delta}}_{\text{mRNA}}$-3 behaves similar to the HER2-enriched subtype, but is 
notably different with all other identified clusters and intrinsic subtypes.
This is further confirmed in Table~\ref{tab: log-rank pairwise} by the minimum p-value of corresponding log-rank test and 
Peto-Peto’s Wilcoxon test.
Also seen in the table,
the other three $\widehat{\mb{\Delta}}_{\text{mRNA}}$-identified clusters have no significant survival differences with large p-values $\ge 0.320$. Moreover,
the matching matrix in Table~\ref{tab: match table}
shows that most of 
$\widehat{\mb{\Delta}}_{\text{mRNA}}$-1 and $\widehat{\mb{\Delta}}_{\text{mRNA}}$-2 samples belong to the Basal-like and Luminal A subtypes, respectively.
Hence, the other three $\widehat{\mb{\Delta}}_{\text{mRNA}}$-identified clusters are less of interest to be new subtypes, 
and we focus on $\widehat{\mb{\Delta}}_{\text{mRNA}}$-3 which has the poorest survival,
and further compare it with the HER2-enriched subtype.
From Table~\ref{tab: match table},
we see that the $\widehat{\mb{\Delta}}_{\text{mRNA}}$-3 cluster (118 samples) and the HER2-enriched subtype (58 samples) share only 10 samples and have substantially distinct clinical features in terms of the three important receptors' status.
In particular, the $\widehat{\mb{\Delta}}_{\text{mRNA}}$-3 cluster primarily includes
those samples that are ER+ and/or PR+, whereas
the HER2-enriched subtype contains those that are HER2+ and/or PR$-$.
To conclude,
the $\widehat{\mb{\Delta}}_{\text{mRNA}}$-3 cluster, with a low survival rate, is remarkably different from the four PAM50 subtypes and appears to be an important new breast cancer subtype worth further investigation.

\section{Discussion}\label{sec: discussion}
In this paper, we propose a new decomposition method, called CDPA, to extract
the common and distinctive patterns of two correlated datasets
by incorporating the conventionally ignored common and distinctive patterns between the two coefficient matrices of common latent factors.
We also develop a graph-matching based approach to match the unpaired rows between the coefficient matrices. 
Consistent CDPA matrix estimation is established under high-dimensional settings 
and is supported by simulations. 
Our simulation studies and two real-data examples
show that CDPA
can better delineate the common and distinctive patterns 
between datasets
than D-CCA-type methods, 
thereby benefiting data mining applications.

There are two possible extensions of the CDPA. 
The first is to extend it to three or more datasets. 
One may construct a multi-set CDPA method by first developing a multi-set D-CCA
from the generalized CCA \citep{Kett71}.
The next challenge is how to appropriately match the rows of the multiple coefficient matrices of the resulting common latent factors. The second extension is to incorporate the nonlinear patterns between the two datasets. The CDPA only considers
the linear patterns extracted from the inner product spaces 
$(\mathcal{L}_0^2,\cov)$ and $(\mathbb{R}^{p_1\vee p_2},\cdot)$.
A nonlinear version of our row-matching approach and a nonlinear D-CCA may 
be expected in this extension, where the latter is possibly developed from the kernel CCA \citep{Fuku07} or the deep CCA \citep{Andr13}.

\appendix

\renewcommand\theequation{A\arabic{equation}}
\setcounter{equation}{0}

\renewcommand\thetable{A\arabic{table}}
\setcounter{table}{0}

\renewcommand\thefigure{A\arabic{figure}}
\setcounter{figure}{0}

\renewcommand\thesection{A\arabic{section}}
\setcounter{section}{0}

\section{Theoretical Proofs}\label{sec: tech proofs}
\subsection{Proof of Theorem~\ref{thm: uniq}}
For $k=1,2$,
denote $\bd{z}_k^{[1:r_{12}]}$ and $\tilde{\bd{z}}_k^{[1:r_{12}]}$
to be the vectors containing two different sets of the first $r_{12}$ canonical variables associated with $\bd{x}_k$.
By the first paragraph of page 5 in the supplement of \citep{Shu18},
there exists an orthogonal matrix $\mb{O}_{zk}$ such that $\tilde{\bd{z}}_k^{[1:r_{12}]}=\mb{O}_{zk}\bd{z}_k^{[1:r_{12}]}$.
Let $\mb{B}_k=\cov(\bd{x}_k, \bd{z}_k^{[1:r_{12}]})$ and $\widetilde{\mb{B}}_k=\cov(\bd{x}_k, \tilde{\bd{z}}_k^{[1:r_{12}]})$.
We have $\widetilde{\mb{B}}_k=\cov(\bd{x}_k, \bd{z}_k^{[1:r_{12}]})\mb{O}_{zk}^\top=
\mb{B}_k\mb{O}_{zk}^\top$.
Thus, $\colsp(\widetilde{\mb{B}}_k)=\colsp(\mb{B}_k)$.
Define $\widetilde{\mb{B}}_{2A}=[\widetilde{\mb{B}}_2;\mb{0}_{ (p_1-p_2)\times r_{12}}]$.
We still have $\colsp(\mb{P}\widetilde{\mb{B}}_{2A})=\colsp(\mb{P}\mb{B}_{2A})$.
For $\ell=1,\dots, r_{12}$,
recall that 
$\mb{V}_{B_1}^{[:,\ell]}$ and $\mb{V}_{B_2}^{[:,\ell]}$ are 
the $\ell$-th pair of principal vectors of
$\colsp(\mb{B}_1)$ and $\colsp(\mb{P}\mb{B}_{2A})$.
Let $\{\widetilde{\mb{V}}_{B_k}\}_{k=1}^2$ be the matrices whose columns  
$\{\widetilde{\mb{V}}_{B_1}^{[:,\ell]},\widetilde{\mb{V}}_{B_2}^{[:,\ell]}\}_{\ell=1}^{r_{12}}$ are another set of principal vectors
of $\colsp(\mb{B}_1)$ and $\colsp(\mb{P}\mb{B}_{2A})$ with $ \theta(\widetilde{\mb{V}}_{B_1}^{[:,\ell]},\widetilde{\mb{V}}_{B_2}^{[:,\ell]})=\theta_{B\ell}$.
There exist orthogonal matrices $\{\mb{O}_{V_k}\}_{k=1}^2$
such that
$\widetilde{\mb{V}}_{B_k}=\mb{V}_{B_k}\mb{O}_{V_k}$.
Let $\mb{\Lambda}_B=\diag(\cos\theta_{B1},\dots,\cos\theta_{Br_{12}})$.
Note that
$
\mb{\Lambda}_B=\widetilde{\mb{V}}_{B_1}^\top \widetilde{\mb{V}}_{B_2}
=\mb{O}_{V_1}^\top \mb{V}_{B_1}^\top \mb{V}_{B_2} \mb{O}_{V_2}
=\mb{O}_{V_1}^\top\mb{\Lambda}_B\mb{O}_{V_2}.
$
Then,
$\mb{O}_{V_k}=\diag(\mb{M}_{k,1},\dots,$
$\mb{M}_{k,m},\mb{M}_{k,m+1})$,
where $\mb{M}_{k,\ell},\ell\le m$ is an orthogonal matrix with column 
dimension equal to the repetition number of the $\ell$-th
largest distinct nonzero singular value of $\mb{\Lambda}_B$,
and $\mb{M}_{k,m+1}$ might be an empty matrix.
By $\mb{O}_{V_1}\mb{\Lambda}_B=\mb{\Lambda}_B \mb{O}_{V_2}$,
we obtain $\mb{M}_{1,\ell}=\mb{M}_{2,\ell}$ for all $\ell \le m$.
Define $r_{\lambda}=\rank(\mb{\Lambda}_B)$, 
\[
\tilde{\bd{c}}_{B\ell}=
\frac{1}{2}\left(1-\sqrt{\frac{1-\cos\theta_{B\ell}}{1+\cos\theta_{B\ell}  }}\right) \left(\widetilde{\mb{V}}_{B_1}^{[:,\ell]}+\widetilde{\mb{V}}_{B_2}^{[:,\ell]}\right),
\]
and $\mb{A}_B=\diag(a_{B1},\dots,a_{Br_{12}})$ with 
$a_{B\ell}=\frac{1}{2}\Big[1-\big(\frac{1-\cos\theta_{B\ell}}{1+\cos\theta_{B\ell}}     \big)^{1/2}\Big]$ for $\ell\le r_{12}$.
Note that
\begin{align*}
	[\tilde{\bd{c}}_{B\ell}]_{\ell=1}^{r_{12}}\widetilde{\mb{V}}_{B_k}^\top
	&=[\tilde{\bd{c}}_{B\ell}]_{\ell=1}^{r_\lambda}(\widetilde{\mb{V}}_{B_k}^{[:,1:r_\lambda]})^\top\\
	&=(\widetilde{\mb{V}}_{B_1}^{[:,1:r_\lambda]}+\widetilde{\mb{V}}_{B_2}^{[:,1:r_\lambda]})\mb{A}_B^{[1:r_\lambda,1:r_\lambda]}(\widetilde{\mb{V}}_{B_k}^{[:,1:r_\lambda]})^\top\\
	&=(\mb{V}_{B_1}^{[:,1:r_\lambda]}+\mb{V}_{B_2}^{[:,1:r_\lambda]})
	\diag(\mb{M}_{1,1},\dots,\mb{M}_{1,m})\mb{A}_B^{[1:r_\lambda,1:r_\lambda]}\\
	&\qquad\cdot
	[\diag(\mb{M}_{1,1},\dots,\mb{M}_{1,m})]^\top (\mb{V}_{B_k}^{[:,1:r_\lambda]})^\top\\
	&=(\mb{V}_{B_1}^{[:,1:r_\lambda]}+\mb{V}_{B_2}^{[:,1:r_\lambda]})
	\mb{A}_B^{[1:r_\lambda,1:r_\lambda]}(\mb{V}_{B_k}^{[:,1:r_\lambda]})^\top\\
	&=[\bd{c}_{B\ell}]_{\ell=1}^{r_\lambda}(\mb{V}_{B_k}^{[:,1:r_\lambda]})^\top\\
	&=[\bd{c}_{B\ell}]_{\ell=1}^{r_{12}}\mb{V}_{B_k}^\top.
\end{align*}
Hence, $[\bd{c}_{B\ell}]_{\ell=1}^{r_{12}}\mb{V}_{B_k}^\top$ is unique for $k=1,2$.
By Theorem~2 in \citep{Shu18}, we have that $\bd{c}_k$ in~\eqref{X=betaC+betaD} is unique for $k=1,2$.
Then by $\mb{B}_1 ([c_\ell]_{\ell=1}^{r_{12}})^\top=\bd{c}_1$
and $\mb{P}\mb{B}_{2A} ([c_\ell]_{\ell=1}^{r_{12}})^\top=\mb{P}(\bd{c}_2^\top, \bd{0}_{1\times (p_1-p_2)})^\top$, we have that
both $\mb{B}_1 ([c_\ell]_{\ell=1}^{r_{12}})^\top$
and $\mb{P}\mb{B}_{2A} ([c_\ell]_{\ell=1}^{r_{12}})^\top$
are unique.
Then by the definition in \eqref{c1*}, we obtain the uniqueness of
$\bd{c}_k^*$ for $k=1,2$.
Hence, 
$\bd{c}=\frac{1}{2}\sum_{k=1}^2[\tr(\mb{\Sigma}_k)]^{-1/2}\bd{c}_k^*$
is unique.

\subsection{Proof of Theorem~\ref{thm: CC}}
Let $\tilde{r}_k=\rank(\widehat{\mb{X}}_k)$.
From (S.17) in \citep{Shu18}, we have $\tilde{r}_k=r_k$
with probability tending to 1 as $n\to \infty$.
Due to Lemma~S.1 in \citep{Shu18}, we simply 
assume $\tilde{r}_k=r_k$ in the rest of the proof.
Thus, $\widehat{\mb{\Lambda}}_k$ is rank-$r_k$, and then
$\widehat{\mb{B}}_k=\widehat{\mb{V}}_k\widehat{\mb{\Lambda}}_k^{1/2}\widehat{\mb{U}}_{\theta k}^{[:,1:r_{12}]}$ 
is rank-$r_{12}$.

From (S.7) of \citep{Shu18}, we have $\lambda_1(\mb{\Sigma}_k)\asymp \lambda_{r_k}(\mb{\Sigma}_k)$.
By Weyl’s inequality \citep[][Theorem 3.3.16(a)]{Horn94}
and Assumption~\ref{assump1} (I) and (V),
$\kappa_1\le
\lambda_{k, p_k}=	\lambda_{k,(r_k+1)+(p_k-r_k)-1}-\lambda_{r_k+1}(\cov(\bd{x}_k))
\le \lambda_{p_k-r_k}(\cov(\bd{e}_k))
\le 
\lambda_1(\cov(\bd{e}_k))=\| \cov(\bd{e}_k)\|_2\le \| \cov(\bd{e}_k)\|_\infty\le s_0$.
Thus, 
\[
\frac{\lambda_1(\cov(\bd{x}_k))}{p_k}\asymp\frac{\tr(\cov(\bd{x}_k))}{\tr(\cov(\bd{e}_k))}=\SNR_k.
\]

Let $\widetilde{\mb{Q}}_k\in \mathbb{R}^{p_k\times r_{12}}$ be the left singular matrix of $\mb{B}_k$.
Note that
$\|\mb{B}_k\|_2
\le \| \mb{V}_k\mb{\Lambda}_k^{1/2}   \|_2 \| \mb{U}_{\theta k}^{[:,1:r_{12}]} \|_2
= \lambda_1^{1/2}(\mb{\Sigma}_k)$.
By (S.31) in \citep{Shu18}, we have
\be\label{diff in B_k}
\| \widehat{\mb{B}}_k-\mb{B}_k  \|_2=O_P(\lambda_1^{1/2}(\mb{\Sigma}_k)\delta_\theta).
\ee
Thus, $\|\widehat{\mb{B}}_k \|_2\le \| \widehat{\mb{B}}_k-\mb{B}_k  \|_2 +\|\mb{B}_k  \|_2=O_P(\lambda_1^{1/2}(\mb{\Sigma}_k))$.
By Lemma~1 of \citep{Lam09} and then Theorem~3 of \citep{Yu15}, there exists an orthogonal matrix $\mb{O}_{k}$ such that
\begin{align}\label{diff in Q_k}
\|\widehat{\mb{Q}}_k-\widetilde{\mb{Q}}_k\mb{O}_{k}\|_F
&\le
\|\widehat{\mb{Q}}_k\mb{O}_{k}^\top-\widetilde{\mb{Q}}_k\|_F\| \mb{O}_{k} \|_2\nonumber\\
&\lesssim_P  \lambda_1^{1/2}(\mb{\Sigma}_k)\|\widehat{\mb{B}}_k-\mb{B}_k\|_2/\lambda_1(\mb{\Sigma}_k)
\lesssim_P \delta_\theta.
\end{align}
Here and in the following text, we write $A\lesssim_P B$ if and only if $A=O_P(B)$.
Note that for any real matrices $\mb{M}_1$ and $\mb{M}_2$, we have
\be\label{(AB)_hat-AB in spectral norm}
\| \widehat{\mb{M}}_1\widehat{\mb{M}}_2-\mb{M}_1\mb{M}_2 \|_2
\le
\begin{cases}
	\| \widehat{\mb{M}}_1 \|_2\|\widehat{\mb{M}}_2- \mb{M}_2    \|_2
	+\| \mb{M}_2 \|_2\|\widehat{\mb{M}}_1- \mb{M}_1    \|_2,\\
	\| \mb{M}_1 \|_2\|\widehat{\mb{M}}_2- \mb{M}_2    \|_2
	+\| \widehat{\mb{M}}_2 \|_2\|\widehat{\mb{M}}_1- \mb{M}_1    \|_2,
\end{cases}
\ee
and
\be\label{(AB)_hat-AB in Frobenius norm}
\| \widehat{\mb{M}}_1\widehat{\mb{M}}_2-\mb{M}_1\mb{M}_2 \|_F
\le
\begin{cases}
	\| \widehat{\mb{M}}_1 \|_2\|\widehat{\mb{M}}_2- \mb{M}_2    \|_F
	+\| \mb{M}_2 \|_2\|\widehat{\mb{M}}_1- \mb{M}_1    \|_F,\\
	\| \mb{M}_1 \|_2\|\widehat{\mb{M}}_2- \mb{M}_2    \|_F
	+\| \widehat{\mb{M}}_2 \|_2\|\widehat{\mb{M}}_1- \mb{M}_1    \|_F.
\end{cases}
\ee
Let
$\mb{Q}_k=\widetilde{\mb{Q}}_k\mb{O}_k$ and $\mb{Q}_{2A}=[\mb{Q}_2;\mb{0}_{(p_1-p_2)\times r_{12}}]$. Note that the columns of $\mb{Q}_k$ form an orthonormal basis of $\colsp(\mb{B}_k)$, and those of $\mb{P}\mb{Q}_{2A}$ also form an orthonormal basis of $\colsp(\mb{P}\mb{B}_{2A})$. Let $\mb{\Theta}_B=\mb{Q}_1^\top\mb{P}\mb{Q}_{2A}$.
Then by \eqref{(AB)_hat-AB in Frobenius norm} and \eqref{diff in Q_k}, we have
\begin{align*}
	\|  \widehat{\mb{\Theta}}_B-\mb{\Theta}_B\|_F
	&\le \|  \widehat{\mb{Q}}_1^\top \|_2 \| \mb{P} \widehat{\mb{Q}}_{2A}- \mb{P} \mb{Q}_{2A}  \|_F
	+
	\|  \mb{P} \mb{Q}_{2A} \|_2\|  \widehat{\mb{Q}}_1^\top-\mb{Q}_1^\top \|_F\\
	&\le \|  \widehat{\mb{Q}}_1^\top \|_2\|\mb{P} \|_2 \| \widehat{\mb{Q}}_{2A}- \mb{Q}_{2A}  \|_F
	+
	\|  \mb{P}\|_2\| \mb{Q}_{2A} \|_2\|  \widehat{\mb{Q}}_1^\top-\mb{Q}_1^\top \|_F\\
	&= \|  \widehat{\mb{Q}}_2-\mb{Q}_2 \|_F+\|  \widehat{\mb{Q}}_1-\mb{Q}_1\|_F
	\\
	&\lesssim_P\delta_\theta,
\end{align*}
and
\begin{align*}
	&\max\{\| \widehat{\mb{\Theta}}_B\widehat{\mb{\Theta}}_B^\top -\mb{\Theta}_B\mb{\Theta}_B^\top \|_F,
	\| \widehat{\mb{\Theta}}_B^\top\widehat{\mb{\Theta}}_B -\mb{\Theta}_B^\top\mb{\Theta}_B \|_F
	\}\\
	&\quad\le (\|\widehat{\mb{\Theta}}_B\|_2+\|\mb{\Theta}_B\|_2)\|  \widehat{\mb{\Theta}}_B-\mb{\Theta}_B\|_F\le 2\|\widehat{\mb{\Theta}}_B-\mb{\Theta}_B\|_F
	\lesssim_P\delta_\theta.
\end{align*}
By Weyl's inequality (see Theorem 3.3.16(c) in \citep{Horn94}),
\be\label{diff in sig of Theta_B}
\max_{1\le \ell\le r_{12}}
| \sigma_\ell(\widehat{\mb{\Theta}}_B)- \sigma_\ell(\mb{\Theta}_B)|
\le \|\widehat{\mb{\Theta}}_B-\mb{\Theta}_B   \|_2
\le \|\widehat{\mb{\Theta}}_B-\mb{\Theta}_B   \|_F
\lesssim_P  \delta_\theta.
\ee
Denote $\{\widetilde{\mb{U}}_{B_k}
\}_{k=1}^2$ to be one pair of orthogonal matrices such that $\mb{\Theta}_B=\widetilde{\mb{U}}_{B_1}\mb{\Lambda}_B\widetilde{\mb{U}}_{B_2}^\top$.
Let
$\sigma_{B,1}>\dots>\sigma_{B,r_B}$ be the distinct singular values of $\mb{\Theta}_B$, 
and define $\sigma_{B,r_{12}+1}^2=-\infty$.
By Lemma~1 of \citep{Lam09} and then Theorem~2 of \citep{Yu15}, there exists
a matrix $\mb{O}_{B_k}=\diag(\mb{O}_{B_k,1},\dots,\mb{O}_{B_k,r_B})$, 
where $\mb{O}_{B_k,\ell}$ is an orthogonal matrix with column dimension equal to the repetition number of $\sigma_{B, \ell}$, such that 
\begin{align}\label{U-UO}
	\| \widehat{\mb{U}}_{B_k}-\widetilde{\mb{U}}_{B_k}\mb{O}_{B_k} \|_F
	&	\le 
	\| \widehat{\mb{U}}_{B_k}\mb{O}_{B_k}^\top-\widetilde{\mb{U}}_{B_k} \|_F\| \mb{O}_{B_k} \|_2\nonumber\\
	&	\lesssim_P  \min\left\{\delta_\theta \Big/\min_{1\le \ell\le r_B}\{\sigma^2_{B,\ell}-\sigma^2_{B,\ell+1}\},1\right\}
	\lesssim_P \delta_\theta.
\end{align}
We	define $\widetilde{\mb{O}}_{B_2}=\diag(\mb{O}_{B_1,1},\dots,\mb{O}_{B_1,r_B-1},\mb{O}_{B_1,r_B})$
if $\sigma_{B,r_B}\ne 0$, and otherwise let 
$\widetilde{\mb{O}}_{B_2}
=\diag(\mb{O}_{B_1,1},\dots,\mb{O}_{B_1,r_B-1},\mb{O}_{B_2,r_B})$.
Let $\mb{U}_{B_1}=\widetilde{\mb{U}}_{B_1} \mb{O}_{B_1}$ and $\mb{U}_{B_2}=\widetilde{\mb{U}}_{B_2} \widetilde{\mb{O}}_{B_2}$.
We have	
$\mb{U}_{B_1}\mb{\Lambda}_B\mb{U}_{B_2}=\widetilde{\mb{U}}_{B_1} \mb{O}_{B_1}\mb{\Lambda}_B\widetilde{\mb{O}}_{B_2}^\top\widetilde{\mb{U}}_{B_2}^\top
=\widetilde{\mb{U}}_{B_1} \mb{\Lambda}_B\widetilde{\mb{U}}_{B_2}^\top=\mb{\Theta}_B$.
Define $\mb{U}^\star_{B_2}=\widetilde{\mb{U}}_{B_2}\mb{O}_{B_2}$ and $r_{\theta_B}=\rank(\mb{\Theta}_B)$.
Then, 
\be\label{U_B2=U_B2*}
\mb{U}_{B_2}^{[:,(r_{\theta_B}+1):r_{12}]}=\mb{U}_{B_2}^{\star [:,(r_{\theta_B}+1):r_{12}]}\quad\text{if}\quad r_{\theta_B}<r_{12},
\ee
\be\label{err bd for U_B1}
\| \widehat{\mb{U}}_{B_1}-\mb{U}_{B_1} \|_F 
\lesssim_P 
\delta_\theta,
\ee
and
\be\label{err bd for U*_B2}
\| \widehat{\mb{U}}_{B_2}-\mb{U}_{B_2}^{\star} \|_F 
\lesssim_P 
\delta_\theta.
\ee
By \eqref{(AB)_hat-AB in spectral norm}, \eqref{diff in sig of Theta_B} and the above two inequalities,
\begin{align*}
	\lefteqn{\left\|\widehat{\mb{U}}_{B_1}\widehat{\mb{\Lambda}}_B \widehat{\mb{U}}_{B_2}^\top -  \mb{U}_{B_1}\mb{\Lambda}_{B} \mb{U}_{B_2}^{\star\top} \right\|_2}\\
	&\le\| \widehat{\mb{U}}_{B_1}\widehat{\mb{\Lambda}}_B-  \mb{U}_{B_1}\mb{\Lambda}_{B}  \|_2\| \widehat{\mb{U}}_{B_2}^\top\|_2
	+\|\mb{U}_{B_1}\mb{\Lambda}_{B}   \|_2\| \widehat{\mb{U}}_{B_2}^\top  -\mb{U}_{B_2}^{\star\top}\|_2\\
	&\le \| \widehat{\mb{U}}_{B_1}-\mb{U}_{B_1} \|_2\| \mb{\Lambda}_{B}  \|_2
	+\|\widehat{\mb{U}}_{B_1}   \|_2\|\widehat{\mb{\Lambda}}_B-\mb{\Lambda}_{B}  \|_2
	+ \| \mb{\Lambda}_{B}   \|_2\| \widehat{\mb{U}}_{B_2}^\top  -\mb{U}_{B_2}^{\star\top}\|_2\\
	&\lesssim_P  
	\delta_\theta.
\end{align*}
By the above inequality, $\| \widehat{\mb{\Theta}}_B-\mb{\Theta}_B   \|_2\lesssim_P \delta_\theta$, and the triangular inequality of matrix norms, we have
\[
\| \mb{U}_{B_1}\mb{\Lambda}_B (\mb{U}_{B_2}-\mb{U}_{B_2}^{\star})^\top \|_2
\lesssim_P   \delta_\theta.
\]
It follows that
\begin{align}\label{D_B(U_B-U_B)}
	\|\mb{U}_{B_2}^{[:,1:r_{\theta_B}]}-\mb{U}_{B_2}^{\star [:,1:r_{\theta_B}]}  \|_F
	&
	\le\sqrt{r_{12}}\|\mb{U}_{B_2}^{[:,1:r_{\theta_B}]}-\mb{U}_{B_2}^{\star[:,1:r_{\theta_B}]}  \|_2\nonumber\\
	&\le\sqrt{r_{12}}\left\|\mb{\Lambda}_B^\dag \right\|_2\left\| \mb{U}_{B_1}^\top \right\|_2\left\| \mb{U}_{B_1}\mb{\Lambda}_{B} (\mb{U}_{B_2}-\mb{U}_{B_2}^{\star})^\top \right\|_2\nonumber\\
	&\lesssim_P 
	\delta_\theta.
\end{align}
Combining  \eqref{D_B(U_B-U_B)}, \eqref{U_B2=U_B2*} and \eqref{err bd for U*_B2} yields
\be\label{err bd for U_B2}
\| \widehat{\mb{U}}_{B_2}-\mb{U}_{B_2} \|_F 
\lesssim_P 
\delta_\theta.
\ee
By \eqref{PA,PV}, we have that the $\ell$-th columns of
$\mb{V}_{B_1}:=\mb{Q}_1\mb{U}_{B_1}$
and 
$\mb{V}_{B_2}:=\mb{P}\mb{Q}_{2A}\mb{U}_{B_2}$
are the $\ell$-th pair of principal vectors
of $\colsp(\mb{B}_1)$ and $\colsp(\mb{P}\mb{B}_{2A})$.
By \eqref{(AB)_hat-AB in Frobenius norm}, \eqref{diff in Q_k} and \eqref{err bd for U_B1}, 
we have
\begin{align}\label{diff in V_B1}
	\|\widehat{\mb{V}}_{B_1}-\mb{V}_{B_1}  \|_F
	&=\|\mb{Q}_1\mb{U}_{B_1}- \widehat{\mb{Q}}_1\widehat{\mb{U}}_{B_1} \|_2
	\nonumber\\
	&\le 
	\|\widehat{\mb{U}}_{B_1} \|_2
	\|  \widehat{\mb{Q}}_1-\mb{Q}_1\|_F
	+
	\|\mb{Q}_1 \|_2
	\| \widehat{\mb{U}}_{B_1}-\mb{U}_{B_1}\|_F
	\nonumber\\
	&\lesssim_P 
	\delta_\theta.
\end{align}
Similarly, by \eqref{err bd for U_B2} we
obtain
\be\label{diff in V_B2}
\|\widehat{\mb{V}}_{B_2}-\mb{V}_{B_2}  \|_F\lesssim_P 
\delta_\theta.
\ee
Then, together with \eqref{(AB)_hat-AB in Frobenius norm} and \eqref{diff in B_k},
we have
\begin{align}\label{diff in VB1}
\| \widehat{\mb{V}}_{B_1}^\top \widehat{\mb{B}}_1-\mb{V}_{B_1}^\top \mb{B}_1\|_F
&\le \|\widehat{\mb{B}}_1 \|_2\|\widehat{\mb{V}}_{B_1}^\top-\mb{V}_{B_1}^\top\|_F
+\|\mb{V}_{B_1}^\top\|_2 \|\widehat{\mb{B}}_1-\mb{B}_1\|_F\nonumber\\
&\lesssim_P  \lambda_1^{1/2}(\mb{\Sigma}_1)\delta_\theta,
\end{align}
and similarly,
\be\label{diff in VB2}
\| \widehat{\mb{V}}_{B_2}^\top\mb{P} \widehat{\mb{B}}_{2A}-\mb{V}_{B_2}^\top\mb{P} \mb{B}_{2A}\|_F
\lesssim_P  \lambda_1^{1/2}(\mb{\Sigma}_2)\delta_\theta.
\ee
By the results given in (S.16), (S.17) and (S.7) of \citep{Shu18}, we have 
$|\lambda_\ell(\widehat{\mb{\Sigma}}_k)-\lambda_\ell(\mb{\Sigma}_k)|\lesssim_P\lambda_1(\mb{\Sigma}_k)/\sqrt{n}$ for all $\ell\le r_k$, 
$[\tr(\widehat{\mb{\Sigma}}_k)]^{1/2}
=[\sum_{\ell=1}^{r_k}\lambda_\ell(\widehat{\mb{\Sigma}}_k)]^{1/2}
\ge [r_k(1-o_P(1))\lambda_{r_k}(\mb{\Sigma}_k)]^{1/2}$, and $\lambda_1(\mb{\Sigma}_k)\asymp \lambda_{r_k}(\mb{\Sigma}_k)$.
Then by the mean value theorem, we obtain
\begin{align}
\lefteqn{	\big| [\tr(\widehat{\mb{\Sigma}}_k)]^{1/2}  -[\tr(\mb{\Sigma}_k)]^{1/2}  \big|}\nonumber\\
	&\le\frac{1}{2}\big|\tr(\widehat{\mb{\Sigma}}_k) -\tr(\mb{\Sigma}_k)  \big|
	\cdot\max\big\{[\tr(\widehat{\mb{\Sigma}}_k)]^{-1/2},[\tr(\mb{\Sigma}_k)]^{-1/2}\big\}\nonumber\\
	&\lesssim_P\lambda_1^{1/2}(\mb{\Sigma}_k)/\sqrt{n}.\label{diff in tr_cov_x}
\end{align}
Hence, 
\begin{align}\label{diff in B_kz}
\lefteqn{	\big| [\tr(\widehat{\mb{\Sigma}}_k)]^{-1/2}  -[\tr(\mb{\Sigma}_k)]^{-1/2} \big|}\nonumber\\
	&=\big| [\tr(\widehat{\mb{\Sigma}}_k)]^{1/2}  -[\tr(\mb{\Sigma}_k)]^{1/2} \big|\Big/
	\big([\tr(\widehat{\mb{\Sigma}}_k)]^{1/2}[\tr(\mb{\Sigma}_k)]^{1/2}\big)\nonumber\\
	&\lesssim_P\lambda_1^{-1/2}(\mb{\Sigma}_k)/\sqrt{n}.
\end{align}
By \eqref{(AB)_hat-AB in Frobenius norm}, 
\eqref{diff in VB1} and \eqref{diff in B_kz},
\begin{align*}
\lefteqn{	\left\|\widehat{\mb{V}}_{B_1}^\top \widehat{\mb{B}}_1[\tr(\widehat{\mb{\Sigma}}_1)]^{-1/2} 
	-\mb{V}_{B_1}^\top \mb{B}_1[\tr(\mb{\Sigma}_1)]^{-1/2} \right\|_F}\\
	&	\lesssim_P \lambda_1^{-1/2}(\mb{\Sigma}_1) ( \lambda_1^{1/2}(\mb{\Sigma}_1)\delta_\theta)+\lambda_1^{1/2}(\mb{\Sigma}_1)\lambda_1^{-1/2}(\mb{\Sigma}_1)/\sqrt{n}\nonumber\\
	&\lesssim_P \delta_\theta.
\end{align*}
Similarly, by \eqref{diff in VB2},
$
\big\|\widehat{\mb{V}}_{B_2}^\top\mb{P} \widehat{\mb{B}}_{2A}[\tr(\widehat{\mb{\Sigma}}_2)]^{-1/2}  
-\mb{V}_{B_2}^\top \mb{P} \mb{B}_{2A}[\tr(\mb{\Sigma}_2)]^{-1/2}  \big\|_F
\lesssim_P \delta_\theta.
$
Thus, 
\begin{align}\label{diff in S*}
	&\Big\|
	\big(\widehat{\mb{V}}_{B_1}^\top \widehat{\mb{B}}_1[\tr(\widehat{\mb{\Sigma}}_1)]^{-1/2} 	+\widehat{\mb{V}}_{B_2}^\top\mb{P} \widehat{\mb{B}}_{2A}[\tr(\widehat{\mb{\Sigma}}_2)]^{-1/2} \big)\nonumber\\
	&\qquad	-\big(\mb{V}_{B_1}^\top \mb{B}_1[\tr(\mb{\Sigma}_1)]^{-1/2} 
	+\mb{V}_{B_2}^\top \mb{P} \mb{B}_{2A}[\tr(\mb{\Sigma}_2)]^{-1/2} 
	\big)\Big\|_F 
	\lesssim_P \delta_\theta.
\end{align}

Define $\mb{C}_B=[\bd{c}_{B\ell}]_{\ell=1}^{r_{12}}$ and
$\mb{A}_B=\diag(a_{B1},\dots,a_{Br_{12}})$ with 
$a_{B\ell}=\frac{1}{2}\Big[1-\big(\frac{1-\mb{\Lambda}_B^{[\ell,\ell]}}{1+\mb{\Lambda}_B^{[\ell,\ell]}}     \big)^{1/2}\Big]$.
We have
$
\mb{C}_B=(\mb{V}_{B_1}+\mb{V}_{B_2})\mb{A}_B.
$
By the same technique used to derive (S.32) in \citep{Shu18},
we have
$
\|  \widehat{\mb{A}}_B-\mb{A}_B \|_F\lesssim_P  \delta_\theta^{1/2}.
$
From \eqref{diff in V_B1} and \eqref{diff in V_B2},
$
\|(\widehat{\mb{V}}_{B_1}+\widehat{\mb{V}}_{B_2})-(\mb{V}_{B_1}+\mb{V}_{B_2})\|_F
\lesssim_P \delta_\theta.
$
Then by \eqref{(AB)_hat-AB in Frobenius norm},
\be\label{diff in C_B}
\|\widehat{\mb{C}}_B-
\mb{C}_B\|_F
\lesssim_P \delta_\theta^{1/2}+ \delta_\theta
\lesssim_P  \delta_\theta^{1/2}.
\ee
From (S.23) in \citep{Shu18}, $\|\widehat{\mb{\Theta}}-\mb{\Theta}\|_F\lesssim \delta_\theta$.
Using the same proof technique for \eqref{err bd for U_B1} and \eqref{err bd for U_B2},
we have $\| \widehat{\mb{U}}_{\theta k}^{[:,1:r_{12}]}-\mb{U}_{\theta k}^{[:,1:r_{12}]} \|_F 
\lesssim_P 
\delta_\theta$.
Then following the same proof lines for (S.28) in \citep{Shu18},
we can obtain
\[
\left\| (\widehat{\mb{U}}_{\theta k}^{[:,1:r_{12}]})^\top \widehat{\mb{\Lambda}}_k^{-1/2} \widehat{\mb{V}}_k^\top-
({\mb{U}}_{\theta k}^{[:,1:r_{12}]})^\top {\mb{\Lambda}}_k^{-1/2} {\mb{V}}_k^\top \right\|_F\lesssim_P \lambda_1^{-1/2}(\mb{\Sigma}_k)\delta_\theta.
\]
From the results given in (S.9), (S.13), (S.15) and (S.32) of \citep{Shu18}, we have that
$\max\{\|\widehat{\mb{X}}_k\|_F, \|{\mb{X}}_k\|_F\}  \lesssim_P \sqrt{n\lambda_1(\mb{\Sigma}_k)}$,
$
\|\widehat{\mb{X}}_k-\mb{X}_k \|_F\lesssim_P \min\{
\sqrt{\lambda_1(\mb{\Sigma}_k)/n}+\sqrt{p_k\log p_k},\sqrt{n\lambda_1(\mb{\Sigma}_k)}
\}
$,
and
$
\|  \widehat{\mb{A}}_C-\mb{A}_C \|_F\lesssim_P  \delta_\theta^{1/2}
$, 
where $\mb{A}_C=\diag(a_1,\dots,a_{r_{12}})$ with
$a_\ell=\frac{1}{2}\big[1-(\frac{1-\sigma_\ell(\mb{\Theta})}{1+\sigma_\ell(\mb{\Theta})})^{1/2}\big]$.
Let $\mb{Z}_k=	{\mb{U}}_{\theta k}^\top {\mb{\Lambda}}_k^{-1/2} {\mb{V}}_k^\top\mb{X}_k$
and
$\mb{C}_0=\mb{A}_C\sum_{j=1}^2\mb{Z}_j^{[1:r_{12},:]}$, which are the sample matrices of $\bd{z}_k$ and $(c_1,\dots,c_{r_{12}})^\top$, respectively.
Then by \eqref{(AB)_hat-AB in Frobenius norm},
\begin{align*}
	&\left\|\widehat{\mb{Z}}_k^{[1:r_{12},:]}
	-{\mb{Z}}_k^{[1:r_{12},:]}\right\|_F\\
	&=
	\left\|
	(\widehat{\mb{U}}_{\theta k}^{[:,1:r_{12}]})^\top \widehat{\mb{\Lambda}}_k^{-1/2} \widehat{\mb{V}}_k^\top\widehat{\mb{X}}_k-
	({\mb{U}}_{\theta k}^{[:,1:r_{12}]})^\top {\mb{\Lambda}}_k^{-1/2} {\mb{V}}_k^\top\mb{X}_k
	\right\|_F\\
	&\lesssim_P \lambda_1^{-1/2}(\mb{\Sigma}_k)\delta_\theta\sqrt{n\lambda_1(\mb{\Sigma}_k)}\\
&\qquad	+ \min\{
	\sqrt{\lambda_1(\mb{\Sigma}_k)/n}+\sqrt{p_k\log p_k},\sqrt{n\lambda_1(\mb{\Sigma}_k)}\}/\sqrt{\lambda_1(\mb{\Sigma}_k)}\\
	&\lesssim_P \delta_\theta\sqrt{n},
\end{align*}
and thus,
\be\label{diff in C_0}
\|\widehat{\mb{C}}_0-\mb{C}_0\|_F\lesssim_P  \delta_\theta^{1/2}\sqrt{n}.
\ee
From \eqref{diff in S*}, \eqref{diff in C_B}, \eqref{diff in C_0} and \eqref{(AB)_hat-AB in Frobenius norm},
we obtain
\be\label{diff in C}
\|  \widehat{\mb{C}}-   \mb{C}\|_2\le \|  \widehat{\mb{C}}-   \mb{C}\|_F=O_P( \delta_\theta^{1/2}\sqrt{n}).
\ee
Combining \eqref{diff in tr_cov_x} and \eqref{diff in C} yields
\[
\|  \widehat{\mb{C}}^{(k)}-   \mb{C}^{(k)}\|_2\le \|  \widehat{\mb{C}}^{(k)}-   \mb{C}^{(k)}\|_F\lesssim_P \delta_\theta^{1/2}\sqrt{n}\cdot\lambda_1^{1/2}(\mb{\Sigma}_k).
\]	
By (S.14) in \citep{Shu18}, there exists a constant $\kappa_3\in (0,1]$
such that 
$
\| \mb{X}_k\|_F\ge \|\mb{X}_k \|_2\ge [\kappa_3+o_P(1)]\sqrt{n\lambda_1(\mb{\Sigma}_k)}.
$
Hence, \[
\frac{\|  \widehat{\mb{C}}-   \mb{C}\|_\star^2}{\frac{1}{2}(\| \mb{X}_1^S \|_\star^2+\| \mb{X}_2^S \|_\star^2 )}=O_P( \delta_\theta),
\]
and
\[
\frac{\|\widehat{\mb{C}}^{(k)}-\mb{C}^{(k)} \|_\star^2}{\|\mb{X}_k \|_\star^2}=O_P(\delta_\theta).
\]

Let $\bd{c}_0=(c_1,\dots,c_{r_{12}})^\top$ and $\bd{z}_c=[\bd{z}_1^{[1:r_{12}]};\bd{z}_2^{[1:r_{12}]}]$.
Define $\mb{Z}_c=[\mb{Z}_1^{[1:r_{12},:]};\mb{Z}_2^{[1:r_{12},:]}]$, which is the sample matrix of $\bd{z}_c$.
We have $\bd{c}_0=\mb{A}_C[\mb{I}_{r_{12}\times r_{12}},\mb{I}_{r_{12}\times r_{12}}]\bd{z}_c$
and $\mb{C}_0=\mb{A}_C[\mb{I}_{r_{12}\times r_{12}},\mb{I}_{r_{12}\times r_{12}}]\mb{Z}_c$.
From the central limit theorem,
\[
\left\|\frac{1}{n}\mb{Z}_c\mb{Z}_c^\top -\cov(\bd{z}_c)\right\|_F \le 2r_{12}
\left\| \frac{1}{n}\mb{Z}_c\mb{Z}_c^\top -\cov(\bd{z}_c)\right\|_{\max}
\lesssim_P n^{-1/2}.
\]
Hence,
\begin{align*}
\left\|\frac{1}{n}\mb{C}_0\mb{C}_0^\top-\cov(\bd{c}_0) \right\|_F
&\le \left\|\frac{1}{n}\mb{Z}_c\mb{Z}_c^\top -\cov(\bd{z}_c)\right\|_F
\Big\| \mb{A}_C[\mb{I}_{r_{12}\times r_{12}},\mb{I}_{r_{12}\times r_{12}}]\Big\|_F^2\\
&\lesssim_P n^{-1/2}.
\end{align*}
Then,
\[
\left\|  \frac{1}{n} \mb{C}\mb{C}^\top-\cov(\bd{c})    \right\|_F
\le \left\|\frac{1}{n}\mb{C}_0\mb{C}_0^\top-\cov(\bd{c}_0) \right\|_F
\|\mb{C}_B\mb{S}\|_F^2
\lesssim_P n^{-1/2}.
\]
By \eqref{diff in C}, we have
\[
\left\|\frac{1}{n}\widehat{\mb{C}}\widehat{\mb{C}}^\top- \frac{1}{n} \mb{C}\mb{C}^\top\right\|_F
\le \frac{1}{n} \| \widehat{\mb{C}}-\mb{C}\|_F(\|\widehat{\mb{C}}\|_F+\|\mb{C}\|_F)\lesssim_P\delta_\theta^{1/2}.
\]
Combining the above two inequalities yields
\[
\left\|  \frac{1}{n}\widehat{\mb{C}}\widehat{\mb{C}}^\top-\cov(\bd{c})    \right\|_F
\lesssim_P\delta_\theta^{1/2}.
\]
By Weyl's inequality (see Theorem 3.3.16(c) in \citep{Horn94}),
\begin{align*}
\lefteqn{\max_{\ell\le r_{12}}
\left|
\lambda_\ell(\frac{1}{n}\widehat{\mb{C}}\widehat{\mb{C}}^\top)
-\lambda_\ell(\cov(\bd{c}))
\right|}\\
&\le
\left\|  \frac{1}{n}\widehat{\mb{C}}\widehat{\mb{C}}^\top-\cov(\bd{c})    \right\|_2
\le
\left\|  \frac{1}{n}\widehat{\mb{C}}\widehat{\mb{C}}^\top-\cov(\bd{c})    \right\|_F\\
&\lesssim_P\delta_\theta^{1/2}.
\end{align*}
Then,
\[
\left|
\tr(\frac{1}{n}\widehat{\mb{C}}\widehat{\mb{C}}^\top)
-\tr(\cov(\bd{c}))
\right|
\le\sum_{\ell=1}^{r_{12}}
\left|
\lambda_\ell(\frac{1}{n}\widehat{\mb{C}}\widehat{\mb{C}}^\top)
-\lambda_\ell(\cov(\bd{c}))
\right|\lesssim_P\delta_\theta^{1/2}.
\]

The proof is complete.

\subsection{Proof of Theorem~\ref{thm: P converge}}

By \eqref{diff in Q_k}, 
there exists a matrix $\mb{Q}_k$, whose columns form an orthonormal basis
of $\colsp(\mb{B}_k)$, such that
$
\|\widehat{\mb{Q}}_k-\mb{Q}_k\|_F
=O_p(\delta_\theta).
$
Note that 
$\tr(\mb{Q}_1^\top \mb{P} \mb{Q}_{2A} (\mb{Q}_1^\top \mb{P} \mb{Q}_{2A})^\top)
=\|  \mb{Q}_1^\top \mb{P} \mb{Q}_{2A}  \|_F^2$.
Then by \eqref{(AB)_hat-AB in Frobenius norm},
for any $\mb{P}\in \Pi_{p_1}$, we have
\begin{align*}
	\lefteqn{\big|\| \mb{Q}_1^\top \mb{P} \mb{Q}_{2A} \|_F^2-\| \widehat{\mb{Q}}_1^\top \mb{P} \widehat{\mb{Q}}_{2A}  \|_F^2     \big|}\\
	&\le
	\big|\| \mb{Q}_1^\top \mb{P} \mb{Q}_{2A} \|_F-\| \widehat{\mb{Q}}_1^\top \mb{P} \widehat{\mb{Q}}_{2A}  \|_F     \big|
	(\| \mb{Q}_1^\top \mb{P} \mb{Q}_{2A} \|_F+\| \widehat{\mb{Q}}_1^\top \mb{P} \widehat{\mb{Q}}_{2A}  \|_F)\\
	&\le
	\|\mb{Q}_1^\top \mb{P} \mb{Q}_{2A}-\widehat{\mb{Q}}_1^\top \mb{P} \widehat{\mb{Q}}_{2A}\|_F
	(\| \mb{Q}_1^\top\mb{P}\|_F \|\mb{Q}_{2A} \|_F+\| \widehat{\mb{Q}}_1^\top \mb{P}\|_F \|\widehat{\mb{Q}}_{2A}  \|_F)\\
	&\le (\| \widehat{\mb{Q}}_1^\top \|_2 \| \mb{P} \mb{Q}_{2A}-  \mb{P} \widehat{\mb{Q}}_{2A}\|_F
	+\|  \mb{P} \mb{Q}_{2A}\|_2\|\mb{Q}_1^\top- \widehat{\mb{Q}}_1^\top \|_F)2r_k\\
	&=O_P(\delta_\theta).
\end{align*}
Hence, 
$\big|\| \mb{Q}_1^\top \mb{P}_* \mb{Q}_{2A} \|_F^2-\| \widehat{\mb{Q}}_1^\top \mb{P}_* \widehat{\mb{Q}}_{2A}  \|_F^2     \big|=O_P(\delta_\theta)=
\big|\| \mb{Q}_1^\top \widehat{\mb{P}}_* \mb{Q}_{2A} \|_F^2-\| \widehat{\mb{Q}}_1^\top \widehat{\mb{P}}_* \widehat{\mb{Q}}_{2A}  \|_F^2     \big|$.
Note that 
$
\| \mb{Q}_1^\top \widehat{\mb{P}}_* \mb{Q}_{2A} \|_F^2\le \| \mb{Q}_1^\top \mb{P}_* \mb{Q}_{2A} \|_F^2
$
and
$
\| \widehat{\mb{Q}}_1^\top \mb{P}_* \widehat{\mb{Q}}_{2A}  \|_F^2
\le \| \widehat{\mb{Q}}_1^\top \widehat{\mb{P}}_* \widehat{\mb{Q}}_{2A}  \|_F^2
$.
We have
\begin{align*}
	0&\le \| \mb{Q}_1^\top \mb{P}_* \mb{Q}_{2A} \|_F^2
	-\| \mb{Q}_1^\top \widehat{\mb{P}}_* \mb{Q}_{2A} \|_F^2\\
	&= 
	( \| \mb{Q}_1^\top \mb{P}_* \mb{Q}_{2A} \|_F^2
	-\| \widehat{\mb{Q}}_1^\top \mb{P}_* \widehat{\mb{Q}}_{2A} \|_F^2)
	+
	(\| \widehat{\mb{Q}}_1^\top \widehat{\mb{P}}_* \widehat{\mb{Q}}_{2A} \|_F^2-\| \mb{Q}_1^\top \widehat{\mb{P}}_* \mb{Q}_{2A} \|_F^2)\\
	&~~~~~~+(\| \widehat{\mb{Q}}_1^\top \mb{P}_* \widehat{\mb{Q}}_{2A} \|_F^2
	-\| \widehat{\mb{Q}}_1^\top \widehat{\mb{P}}_* \widehat{\mb{Q}}_{2A} \|_F^2)
	\\
	&\le \big| \| \mb{Q}_1^\top \mb{P}_* \mb{Q}_{2A} \|_F^2
	-\| \widehat{\mb{Q}}_1^\top \mb{P}_* \widehat{\mb{Q}}_{2A} \|_F^2\big|
	+
	\big|\| \widehat{\mb{Q}}_1^\top \widehat{\mb{P}}_* \widehat{\mb{Q}}_{2A} \|_F^2-\| \mb{Q}_1^\top \widehat{\mb{P}}_* \mb{Q}_{2A} \|_F^2\big|\\
	&=O_P(\delta_\theta).
\end{align*}
Hence,
$
\left|
\tr\big(\mb{Q}_1^\top \widehat{\mb{P}}_* \mb{Q}_{2A} (\mb{Q}_1^\top \widehat{\mb{P}}_* \mb{Q}_{2A})^\top\big)
-
\tr\big(\mb{Q}_1^\top \mb{P}_* \mb{Q}_{2A} (\mb{Q}_1^\top \mb{P}_* \mb{Q}_{2A})^\top\big)
\right|=O_P(\delta_\theta).
$

\section{Selection of Matrix Ranks}\label{sec: rank selection}

Following \citep{Shu18}, we select $r_k=\rank(\mb{\Sigma}_k)$ for $k=1,2$ and $r_{12}=\rank(\mb{\Sigma}_{12})$ by the ED method of \citep{Onat10} and the MDL-IC method of \citep{Song16}, respectively. 
Specifically, the ED method estimates $r_k$ by
\[
\hat{r}_k=\max\{\ell\le T_k: \hat{\lambda}_{k,\ell}-\hat{\lambda}_{k,\ell+1}\ge \delta\},
\]
where $\hat{\lambda}_{k,\ell}$ is the $\ell$-th eigenvalue of $\mb{Y}_k\mb{Y}_k^\top/n$, $T_k= \big|\{i: \hat{\lambda}_{k,i}\ge \frac{1}{m_k}\sum_{\ell=1}^{m_k}\hat{\lambda}_{k,\ell}\}\big|\wedge \frac{m_k}{10}$ with $m_k=n\wedge p_k$, and $\delta$ is calibrated as in Section IV of \citep{Onat10}.
If there exit two variables from different denoised datsets have a significant nonzero correlation detected by the normal approximation test of \citep{DiCi17},
then we conclude $r_{12}>0$. Otherwise,
the CDPA method is unnecessary due to no correlation between the two signal datasets. 
The MDL-IC method estimates nonzero $r_{12}$ by
\[
\hat{r}_{12}=\argmin_{r\in [1,\hat{r}_1\wedge \hat{r}_2]}\left\{
n\sum_{\ell=1}^r \log(1-s_\ell^2)+r(\hat{r}_1+\hat{r}_2-r)\log n
\right\},
\]
where $s_\ell$ is the $\ell$-th largest singular value of $(\mb{U}_{12}^{[:,1:\hat{r}_1]})^\top \mb{U}_{22}^{[:,1:\hat{r}_2]}$,
and the $i$-th column of $\mb{U}_{k2}\in \mathbb{R}^{n\times n}$ 
is the right-singular vector of $\mb{Y}_k$  corresponding to its $i$-th largest singular value.

\section{Additional Simulation Results}\label{sec: add siml}

Figures \ref{Fig: Setup 1, theta=0,15,30}--\ref{Fig: Setup 2, theta=45,60,75} display the simulation results for the CDPA estimators under Setups~1 and~2. The result analysis given in Section~\ref{CDPA sec: simulations} generally holds here.

\begin{sidewaysfigure}[p!]
	\begin{subfigure}{0.28\textwidth}
		\includegraphics[width=1\textwidth]{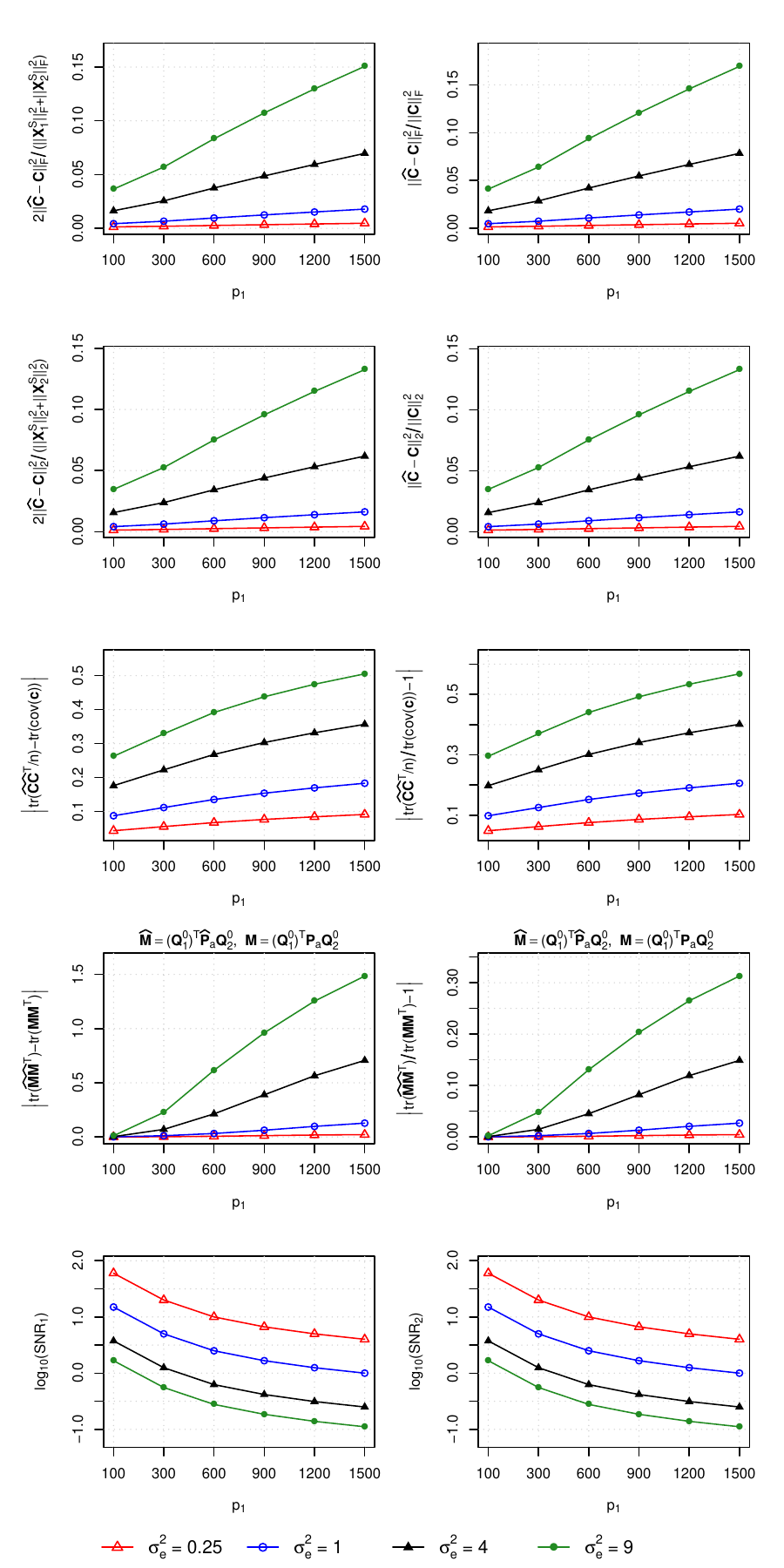}
		\caption{Setup 1 with $\theta=0^\circ$}
	\end{subfigure}
	\hspace{0.5cm}
	\begin{subfigure}{0.28\textwidth}
		\includegraphics[width=1\textwidth]{simulation1_angle15_plot.pdf}
		\caption{Setup 1 with  $\theta=15^\circ$}
	\end{subfigure}
	\hspace{0.5cm}
	\begin{subfigure}{0.28\textwidth}
		\includegraphics[width=1\textwidth]{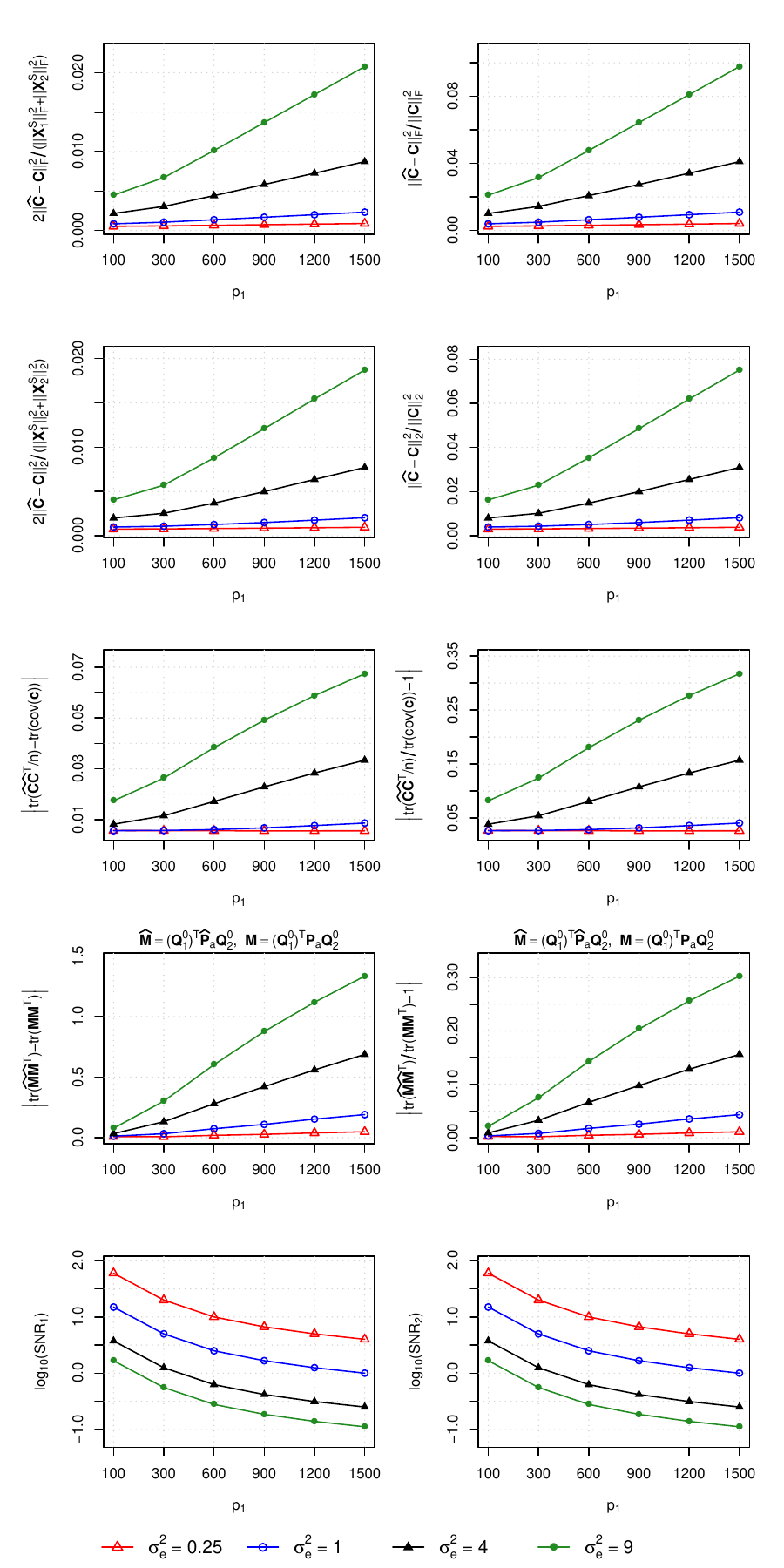}
		\caption{Setup 1 with  $\theta=30^\circ$}
	\end{subfigure}
	\caption{Average errors of CDPA estimates over 1000 replications and the signal-to-noise ratios for Setup 1 with $\theta\in \{0^\circ,15^\circ,30^\circ\}$.}
	\label{Fig: Setup 1, theta=0,15,30}
\end{sidewaysfigure}

\begin{sidewaysfigure}[p!]
	\begin{subfigure}{0.28\textwidth}
		\includegraphics[width=1\textwidth]{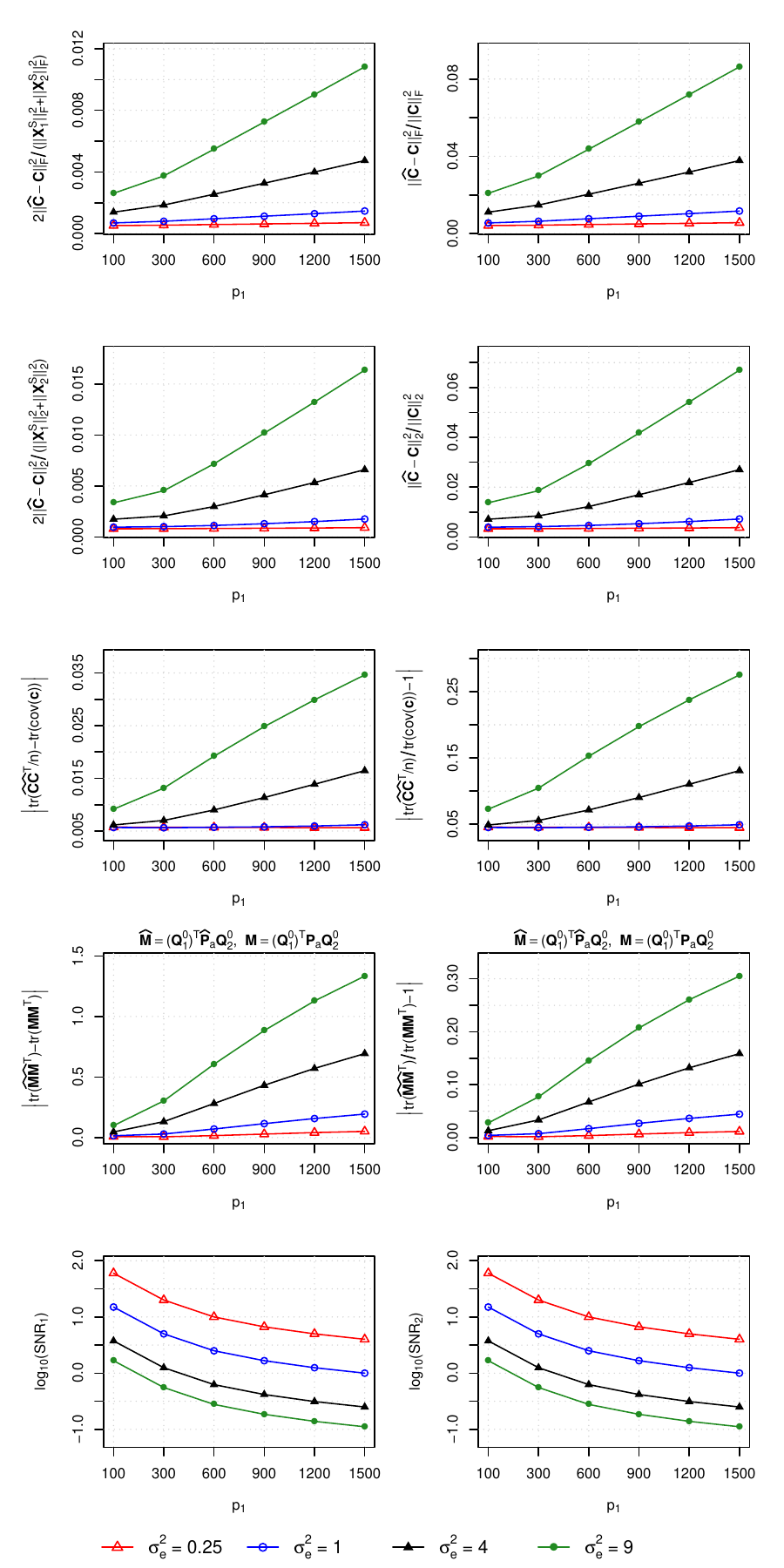}
		\caption{Setup 1 with $\theta=45^\circ$}
	\end{subfigure}
	\hspace{0.5cm}
	\begin{subfigure}{0.28\textwidth}
		\includegraphics[width=1\textwidth]{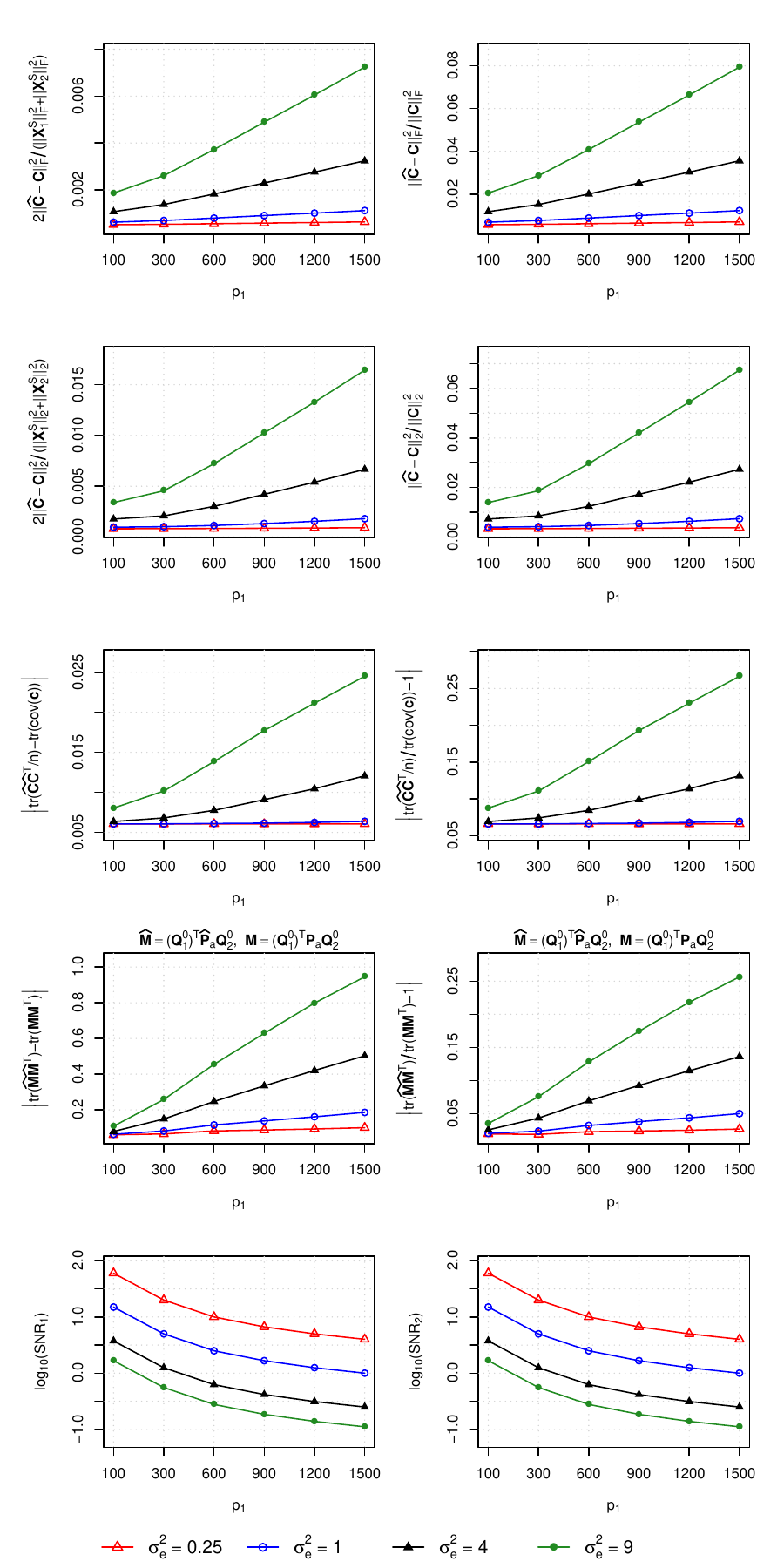}
		\caption{Setup 1 with  $\theta=60^\circ$}
	\end{subfigure}
	\hspace{0.5cm}
	\begin{subfigure}{0.28\textwidth}
		\includegraphics[width=1\textwidth]{simulation1_angle75_plot.pdf}
		\caption{Setup 1 with  $\theta=75^\circ$}
	\end{subfigure}
	\caption{Average errors of CDPA estimates over 1000 replications and the signal-to-noise ratios for Setup 1 with $\theta\in \{45^\circ,60^\circ,75^\circ\}$.}
	\label{Fig: Setup 1, theta=45, 60, 75}
\end{sidewaysfigure}

\begin{sidewaysfigure}[p!]
	\begin{subfigure}{0.28\textwidth}
		\includegraphics[width=1\textwidth]{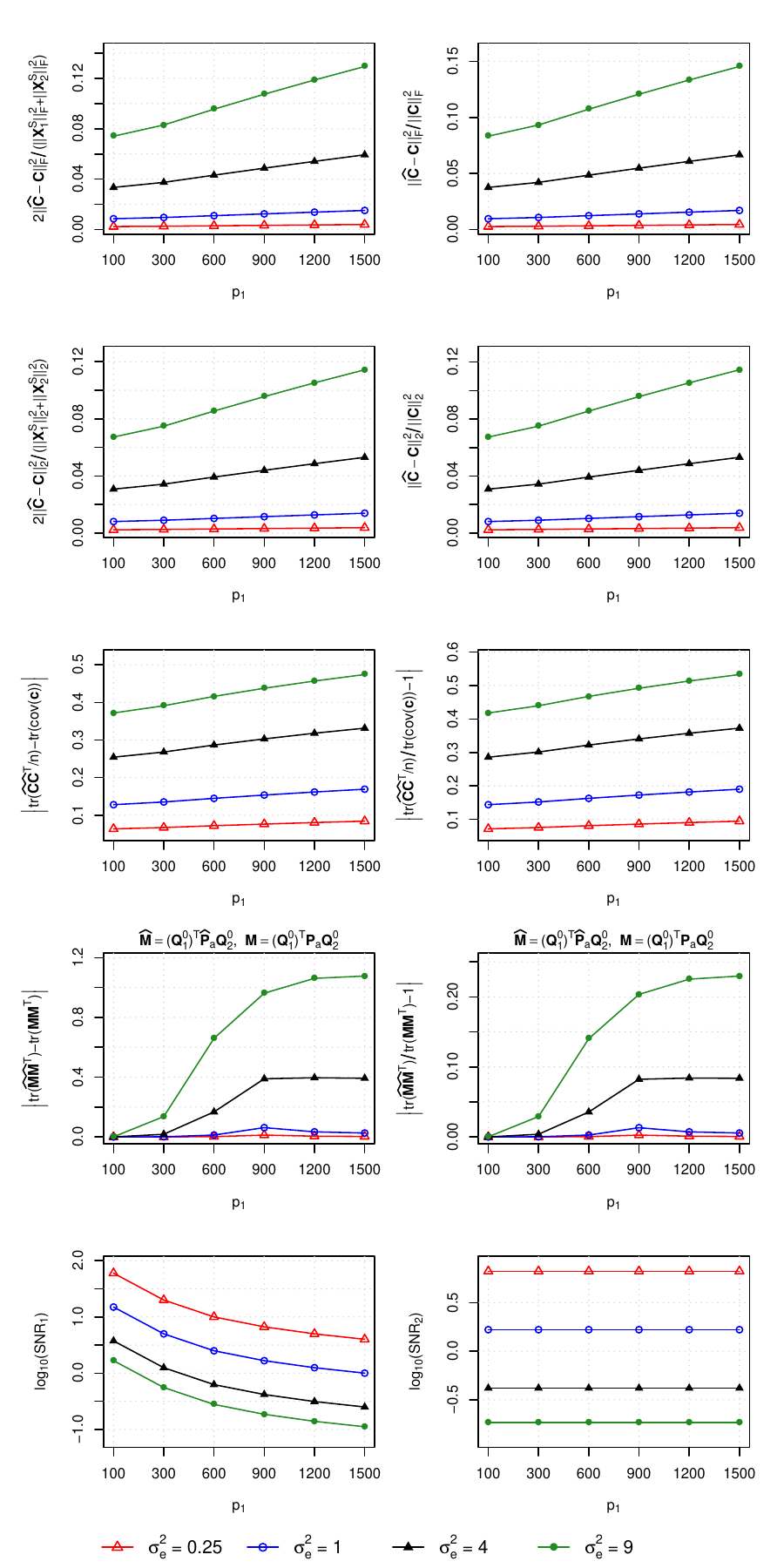}
		\caption{Setup 2 with $\theta=0^\circ$}
	\end{subfigure}
	\hspace{0.5cm}
	\begin{subfigure}{0.28\textwidth}
		\includegraphics[width=1\textwidth]{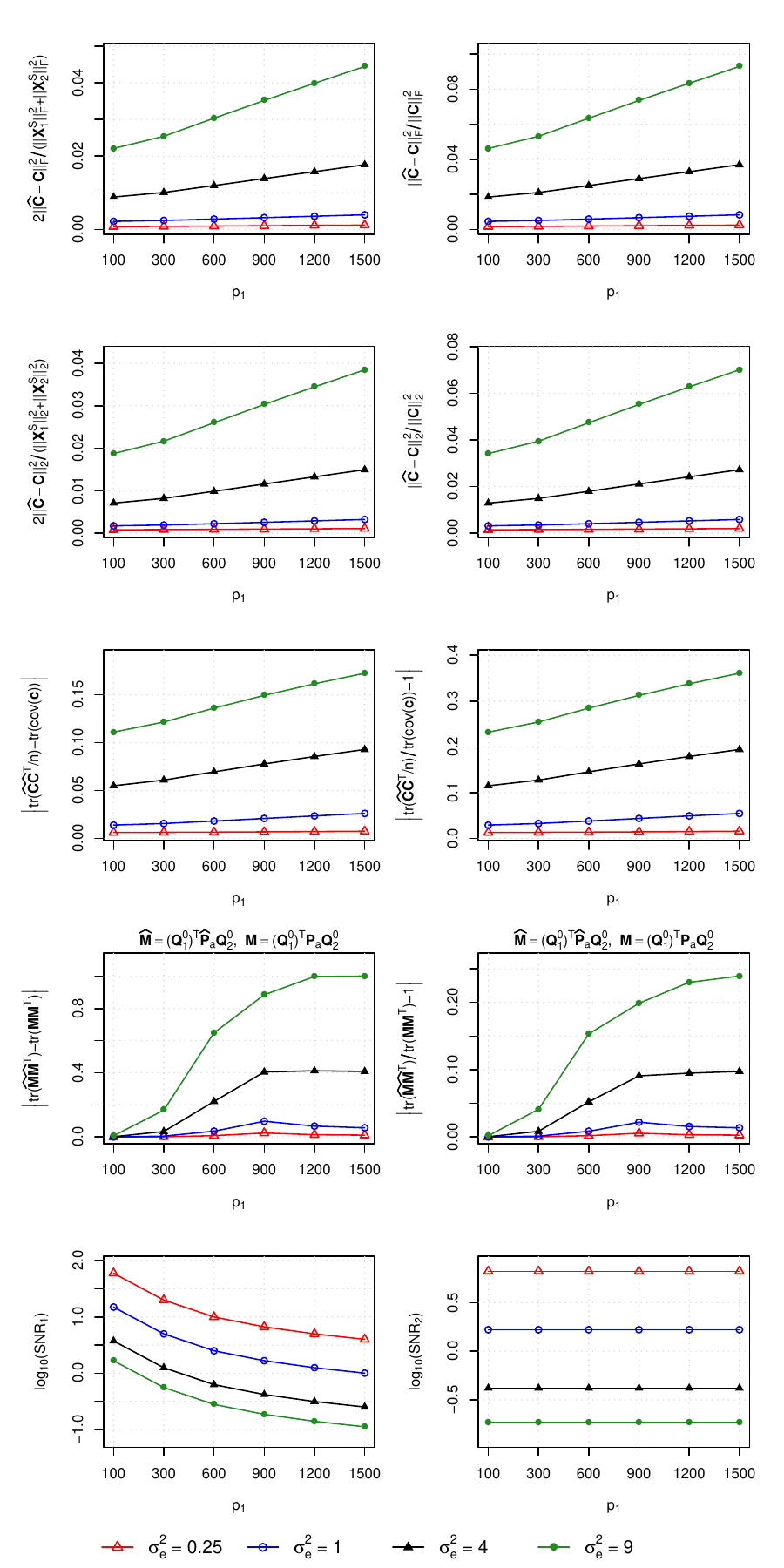}
		\caption{Setup 2 with  $\theta=15^\circ$}
	\end{subfigure}
	\hspace{0.5cm}
	\begin{subfigure}{0.28\textwidth}
		\includegraphics[width=1\textwidth]{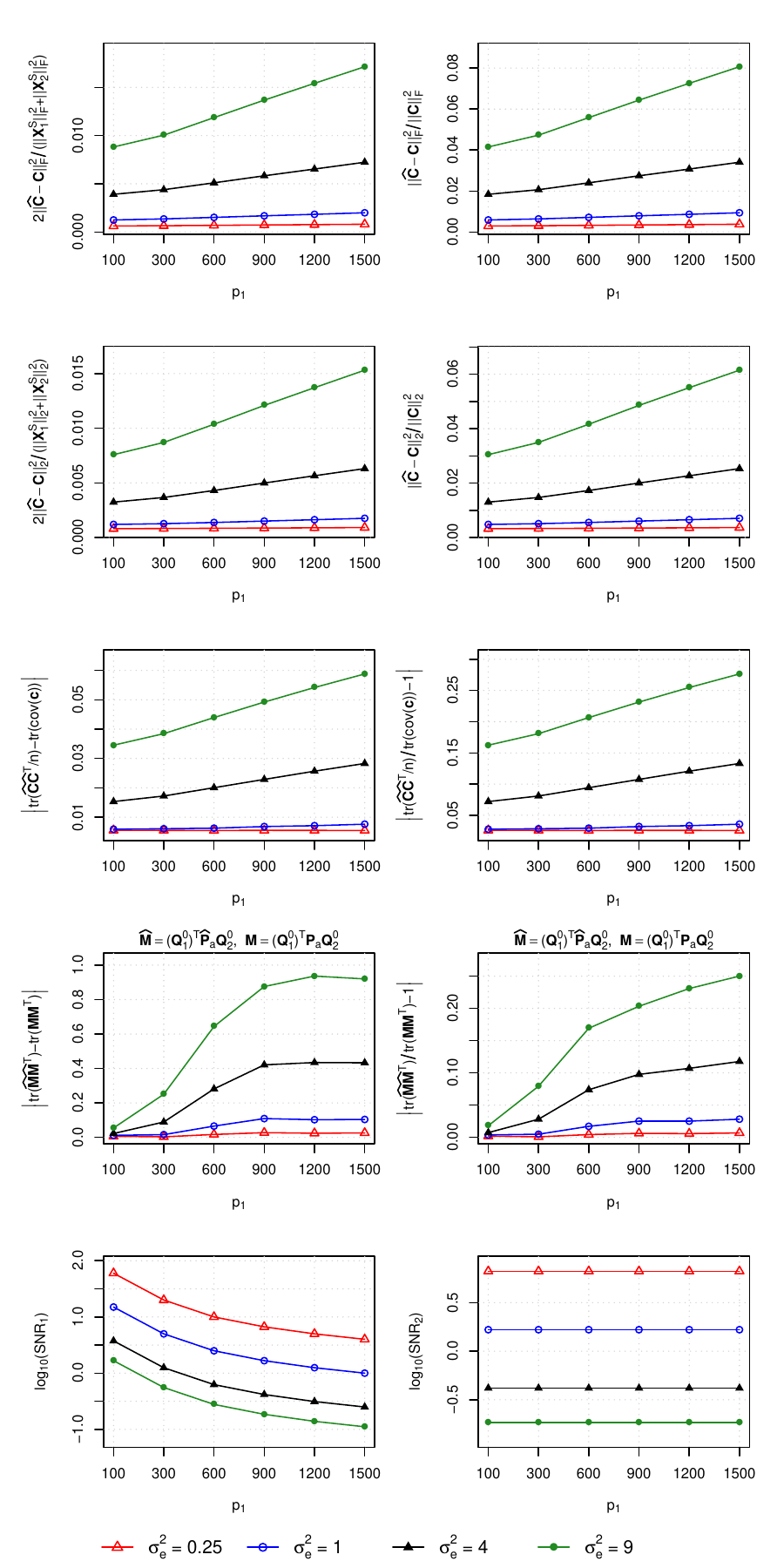}
		\caption{Setup 2 with  $\theta=30^\circ$}
	\end{subfigure}
	\caption{Average errors of CDPA estimates over 1000 replications and the signal-to-noise ratios for Setup 2 with $\theta\in \{0^\circ,15^\circ,30^\circ\}$.}
	\label{Fig: Setup 2, theta=0,15,30}
\end{sidewaysfigure}

\begin{sidewaysfigure}[p!]
	\begin{subfigure}{0.28\textwidth}
		\includegraphics[width=1\textwidth]{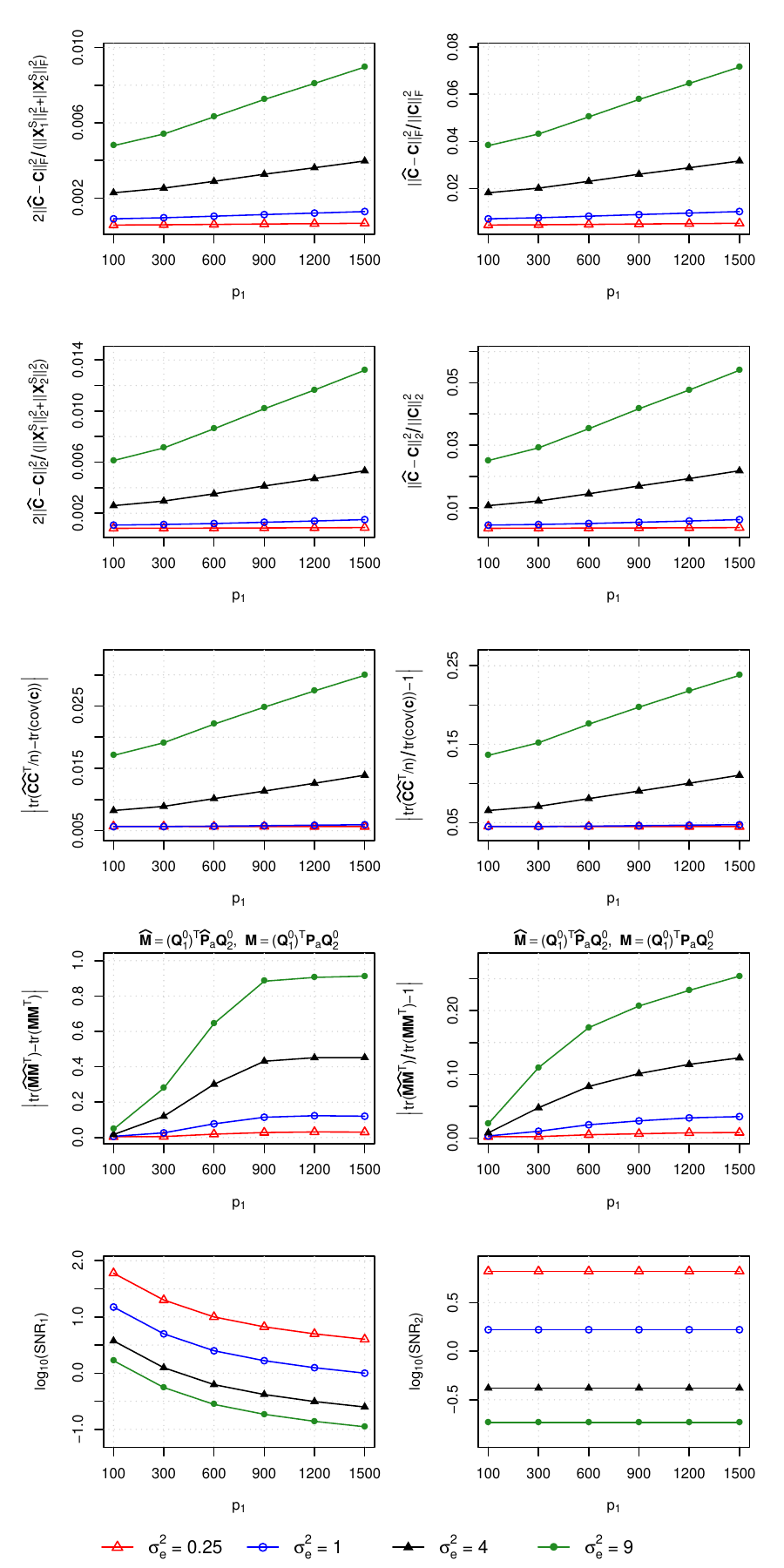}
		\caption{Setup 2 with $\theta=45^\circ$}
	\end{subfigure}
	\hspace{0.5cm}
	\begin{subfigure}{0.28\textwidth}
		\includegraphics[width=1\textwidth]{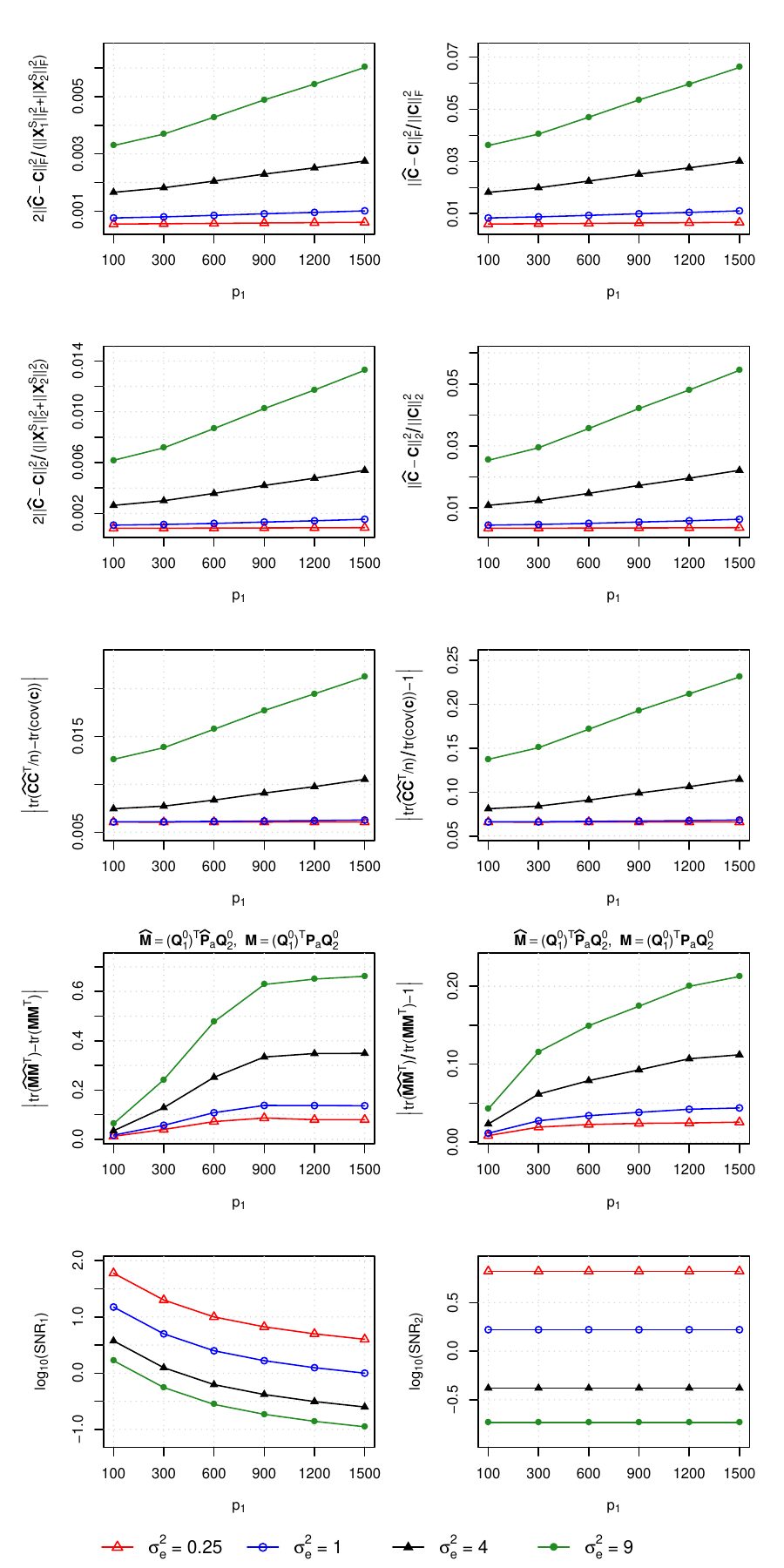}
		\caption{Setup 2 with  $\theta=60^\circ$}
	\end{subfigure}
	\hspace{0.5cm}
	\begin{subfigure}{0.28\textwidth}
		\includegraphics[width=1\textwidth]{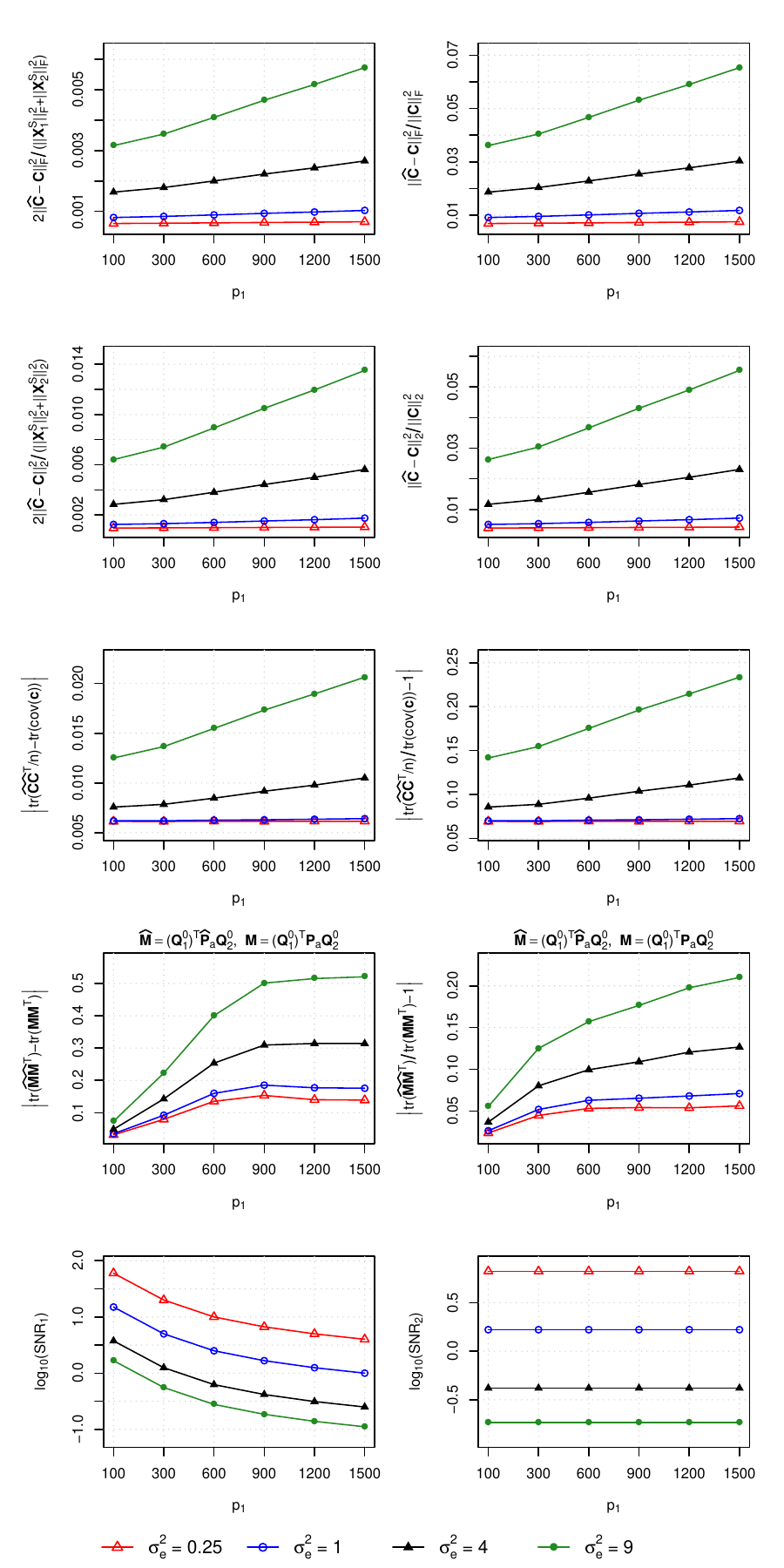}
		\caption{Setup 2 with  $\theta=75^\circ$}
	\end{subfigure}
	\caption{Average errors of CDPA estimates over 1000 replications and the signal-to-noise ratios for Setup 2 with $\theta\in \{45^\circ,60^\circ,75^\circ\}$.}
	\label{Fig: Setup 2, theta=45,60,75}
\end{sidewaysfigure}

\section{Additional Real-Data Results}\label{sec: add data analysis}

\subsection{Additional results of HCP motor-task functional MRI data}\label{sec: HCP app}

\begin{figure}[p!]	
	\begin{subfigure}{0.45\textwidth}
		\centering\includegraphics[width=1\textwidth]{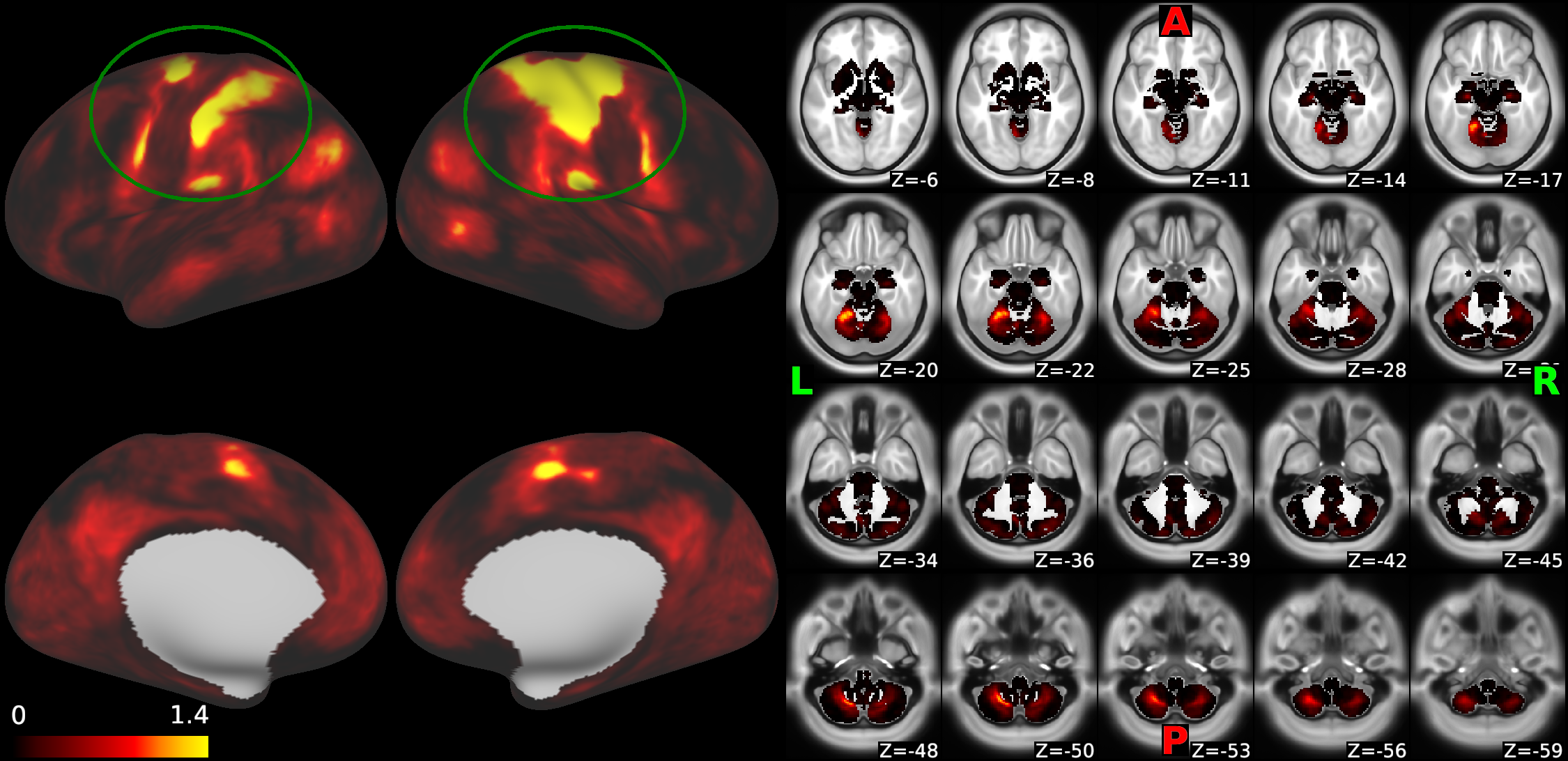}
		\caption{$\widehat{\var}(\bd{c}_L)$ of DISCO-SCA}
	\end{subfigure}
	\begin{subfigure}{0.45\textwidth}
		\centering\includegraphics[width=1\textwidth]{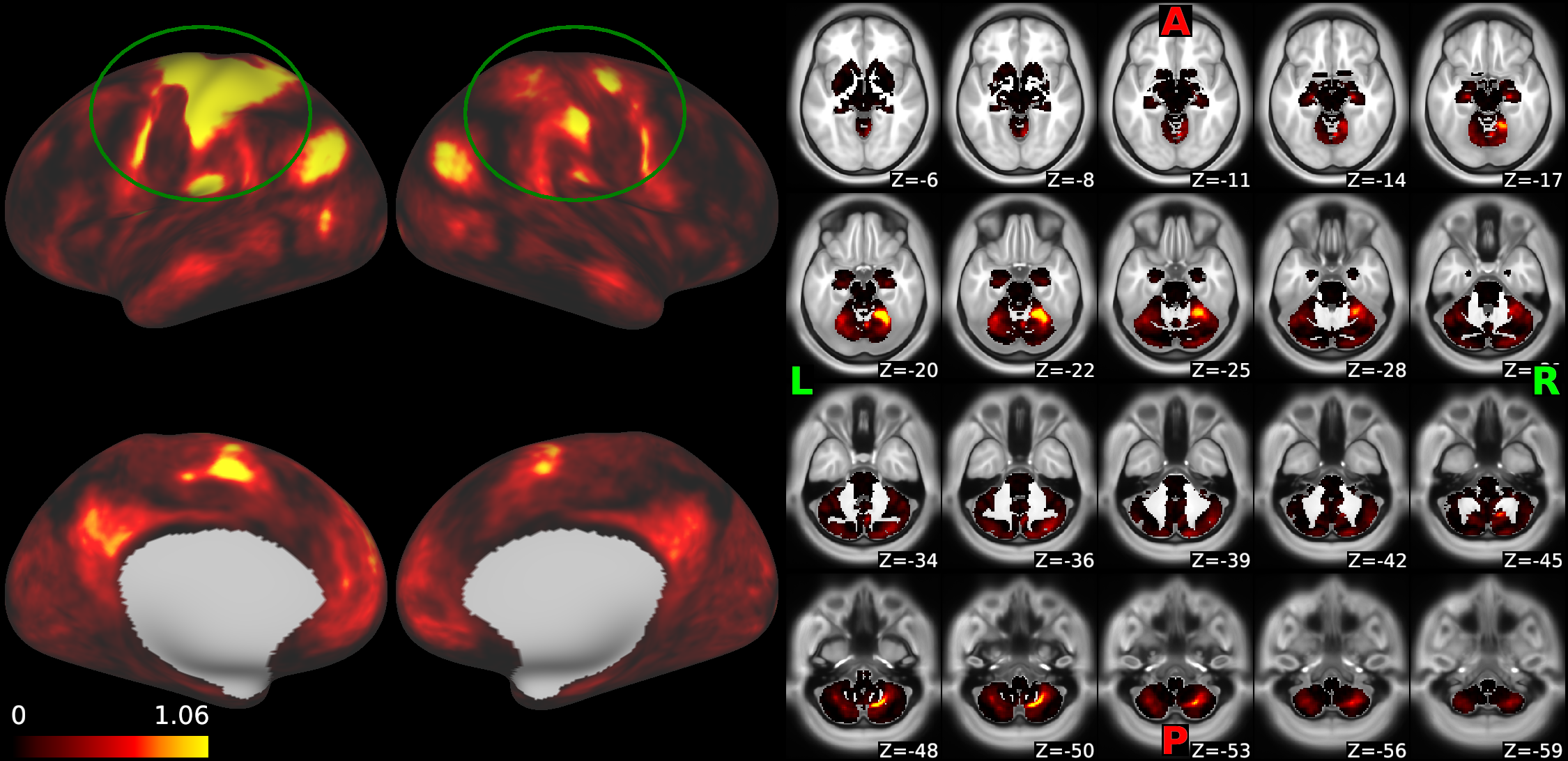}
		\caption{$\widehat{\var}(\bd{c}_R)$ of DISCO-SCA}
	\end{subfigure}
	\begin{subfigure}{0.45\textwidth}\bigskip
		\centering\includegraphics[width=1\textwidth]{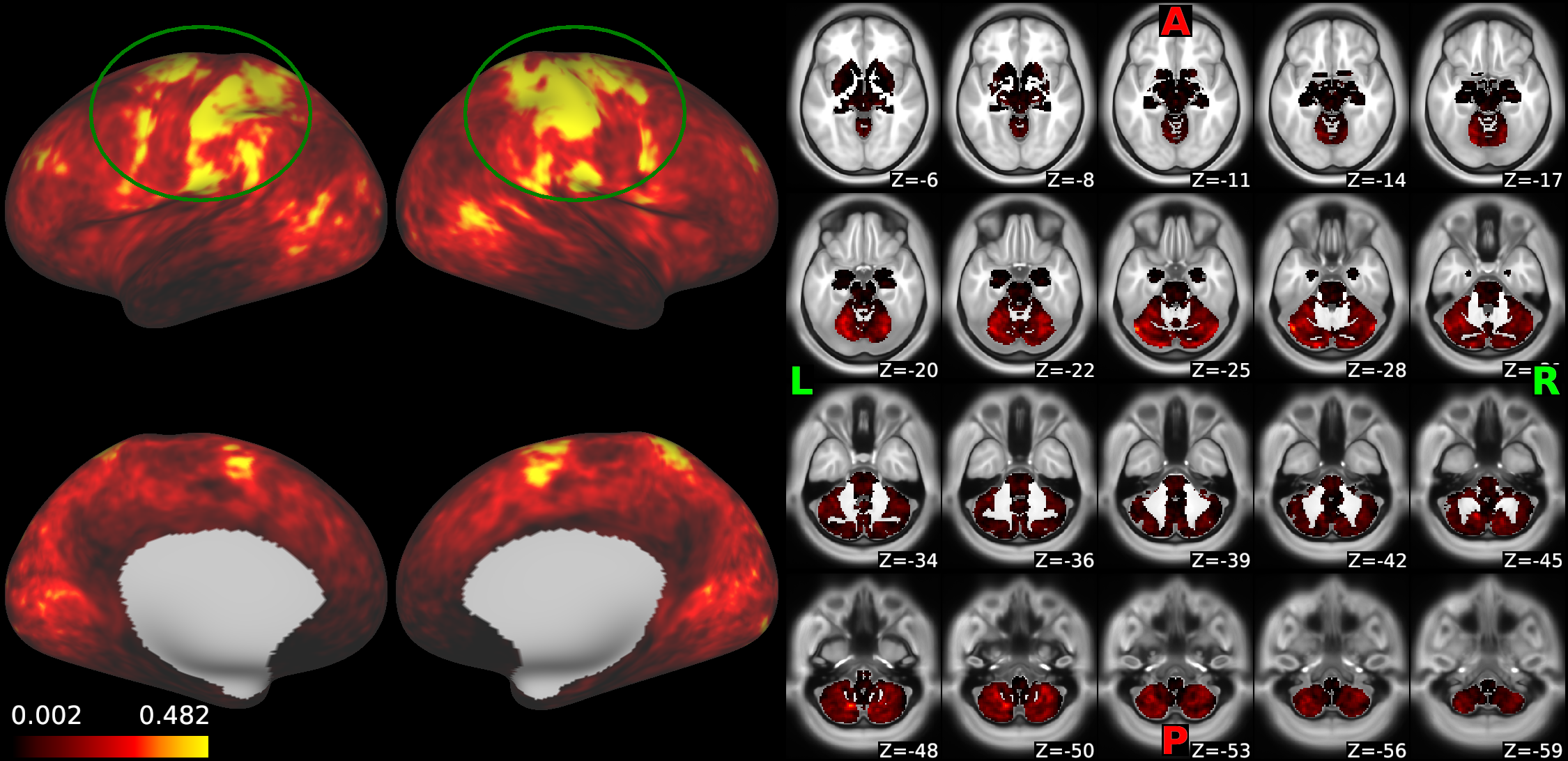}
		\caption{$\widehat{\var}(\bd{c}_L)$ of JIVE}
	\end{subfigure}
	\begin{subfigure}{0.45\textwidth}\bigskip
		\centering\includegraphics[width=1\textwidth]{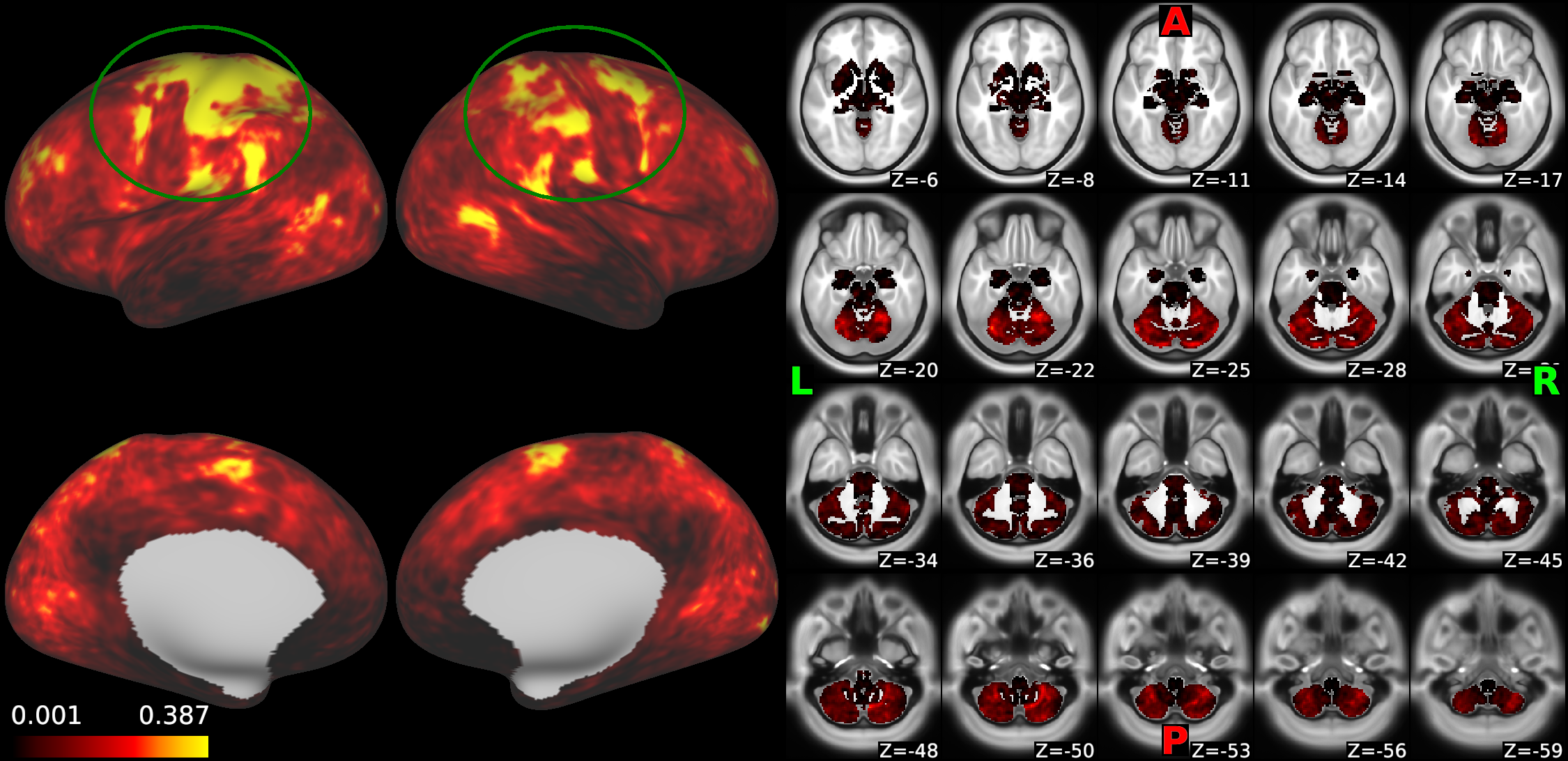}
		\caption{$\widehat{\var}(\bd{c}_R)$ of JIVE}
	\end{subfigure}
	\begin{subfigure}{0.45\textwidth}\bigskip
		\centering\includegraphics[width=1\textwidth]{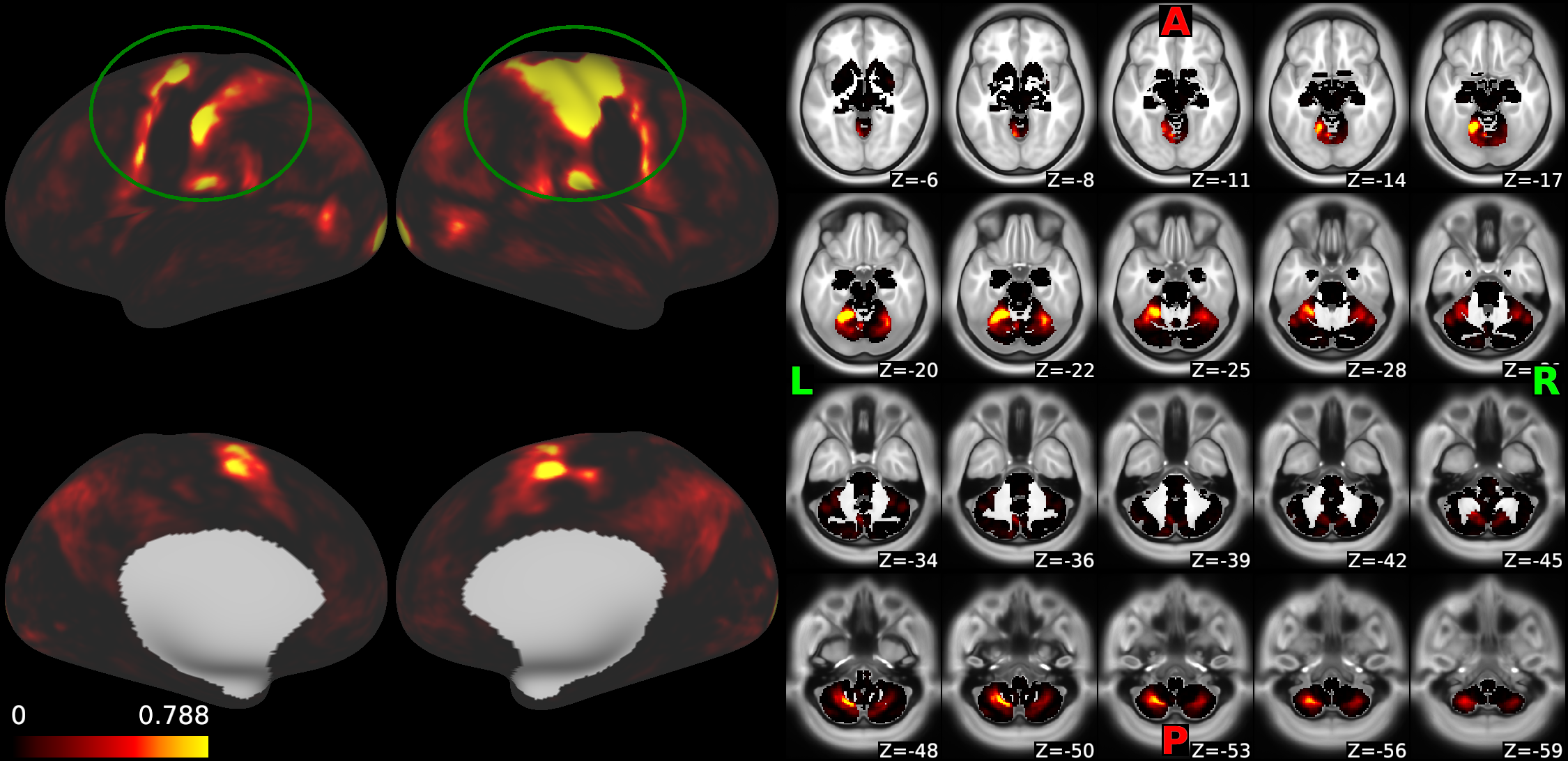}
		\caption{$\widehat{\var}(\bd{c}_L)$ of AJIVE}
	\end{subfigure}
	\begin{subfigure}{0.45\textwidth}\bigskip
		\centering\includegraphics[width=1\textwidth]{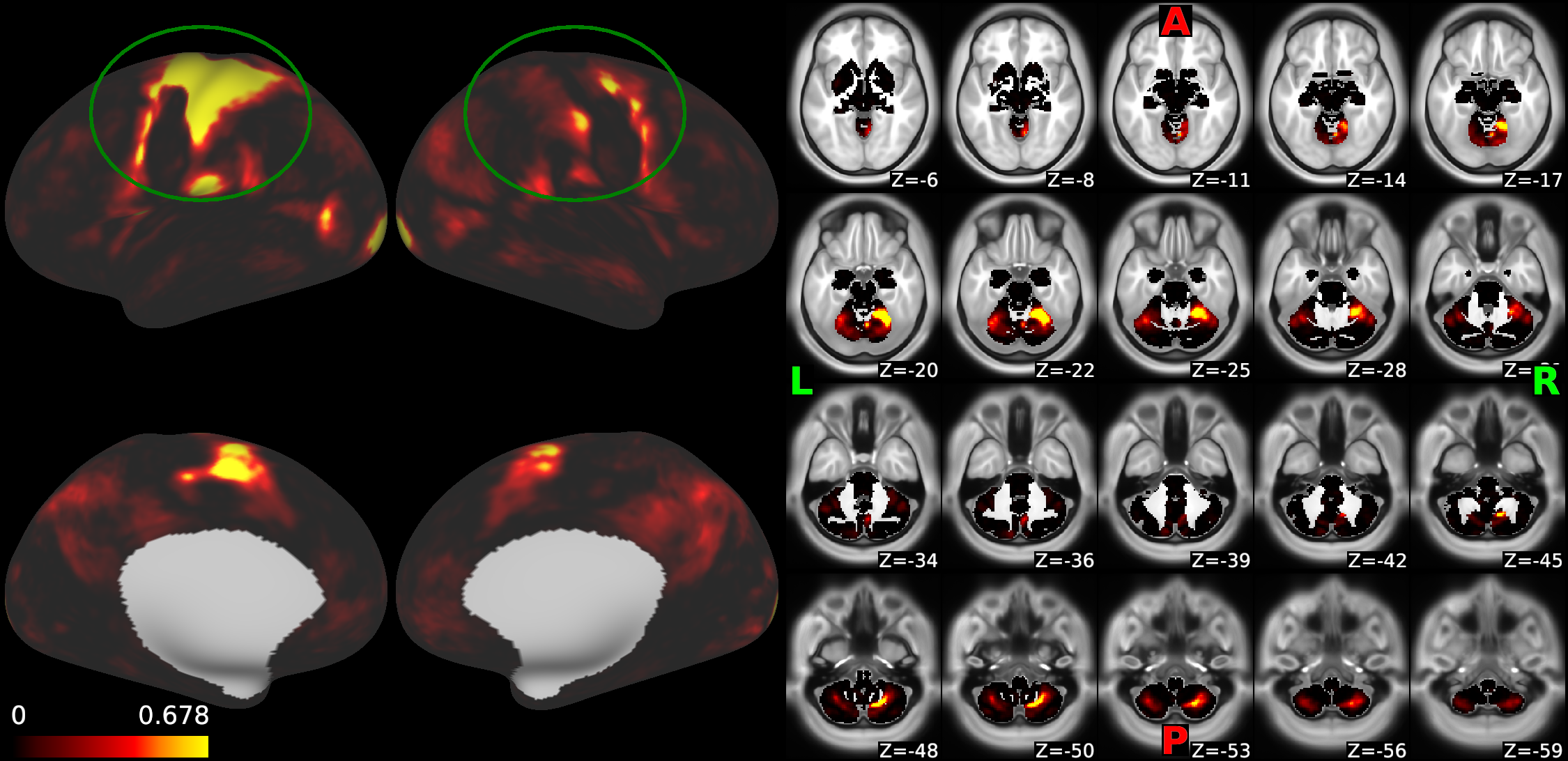}
		\caption{$\widehat{\var}(\bd{c}_R)$ of AJIVE}
	\end{subfigure}
	\caption{The variance maps estimated by the DISCO-SCA, JIVE, and AJIVE methods for HCP motor-task functional MRI data. The notation $\widehat{\var}$ denotes the sample variance vector obtained from the corresponding recovered sample matrix. In each subfigure, the left part displays the cortical surface with the outer side shown in the first row and the inner side in the second row; the right part shows the subcortical area on 20 $xy$ slides at the $z$ axis. The somatomotor cortex is annotated by green circles.}
	\label{Fig: Others: HCP real data}
\end{figure}

We also apply the five D-CCA-type methods (OnPLS, DISCO-SCA, COBE, JIVE, and AJIVE) to 
analyze the HCP motor-task functional MRI data.
The result of OnPLS is not available because this method exceeds the 62GB memory limit of our computing node
due to the SVD computation of the large 91,282$\times$91,282 matrix $\mb{Y}_L\mb{Y}_R^\top$ in its algorithm. The COBE method fails to generate nonzero common-source matrix estimates.
Figure~\ref{Fig: Others: HCP real data} shows the maps of $\widehat{\var}(\bd{c}_L)$ and $\widehat{\var}(\bd{c}_R)$ obtained from the DISCO-SCA, JIVE and AJIVE methods. Similar to those shown in Figure~\ref{CDPA: HCP real data} (c) and (d) for D-CCA, 
the common-source vectors $\bd{c}_L$ and  $\bd{c}_R$ of the three methods have estimated variance maps that are asymmetric on the two hemispheres, and thus are less plausible than the common-pattern vector $\bd{c}$ of CDPA to represent the common pattern of the left-hand and right-hand tasks on the brain.

\subsection{Additional results of TCGA breast cancer genomic datasets}

We also apply the same clustering method used in Section~\ref{sec: TCGA}
to each recovered matrix from the five D-CCA-type methods: OnPLS, COBE, JIVE, AJIVE, and DISCO-SCA.
Table~\ref{tab: log-rank table for others} reports the p-values of the log-rank test and the  Peto-Peto’s Wilcoxon test for the survival differences among the clusters from each of these matrices.
All the five methods have the p-values above 0.05 and thus fail to  discover breast cancer subtypes with significant survival differences.

\begin{table}[h!]
	\caption{Log-rank test and Peto-Peto's Wilcoxon test for survival curve differences among the clusters identified from each matrix of the five D-CCA-type methods for
		TCGA breast cancer datasets. }
	\begin{tabular}{lccc cc c cccc}
		\hline\noalign{\smallskip}
		& \multicolumn{5}{c}{Log-rank/Peto's p-values for competing methods}\\
		Data
		& OnPLS 
		&COBE
		& JIVE & AJIVE
		& DISCO-SCA
		\\
		\noalign{\smallskip}\hline\noalign{\smallskip}
		$\widehat{\mb{X}}_{\text{DNA}}$
		&0.340/0.568 &0.093/0.137 &0.585/0.389 &0.125/0.139 &0.774/0.866 \\
		$\widehat{\mb{X}}_{\text{mRNA}}$
		&0.060/0.078  &0.189/0.107 &0.577/0.589 &0.266/0.192 &0.175/0.116\\
		$[\widehat{\mb{X}}_{\text{DNA}}^N;\widehat{\mb{X}}_{\text{mRNA}}^N]$
		&0.461/0.506 &0.325/0.319 &0.207/0.225 &0.296/0.330 &0.452/0.517 \\ \noalign{\smallskip}
		$\widehat{\mb{C}}_{\text{DNA}}$
		&0.846/0.957  &NA &0.133/0.156 &0.213/0.193 &0.147/0.204\\
		$\widehat{\mb{C}}_{\text{mRNA}}$
		&0.060/0.078 & NA&0.133/0.156 &0.083/0.116 &0.205/0.097 \\
		$[\widehat{\mb{C}}_{\text{DNA}}^N;\widehat{\mb{C}}_{\text{mRNA}}^N]$
		&0.493/0.707  &NA &0.133/0.156 &0.321/0.240 &0.217/0.104\\	 \noalign{\smallskip}
		$\widehat{\mb{D}}_{\text{DNA}}$
		&0.618/0.559 &0.093/0.137 &0.137/0.086 &0.282/0.205 &0.791/0.657 \\
		$\widehat{\mb{D}}_{\text{mRNA}}$
		&NA &0.189/0.107 &0.074/0.076 &0.439/0.141 &0.846/0.842 \\
		$[\widehat{\mb{D}}_{\text{DNA}}^N;\widehat{\mb{D}}_{\text{mRNA}}^N]$
		&NA &0.325/0.319  &0.089/0.062 &0.155/0.187
		&0.614/0.594 	\\
		\noalign{\smallskip}\hline
	\end{tabular}

	\vspace{0.5\baselineskip}	
	{ Notes: Denote $\mb{M}^N=\mb{M}/\|\mb{M}\|_F$ for any matrix $\mb{M}$. NA means that the result is not available due to a zero matrix estimate.}
	\label{tab: log-rank table for others}
\end{table}


\bibliographystyle{imsart-number} 
\bibliography{refs}

\end{document}